\documentclass[lettersize,journal]{IEEEtran}
\usepackage{amsmath,amsfonts}
\usepackage{algorithmic}
\usepackage{algorithm}
\usepackage{color,array}
\usepackage[caption=false,font=normalsize,labelfont=sf,textfont=sf]{subfig}
\usepackage{textcomp}
\usepackage{stfloats}
\usepackage{url}
\usepackage{verbatim}
\usepackage{graphicx}
\usepackage{cite}
\usepackage{xcolor}
\usepackage[dvipsnames]{xcolor}
\usepackage{tgpagella}
\usepackage{multirow}
\usepackage{colortbl}
\usepackage[colorlinks,urlcolor=blue,linkcolor=blue,citecolor=blue]{hyperref}
\usepackage{orcidlink}
\usepackage{balance}
\usepackage{soul}
\usepackage{amssymb}
\usepackage{bm}
\usepackage{afterpage}
\usepackage{placeins}

\newcommand{\step}[2]{\bigskip\noindent{\em \hypertarget{step#1}{Step #1. #2:}}}
\DeclareRobustCommand{\stepref}[1]{Step~\hyperlink{step#1}{#1}}
\DeclareRobustCommand{\stepstwo}[2]{Steps~\hyperlink{step#1}{#1} and \hyperlink{step#2}{#2}}

\begin{document}

\title{Winsor-CAM: Human-Tunable Visual Explanations from Deep Networks via Layer-Wise Winsorization}

\author{Casey Wall\,\orcidlink{0009-0000-9973-6007}, 
Longwei Wang\,\orcidlink{0009-0002-0638-5637}, \textit{Member, IEEE,} 
Rodrigue Rizk\,\orcidlink{0000-0002-4392-4188}, \textit{Member, IEEE,} 
KC Santosh\,\orcidlink{0000-0003-4176-0236}
\textit{Senior, IEEE} 
        % <-this % stops a space
%\thanks{Submitted to IEEE Transactions on Pattern Analysis and Machine Intelligence.}% <-this % stops a space

    % \thanks{This work was supported in part by the National Science Foundation under Award \href{https://www.nsf.gov/awardsearch/showAward?AWD_ID=2346643}{\#2346643} and in part by the U.S. Department of Defense (DoD) under Award \href{https://dtic.dimensions.ai/details/grant/grant.14525543}{\#FA95502310495}.}
    \thanks{This work was supported by the National Science Foundation under Grant No. \href{https://www.nsf.gov/awardsearch/showAward?AWD_ID=2346643}{\#2346643}, the U.S. Department of Defense under Award No. \href{https://dtic.dimensions.ai/details/grant/grant.14525543}{\#FA9550-23-1-0495}, and the U.S. Department of Education under Grant No. P116Z240151.
Any opinions, findings, conclusions or recommendations expressed in this material are those of the author(s) and do not necessarily reflect the views of the National Science Foundation, the U.S. Department of Defense, or the U.S. Department of Education.}

    \thanks{C. Wall, L. Wang, R. Rizk, and KC Santosh are with \href{https://ai-research-lab.org}{USD Artificial Intelligence Research}, Department of Computer Science, University of South Dakota (e-mail: casey.wall@coyotes.usd.edu, \{longwei.wang, rodrigue.rizk, kc.santosh\}@usd.edu).}

    \thanks{We publicly release the code and demo notebooks at: \url{https://github.com/USD-AI-ResearchLab/Winsor-CAM}.}
    \thanks{This manuscript has been accepted for publication in IEEE TPAMI (2026).}
}

\maketitle

\begin{abstract}

Interpreting Convolutional Neural Networks (CNNs) is critical for safety-sensitive applications such as healthcare and autonomous systems. Popular visual explanation methods like Grad-CAM use a single convolutional layer, potentially missing multi-scale cues and producing unstable saliency maps. We introduce Winsor-CAM, a single-pass gradient-based method that aggregates Grad-CAM maps from all convolutional layers and applies percentile-based Winsorization to attenuate outlier contributions. A user-controllable percentile parameter $p$ enables semantic-level tuning from low-level textures to high-level object patterns. We evaluate Winsor-CAM on six CNN architectures using PASCAL VOC 2012 and PolypGen, comparing localization (IoU, center-of-mass distance) and fidelity (insertion/deletion AUC) against seven baselines including Grad-CAM, Grad-CAM++, LayerCAM, ScoreCAM, AblationCAM, ShapleyCAM, and FullGrad. On DenseNet121 with a subset of Pascal VOC 2012, Winsor-CAM achieves 46.8\% IoU and 0.059 CoM distance versus 39.0\% and 0.074 for Grad-CAM, with improved insertion AUC (0.656 vs. 0.623) and deletion AUC (0.197 vs. 0.242). Notably, even the worst-performing fixed $p$-value configuration outperforms FullGrad across all metrics. An ablation study confirms that incorporating earlier layers improves localization. Similar evaluation on PolypGen polyp segmentation further validates Winsor-CAM's effectiveness in medical imaging contexts. Winsor-CAM provides an efficient, robust, and human-tunable explanation tool for expert-in-the-loop analysis.
\end{abstract}

\begin{figure}[!tbp]
    \centering
    \includegraphics[width=0.489\textwidth, trim=8.2 9.3 6 8.1, clip]{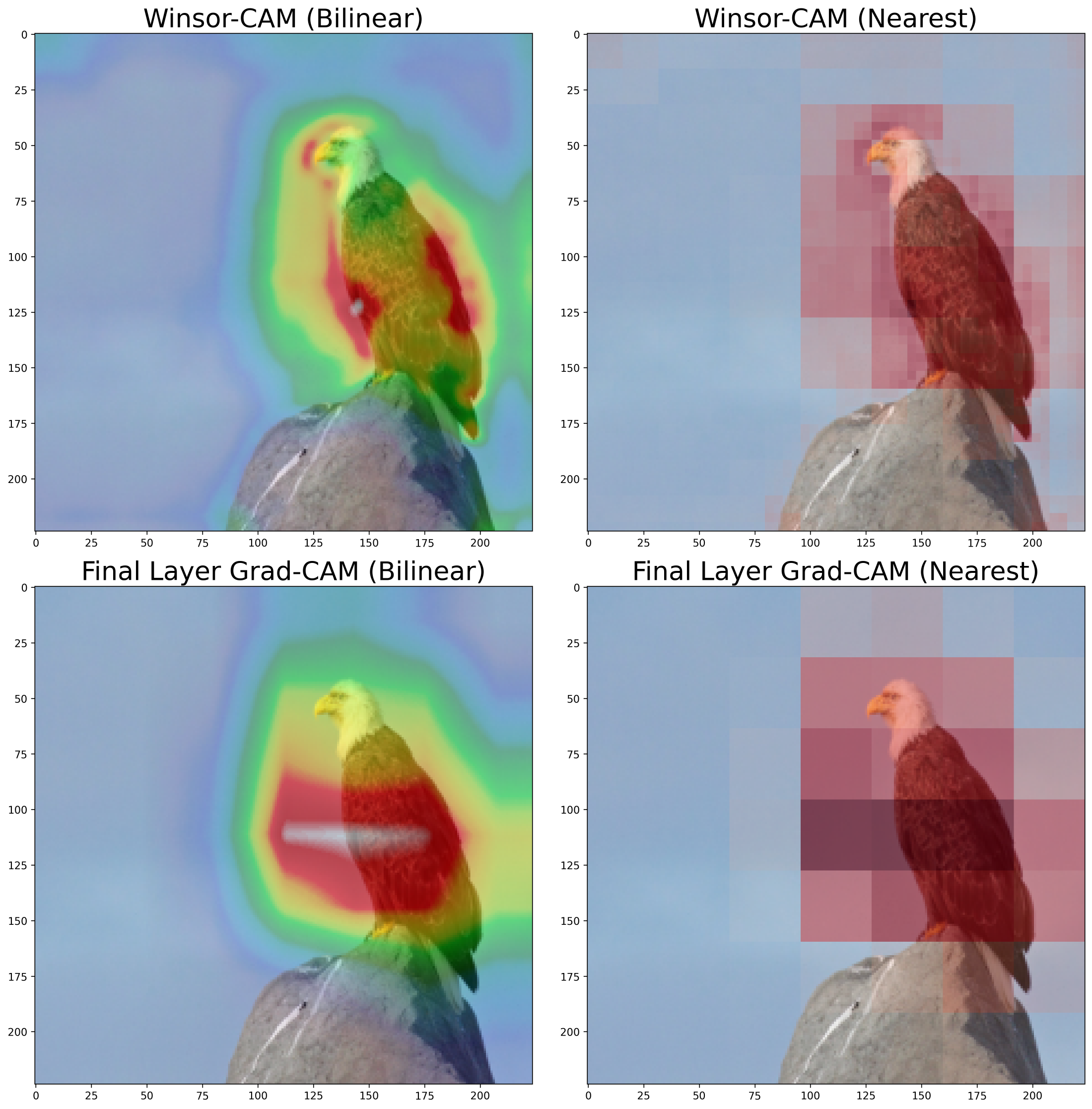}
    \caption{Comparison of Winsor-CAM and standard Grad-CAM outputs on a ResNet-50 model, illustrating improved localization and robustness to interpolation artifacts. Winsor-CAM produces smoother, semantically aligned heatmaps under both bilinear and nearest-neighbor upsampling, while Grad-CAM exhibits spatial distortion and noise, particularly under nearest interpolation. ImageNet validation set, class: ``bald eagle''.}
    \label{fig:Resnet50_interpolation}
\end{figure}

\begin{IEEEkeywords}
Explainable AI (XAI), Convolutional Neural Networks (CNNs), Winsorization, Saliency Maps.
\end{IEEEkeywords}

\section{Introduction}\label{sec:introduction}

\IEEEPARstart{C}{onvolutional} Neural Networks (CNNs) have achieved state-of-the-art performance across a variety of computer vision tasks, including image classification, object detection, and medical imaging. Despite these advances, the decision-making processes of CNNs remain largely opaque, raising concerns about accountability, reliability, and public trust, particularly in high-stakes domains such as healthcare, autonomous systems, and law enforcement. As a result, the field of Explainable Artificial Intelligence (XAI) has emerged with the goal of developing techniques that make neural network predictions more interpretable and transparent to human users~\cite{hamida2024exploring,longo2024explainable}.

Among the growing suite of XAI techniques, visual explanation methods have become especially prominent due to their intuitive appeal and applicability to spatially structured inputs such as images~\cite{simonyan2013deep,zeiler2014visualizing}. These methods generate saliency maps or heatmaps that highlight the most influential regions of an input image relevant to a model's prediction, offering users visual insight into the model's focus during inference. An example of generated heatmaps can be found in Fig.~\ref{fig:Resnet50_interpolation}. This capability supports a range of downstream objectives, from error diagnosis and model debugging to fairness auditing and scientific discovery~\cite{goyal2019counterfactual,hendricks2016generating,hendricks2018grounding,shrikumar2017deeplift,wagner2019interpretable}.

A seminal approach in this space is Class Activation Mapping (CAM), introduced by Zhou et al.~\cite{zhou2016learning}, which localizes discriminative regions by projecting classifier weights onto convolutional feature maps. CAM exploits the hierarchical and spatial structure of CNNs to trace decision-relevant regions at the final convolutional layer. Despite its effectiveness, CAM requires architectural modifications and is thus limited in flexibility. Subsequent developments, such as Gradient-weighted Class Activation Mapping (Grad-CAM)~\cite{gradcam}, addressed this limitation by leveraging gradients to compute class-specific importance scores, enabling visual explanations without modifying the underlying model architecture. Grad-CAM has since become a de facto standard for visual explanation due to its generality and ease of use in CNN-based XAI.

However, Grad-CAM suffers from a key limitation: it typically derives explanations from only the final convolutional layer. While this layer captures high-level semantic features, it may overlook low-level cues, such as textures or edges, learned in earlier layers. Furthermore, naïve extensions that uniformly average Grad-CAM outputs across layers can dilute semantically meaningful patterns by introducing noise from less relevant feature maps. These limitations motivate the need for explanation methods that incorporate multi-layer information while mitigating inter-layer variance and suppressing outlier dominance.

In this paper, we propose \textbf{Winsor-CAM}, a novel extension of Grad-CAM designed to overcome these limitations by leveraging saliency information from all convolutional layers in a CNN. The key innovation in Winsor-CAM is the integration of \textit{Winsorization}, a statistical clipping technique that suppresses extreme layer importance values. This technique enables semantic control via a user-adjustable percentile threshold, allowing users to dynamically adjust the semantic resolution of the output. 

\begin{figure}[!tbp]
    \centering

    \includegraphics[width=0.489\textwidth, trim=22 10 23 11, clip]{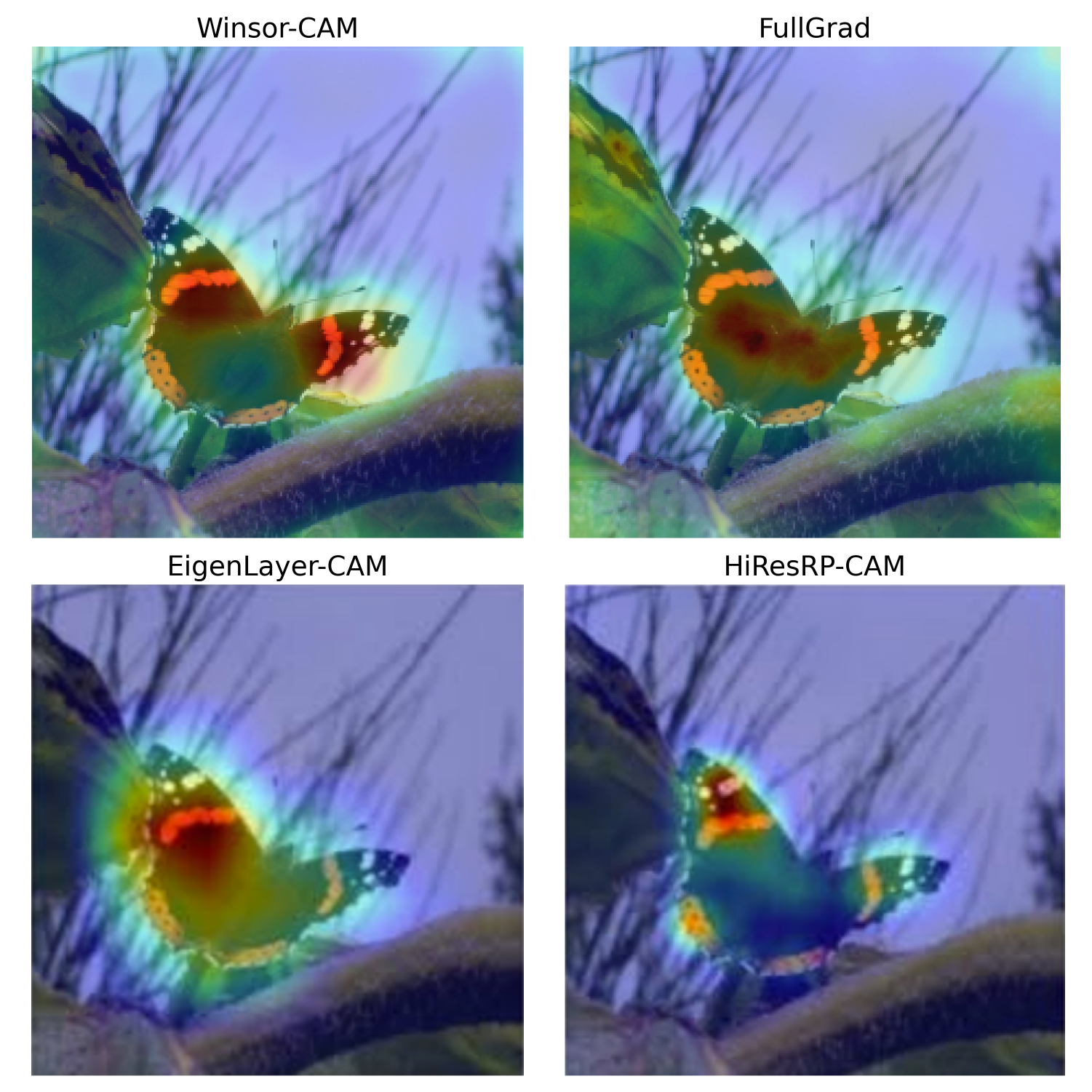}

    \caption{Comparison of multi-layer CAM-based methods on an image from ImageNet with the class ``Admiral'' (butterfly) with VGG16. Winsor-CAM (top left, $p=80$) and FullGrad~\cite{srinivas2019full} (top right) outputs were generated in this work. EigenLayer-CAM (bottom left) and HiResRP-CAM (bottom right) visualizations are adapted from~\cite{Andrei2025} for comparison, licensed under CC BY 4.0.}
    \label{fig:winsor_full_eigen_hires}
\end{figure}

Our contributions are as follows: 

\begin{enumerate}
\item We present Winsor-CAM, the first method to aggregate Grad-CAM explanations across the entire convolutional stack while applying robust outlier attenuation via Winsorization.
\item We introduce a human-controllable percentile parameter to tune the semantic abstraction level of the explanations.
\item We provide a comprehensive evaluation across standard CNN architectures and demonstrate improved interpretability and localization fidelity.
\item We show that Winsor-CAM outperforms commonly used baselines like final layer Grad-CAM, naïve layer aggregation, Grad-CAM++~\cite{chattopadhay2018grad}, LayerCAM~\cite{jiang2021layercam}, ShapleyCAM~\cite{cai2025cams}, ScoreCAM~\cite{wang2020score}, AblationCAM~\cite{desai2020ablationcam}, and FullGrad~\cite{srinivas2019full} in terms of visual coherence and spatial alignment with ground truth segmentation masks.
\item We provide an ablation study linking layer inclusion depth to localization performance.
\end{enumerate}

To validate our method, we conduct comprehensive experiments using six CNN architectures---ResNet50~\cite{Kaiming2016}, DenseNet121~\cite{Huang2017}, VGG16~\cite{simonyan2015}, InceptionV3~\cite{Szegedy2016}, Efficientnet-B0~\cite{efficientnet2019}, and ConvNeXt-Tiny~\cite{convnext2022}---on the PASCAL VOC 2012~\cite{pascal-voc-2012} dataset. We evaluate Winsor-CAM both qualitatively and quantitatively using Intersection over Union (IoU), Euclidean distance between the Center-of-Mass (CoM) of the saliency map and ground truth segmentation mask, insertion AUC, and deletion AUC. To demonstrate Winsor-CAM's upper-bound performance, we select the optimal percentile parameter $p$ per image (maximizing IoU and insertion AUC, minimizing CoM distance and deletion AUC), simulating a human-in-the-loop scenario where users tune $p$ based on interpretive needs. Our results show that Winsor-CAM with optimal per-image $p$-value selection outperforms final-layer Grad-CAM and naïve layer aggregation across all metrics in terms of visual interpretability and spatial localization. Notably, even when $p$ is fixed across all images (eliminating oracle selection), the worst-performing configuration still outperforms FullGrad, a method that also aggregates across all layers, across all metrics, demonstrating Winsor-CAM's robustness independent of parameter tuning. Additionally, we show results on the PolypGen~\cite{PolypGen2023} medical imaging dataset as a domain-specific validation of Winsor-CAM's generalizability. The tunable percentile parameter enables targeted exploration of feature hierarchies, allowing users to emphasize low- or high-level representations as needed, making Winsor-CAM particularly well-suited for expert-in-the-loop interpretability, diagnostic decision support, and interactive explainability workflows.

\section{Related Work}\label{sec:lit_review}

\subsection{Background}\label{sec:background}

Visual explanation techniques for deep neural networks, particularly CNNs, aim to localize input regions most relevant to a model's prediction. Broadly, these methods fall into three categories: \textbf{gradient-based}, \textbf{perturbation-based}, and \textbf{intrinsic methods}. Gradient-based approaches exploit internal model signals such as activations and gradients, requiring white-box access, whereas perturbation-based methods externally probe the model through input modifications (black-box). Intrinsic methods, like attention in transformers, provide interpretability inherently through architectural design. Winsor-CAM focuses on gradient-based multi-layer aggregation for CNNs, but we also discuss hybrid and perturbation-based methods~\cite{gradcam,chattopadhay2018grad,jiang2021layercam,fu2020axiom,yamauchi2024spatial,dong2023rethinking,yamauchi2024explaining,cai2025cams,smilkov2017,omeiza2019smooth}.

\subsubsection{Gradient-Based Methods}\label{sec:gradient_based_methods}

Gradient-based methods compute derivatives of the model output with respect to activations or input features to produce saliency maps. They can be categorized by computational cost: \textbf{single-pass methods}, which require one forward and backward pass, and \textbf{multi-pass methods}, which perform multiple passes for integration or iterative refinement.

% Winsorization - label -> fig:Overlay
\begin{figure*}[t]
    \centering
    \includegraphics[width=1\textwidth, trim=1.5 .5 3 2.5, clip]{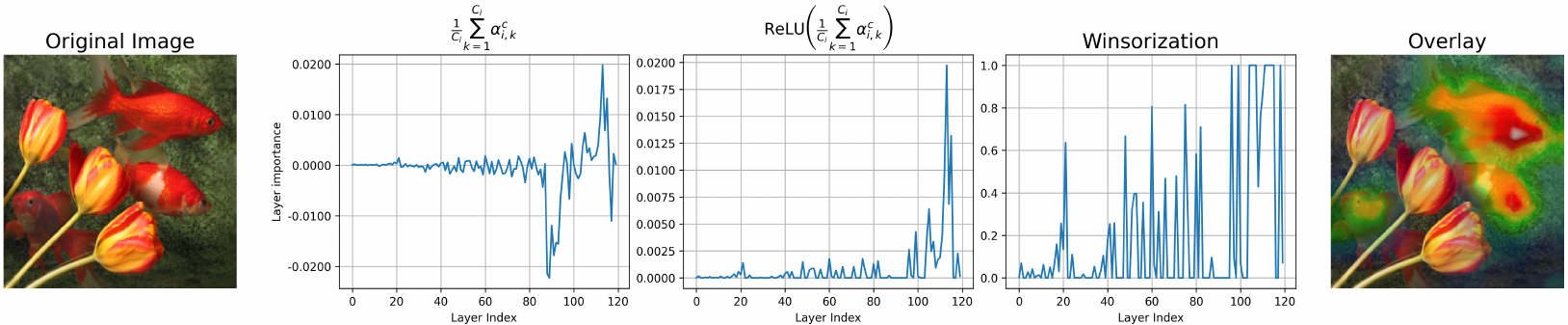}

    \caption{Step-by-step visualization of the Winsor-CAM pipeline for layer-wise mean importance. From left to right: input image (ImageNet validation set, class: ``goldfish''), raw layer-wise importance scores (before ReLU), positive importance scores after ReLU $\Gamma^c_i$ (Eq.~\eqref{eq:layer_importance_mean}), Winsorized importance on normalized scale (\stepstwo{4}{5}), and final Winsor-CAM heatmap overlay. This shows how layer-wise importance is extracted, outliers suppressed via Winsorization, and importance structure preserved.}
    \label{fig:Overlay}
\end{figure*}

\smallskip
\noindent 
\textbf{Single-Pass Gradient Methods.}
Classic single-pass CAM methods, such as Grad-CAM~\cite{gradcam} and representative gradient-based variants (e.g., Grad-CAM++~\cite{chattopadhay2018grad}, LayerCAM~\cite{jiang2021layercam}, Axiom-based Grad-CAM~\cite{fu2020axiom}, HiResRP-CAM~\cite{Andrei2025}, EigenLayer-CAM~\cite{Andrei2025}, and FullGrad~\cite{srinivas2019full}), compute gradient-weighted activations to generate saliency maps. These approaches are efficient and often high-resolution, but they typically focus on individual layers, are sensitive to sparse gradients, and lack systematic mechanisms for aggregating multi-layer information with outlier suppression. Closely related single-pass attribution methods, such as DeepLIFT~\cite{shrikumar2017deeplift} and Layer-wise Relevance Propagation (LRP)~\cite{montavon2019lrp}, propagate relevance scores through the network using reference-dependent or conservation-based rules rather than gradients, which can limit robustness or generality in practice.

\smallskip
\noindent 
\textbf{Multi-Pass Gradient Methods.}
Multi-pass approaches, including Integrated Gradients (IG)~\cite{sundararajan2017integrated} and Expected Grad-CAM~\cite{buono2024expected}, aggregate gradients over multiple forward-backward passes to satisfy axioms such as sensitivity and implementation invariance. ShapleyCAM~\cite{cai2025cams} extends CAM-based attribution by modeling the prediction process as a cooperative game and approximating Shapley values using gradients and Hessian-vector products. This yields theoretically grounded saliency maps with improved fidelity, but requires multiple backward passes and incurs higher computational cost. While effective, these methods achieve faithfulness through repeated inference rather than architectural or statistical mechanisms for balancing multi-layer contributions during aggregation.

\smallskip
\noindent 
\textbf{Multi-Layer Attribution.}
To address the limitations of single-layer saliency, methods like FullGrad~\cite{srinivas2019full} and HiResRP-CAM~\cite{Andrei2025} and EigenLayer-CAM~\cite{Andrei2025} aggregate gradients across all convolutional layers. FullGrad sums gradients and bias contributions uniformly, allowing deeper layers with larger activations to dominate. HiResRP-CAM propagates CAM-based relevance through layers via LRP to achieve higher-resolution, class-sensitive explanations, while EigenLayer-CAM combines SVD of activation maps with positive gradient scaling to yield class-discriminative visualizations. Fig.~\ref{fig:winsor_full_eigen_hires} visually compares Winsor-CAM outputs with these multi-layer aggregation methods. While effective, these approaches typically rely on fixed or implicit weighting schemes across layers and lack explicit mechanisms for controlling the influence of extreme activations and gradients during aggregation.

\subsubsection{Perturbation-Based Methods}\label{sec:perturbation_based_methods}

Perturbation-based methods evaluate feature importance by measuring output changes under input modifications. Examples include occlusion sensitivity~\cite{zeiler2014visualizing}, LIME~\cite{ribeiro2016lime}, SHAP~\cite{lundberg2017shap}, and RISE~\cite{petsiuk2018rise}. Although model-agnostic and theoretically grounded, these approaches are computationally expensive due to repeated forward passes and do not provide layer-wise insight. Extensions such as Collective Attribution~\cite{yamauchi2024explaining} apply Shapley-based techniques to object detection, but remain limited by sampling complexity.

\subsubsection{Intrinsic Methods}\label{sec:intrinsic_methods}
Intrinsic methods derive interpretability directly from model architecture rather than post hoc analysis. In transformer-based models, attention mechanisms provide built-in attribution by quantifying token-level interactions~\cite{vaswani2017attention}. Vision Transformers (ViTs)~\cite{vit2020} leverage multi-head self-attention to offer transparent information flow tracing. While attention is not the only form of intrinsic interpretability, it has become a dominant paradigm in transformer architectures. However, intrinsic methods are limited to specific architectures and may not generalize to CNNs without major modifications.

\subsubsection{Hybrid Methods}\label{sec:hybrid_methods}

Hybrid methods combine gradient-based, perturbation-driven, and attention-based components in flexible configurations to reduce noise and improve attribution quality. SmoothGrad~\cite{smilkov2017} and Smooth Grad-CAM++~\cite{omeiza2019smooth} average gradients over multiple noisy input copies, producing less visually noisy maps but requiring multiple forward-backward passes. Recent hybrid approaches focus on multi-layer fusion: TAME~\cite{TAME2022} employs hierarchical attention to combine feature maps from multiple convolutional layers, while UnionCAM~\cite{UnionCAM2024} fuses multiple CAM variants with denoising. These methods improve robustness and resolution, but generally require additional training, architectural integration, or multi-pass computation.

\subsection{Methodological Motivation}\label{sec:method_motivation}

Gradient-based multi-layer attribution methods face three key limitations that Winsor-CAM addresses through percentile-based outlier suppression:

\smallskip
\noindent
\textbf{Compared to multi-pass methods:} ShapleyCAM~\cite{cai2025cams} and IG~\cite{sundararajan2017integrated} achieve robustness through repeated inference (path integration with 20-50 steps or marginal contribution sampling), requiring $\mathcal{O}(n)$ forward-backward passes. The proposed Winsor-CAM method achieves comparable robustness in a \emph{single} forward-backward pass (\stepref{1}) by statistically suppressing outlier layer contributions (\stepref{4}) rather than averaging over multiple model evaluations, matching the single-pass efficiency of standard Grad-CAM and FullGrad while avoiding the repeated network evaluations required by multi-pass methods.

\smallskip
\noindent
\textbf{Compared to architectural propagation:} LRP~\cite{montavon2019lrp} and HiResRP-CAM~\cite{Andrei2025} enforce local conservation via backward rules but do not globally attenuate layers with disproportionately large contributions. Winsor-CAM operates on \emph{scalar} layer importance scores (\stepref{3}) rather than spatial maps, enabling global suppression across network depth while preserving fine-grained spatial structures within layers.

% pipeline - label -> fig:pipeline
\begin{figure*}[t]
    \centering
    \includegraphics[width=1\textwidth, trim=.1 1.75 .2 .2, clip]{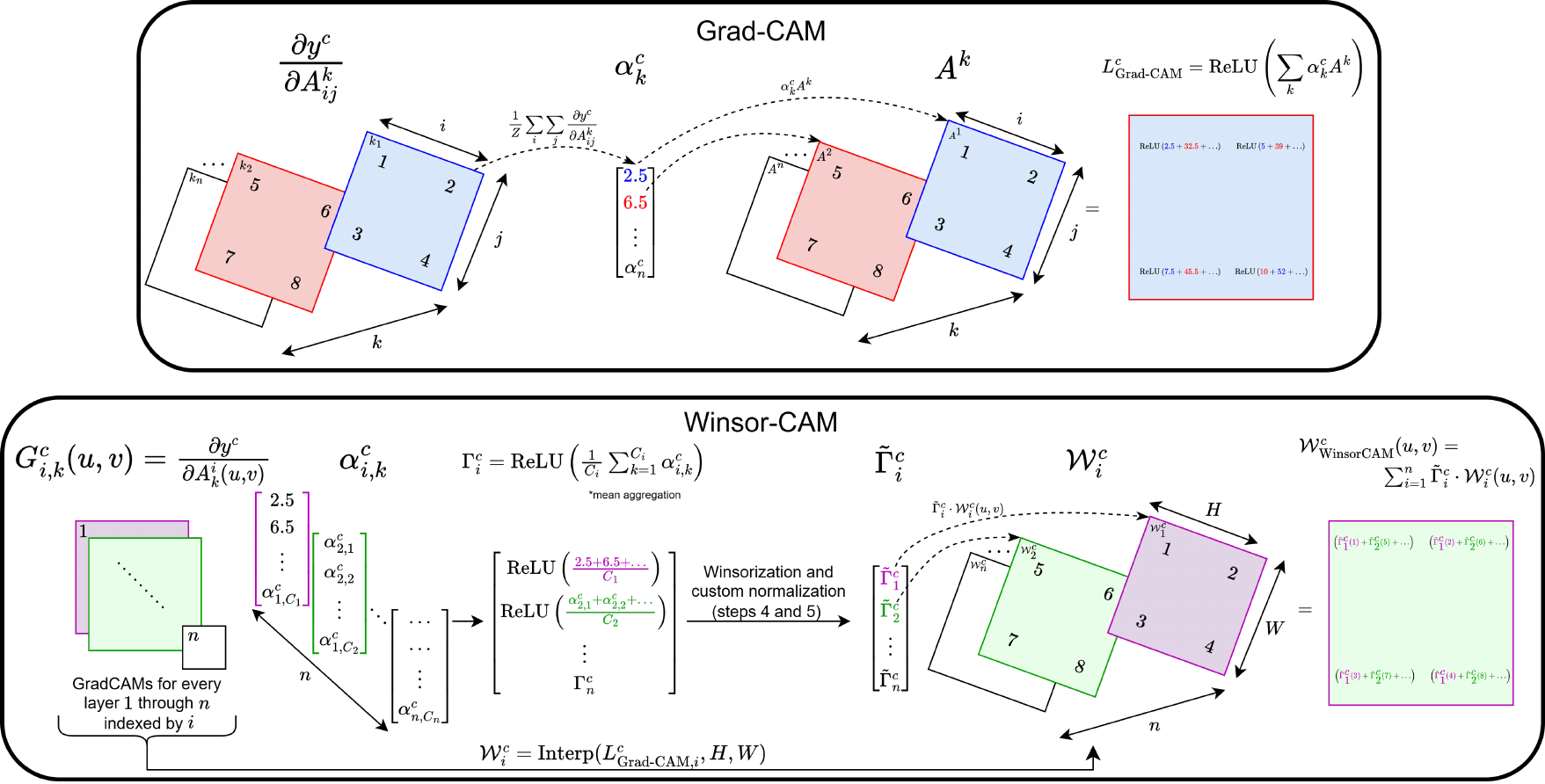}
    \caption{Comparison of Grad-CAM and Winsor-CAM pipelines. \textbf{Top:} Standard Grad-CAM applied to a single convolutional layer. Gradients w.r.t. target class are computed and average-pooled to obtain filter-wise importance weights, which linearly combine activation maps to produce a heatmap at the layer's spatial resolution. \textbf{Bottom:} Winsor-CAM pipeline. Grad-CAM maps are computed for all convolutional layers, and corresponding importance weights yield layer-wise importance scores. These scores are Winsorized to suppress outliers and normalized (both operations account for zero-valued layers, as in Fig.~\ref{fig:Overlay}). Normalized importance scores then weight the interpolated Grad-CAM maps from all layers, producing the final high-resolution heatmap.}
    \label{fig:pipeline}
\end{figure*}

\smallskip
\noindent
\textbf{Compared to uniform aggregation:} FullGrad~\cite{srinivas2019full} sums gradient contributions across all layers, allowing deeper layers (with larger activations due to hierarchical composition) to dominate regardless of discriminative relevance. Winsor-CAM applies percentile-based thresholding (\stepref{4}) to layer importance scores \emph{before} aggregation, bounding outlier contributions while preserving relative ordering (\stepref{6}) without heuristic scaling or manual selection.

The percentile threshold $p$ (Eq.~\ref{eq:winsor_threshold}) controls clipping aggressiveness: lower values emphasize early-layer features (edges, textures), higher values retain deeper-layer abstractions (shapes, categories). Unlike fixed hyperparameters in baselines (e.g., IG's baseline choice), $p$ enables interactive exploration of the feature hierarchy without architectural modification or extra passes, suitable for expert-in-the-loop explanations at varying semantic resolutions.

\section{Methodology}\label{sec:methodology}

This section formally defines Winsor-CAM, our proposed method for interpretable single-pass multi-layer saliency maps. We first recap Grad-CAM (\ref{sec:gradcam_recap}), then describe how Winsor-CAM aggregates feature relevance across all convolutional layers using statistical Winsorization and a user-controllable percentile threshold (\ref{sec:winsor_cam}).

\subsection{Grad-CAM Revisited}\label{sec:gradcam_recap}

Gradient-weighted Class Activation Mapping (Grad-CAM)~\cite{gradcam} visualizes class-specific discriminative regions in CNNs by computing gradients of the target class score with respect to feature maps in, typically, the final convolutional layer. For class $c$, the importance weight for the $k$-th filter is
\[
\alpha_k^c \;=\; \frac{1}{H\cdot W}\sum_{u=1}^{H}\sum_{v=1}^{W}\frac{\partial y^c}{\partial A^k_{uv}},
\]
where $y^c$ is the score for class $c$, $A^k_{uv}$ is the activation of the $k$-th channel at spatial location $(u,v)$, and $H$ and $W$ denote the spatial height and width of $A^k$ (so $A^k\in\mathbb{R}^{H\times W}$). These weights (\(\alpha_k^c\)) are then linearly combined with the feature maps, followed by a ReLU activation:

\[
L^c_{\text{Grad-CAM}} = \text{ReLU} \left( \sum_k \alpha_k^c A^k \right),
\]
where $L^c \in \mathbb{R}^{H\times W}$ denotes the resulting class-specific localization map. The resulting heatmap (\(L^c_{\text{Grad-CAM}}\)) is typically upsampled to the input resolution and normalized for visualization. As Grad-CAM generally focuses only on the final convolutional layer, it potentially misses important low- and mid-level features. Winsor-CAM overcomes this by extending Grad-CAM to all spatial convolutional layers in the network.

\subsection{Winsor-CAM: Layer-Wise Saliency Aggregation via Controlled Importance Scaling}\label{sec:winsor_cam}

Winsor-CAM generalizes Grad-CAM by aggregating class-specific saliency information from all convolutional layers in a CNN, rather than relying solely on the final layer. To ensure stable and interpretable visualizations, it incorporates statistical Winsorization to suppress outlier importance values and balance contributions across the network depth. As provided in Algorithm~\ref{alg:winsorcam}, our method proceeds through six stages: \textit{(1)} layer-wise Grad-CAM computation, \textit{(2)} spatial alignment via interpolation, \textit{(3)} importance score extraction, \textit{(4)} Winsorization-based outlier suppression, \textit{(5)} normalized layer weighting, and \textit{(6)} final saliency fusion. The pipelines for Grad-CAM and Winsor-CAM are visualized in Fig.~\ref{fig:pipeline}.

\step{1}{Grad-CAM Computation Per Layer} Let a CNN have $n$ convolutional layers indexed by $i \in [1, n]$.
For each layer $i$:
\begin{itemize}
    \item $A^i \in \mathbb{R}^{C_i\times H_i\times W_i}$ are the feature maps,
    \item $A^i_k \in \mathbb{R}^{H_i\times W_i}$ is the $k$-th channel,
\item $C_i$ is the number of channels
\item $y^c \in \mathbb{R}$ is the logit for class $c$.
\end{itemize}
The gradient of $y^c$ w.r.t. $A^i_k$ at spatial location $(u, v) \in [1, H_i]\times [1, W_i]$ is given by:
\begin{equation}
    G^c_{i,k}(u, v) = \frac{\partial y^c}{\partial A^i_k(u,v)}.
    \label{eq:grad_map}
\end{equation}

\noindent To compute the importance of channel $k$ at layer $i$ for class $c$, we apply global average pooling over the gradient map:
\begin{equation}
    \alpha^c_{i,k} = \frac{1}{Z_i} \sum_{u=1}^{H_i} \sum_{v=1}^{W_i} G^c_{i,k}(u, v),
    \quad \text{where } Z_i = H_i \cdot W_i.
    \label{eq:gradcam_weight}
\end{equation}
where $A^i\in\mathbb{R}^{C_i\times H_i\times W_i}$ is the output of the $i$-th convolutional layer and $H_i,W_i$ are its spatial height and width. Gradients in early layers can exhibit sign oscillation due to downstream inhibitory/excitatory weights; therefore $\alpha^c_{i,k}$ measures sensitivity of $y^c$ to $A^i_k$ rather than direct correlation, motivating the subsequent ReLU and Winsorization steps.

The per-layer Grad-CAM map is formed as a weighted sum of channel maps followed by ReLU:
\begin{equation}
    L^c_{\text{Grad-CAM},i}(u,v) = \text{ReLU} \left( \sum_{k=1}^{C_i} \alpha^c_{i,k} \cdot A^i_k(u,v) \right).
    \label{eq:gradcam_map}
\end{equation}
where $L^c_{\text{Grad-CAM},i}\in\mathbb{R}^{H_i\times W_i}$ is the nonnegative per-layer class localization map, with larger values indicating stronger positive contribution to $y^c$.

% algorithm - label -> alg:winsorcam
\begin{algorithm}[t]
\scriptsize
\caption{Winsor-CAM: Human-Tunable Visual Explanations from Deep Networks via Layer-Wise Winsorization}
\label{alg:winsorcam}
\algsetup{linenosize=\scriptsize}
\begin{algorithmic}[1]
\REQUIRE Trained CNN model $f$, input image $I$, target class $c$, Winsorization percentile $p$, aggregation method $\in \{\text{mean}, \text{max}\}$
\ENSURE Winsor-CAM heatmap $\mathcal{W}^c_{\text{Winsor-CAM}}$
% \STATE Initialize lists: $\mathcal{W}^c \gets [\,]$, $\Gamma^c \gets [\,]$
\STATE Let $(H, W) \gets (\max_i H_i, \max_i W_i)$ where $(H_i, W_i)$ are spatial dimensions of layer $i$

\FOR{each convolutional layer $i = 1$ to $n$}
    \STATE Compute activations $A^i$ and gradients $G^c_{i,k} = \frac{\partial y^c}{\partial A^i_k}$ \hspace*{\fill} //~\stepref{1}
    \STATE $\alpha^c_{i,k} \gets \frac{1}{H_i \cdot W_i} \sum_{u,v} G^c_{i,k}(u,v)$
    \STATE $L^c_{\text{Grad-CAM},i}(u,v) \gets \text{ReLU}\left(\sum_k \alpha^c_{i,k} \cdot A^i_k\right)$
    \STATE $\mathcal{W}^c_i \gets \text{Interp}(L^c_{\text{Grad-CAM},i}(u,v), H, W)$ \hspace*{\fill} //~\stepref{2}
    \STATE $\Gamma^c_i \gets \text{ReLU}(\text{agg}_k(\alpha^c_{i,k}))$ where $\text{agg} \in \{\text{mean}, \text{max}\}$\hspace*{28pt} //~\stepref{3}
\ENDFOR
\STATE Let $\Gamma^+ \gets [\Gamma^c_i \mid \Gamma^c_i > 0]$\hspace*{\fill} //~\stepref{4}
\IF{$\Gamma^+ = \emptyset$} \RETURN $\mathbf{0}_{H\times W}$ \ENDIF
\STATE Let $T \gets \text{Quantile}(\Gamma^+, p)$ \COMMENT{Obtain upper percentile-based threshold}
\STATE Let $x_{\min} \gets \min(\Gamma^+)$, $x_{\max} \gets \max(\Gamma^+), \epsilon \gets 10^{-6}$ 
\FOR{each layer $i = 1$ to $n$}
    \STATE $\Gamma^{c,\text{win}}_i \gets \min(\Gamma^c_i, T)$ if $\Gamma^c_i > 0$, else $0$
    \STATE $\tilde{\Gamma}^c_i \gets 0.1 + \dfrac{\Gamma^{c,\text{win}}_i - x_{\min}}{\max(x_{\max} - x_{\min}, \epsilon)} \cdot 0.9$ if $\Gamma^{c,\text{win}}_i > 0$, else $0$\hspace*{1pt}//~\stepref{5}
\ENDFOR
\STATE $\mathcal{W}^c_{\text{Winsor-CAM}} \gets \sum_{i=1}^n \tilde{\Gamma}^c_i \cdot \mathcal{W}^c_i$\hspace*{\fill} //~\stepref{6}
\RETURN $\mathcal{W}^c_{\text{Winsor-CAM}}$
\end{algorithmic}

\end{algorithm}

\step{2}{Spatial Alignment via Interpolation} Since $L^c_{\text{Grad-CAM},i}$ from each layer $i$ has spatial dimensions $(H_i, W_i)$, we upsample all maps to a common resolution $(H, W)$:
\begin{equation}
\begin{aligned}
\mathcal{W}^c_i &= \text{Interp}(L^c_{\text{Grad-CAM},i}, H, W),
\end{aligned}
\label{eq:interpolation}
\end{equation}
where $H=\max_i H_i$, $W=\max_i W_i$, and $\mathrm{Interp}(\cdot)$ denotes spatial resampling (e.g., bilinear or nearest-neighbor). Let $\mathcal{W}^c = \{\mathcal{W}^c_i\}_{i=1}^{n}$ denote the set of interpolated maps. We upsample smaller maps to $(H,W)$ so that all $\mathcal{W}^c_i$ can be linearly combined in \stepref{6}.

\begin{figure*}[tbp]
    \centering
    \includegraphics[width=1\textwidth, trim=1.5 .75 1.4 1.5, clip]{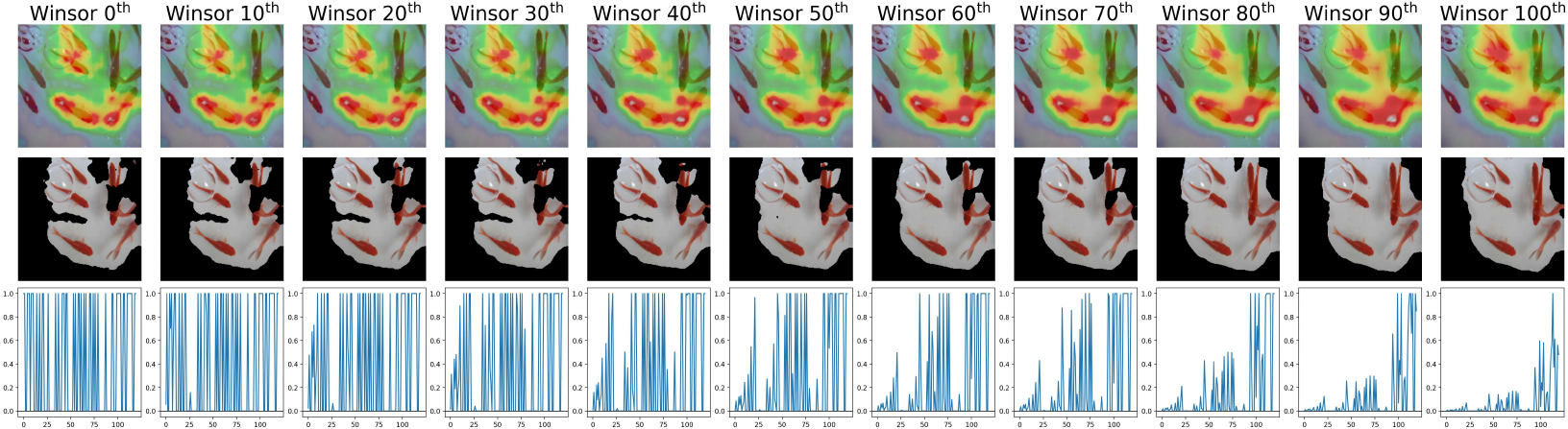}

    \caption{Progression of Winsor-CAM visualizations on DenseNet121 as percentile $p$ varies from 0 to 100 (increments of 10) for class ``goldfish''. Top: raw heatmaps showing saliency shift from fine-grained details to broader patterns as $p$ increases. Middle: binarized masks after thresholding. Bottom: layer-wise importance distributions (x-axis = layer index, early to late; y-axis = normalized importance after Winsorization). Lower $p$-values suppress extreme scores and emphasize early-layer features (textures, edges), while higher $p$-values retain broader contributions from deeper layers (coarser, high-level saliency). This demonstrates Winsor-CAM's semantic-level control over explanation granularity.}

    \label{fig:winsorcam_progession}
\end{figure*}

\step{3}{Per-Layer Importance Score Aggregation} To quantify the overall relevance of each layer $i$ to class $c$, we compute a scalar $\Gamma^c_i$ from its filter-wise weights $\alpha^c_{i,k}$. This can be defined using either of the following: mean or max aggregation. Let $\Gamma^c = \{\Gamma^c_1, \ldots, \Gamma^c_n\}$ be the vector of per-layer importance scores. These scores will guide how strongly each layer's heatmap contributes to the final visualization. Mean aggregation is defined as:

To quantify each layer $i$'s overall relevance to class $c$, we compute a scalar $\Gamma^c_i$ from filter-wise weights $\alpha^c_{i,k}$ using either mean or max aggregation:
\begin{equation}
\Gamma^c_i = \text{ReLU} \left( \frac{1}{C_i} \sum_{k=1}^{C_i} \alpha^c_{i,k} \right)
\label{eq:layer_importance_mean}
.\end{equation}
Similarly, max aggregation can be expressed as,
\begin{equation}
\Gamma^c_i = \text{ReLU} \left( \max_{k \in \{1, \ldots, C_i\}} \alpha^c_{i,k} \right).
\label{eq:layer_importance_max}
\end{equation}

When $\Gamma^c_i = 0$ after ReLU, the layer has no positive influence on the class logit and is excluded from the final heatmap. This occurs when the aggregated importance score, derived from filter-wise gradients $\alpha^c_{i,k}$, is zero or negative, indicating negligible or suppressive contribution for the target class.

In practice, mean aggregation tends to yield lower importance scores for a larger number of layers, reflecting more conservative relevance estimates. In contrast, max aggregation often results in a greater number of layers receiving high importance scores, highlighting the most dominant filter in each layer. The choice between mean and max aggregation depends on the desired trade-off between interpretability and granularity in the resulting heatmap.

\step{4}{Winsorization of Importance Scores} %\label{sec:winsorization_of_importance}
Classical Winsorization replaces values below the lower $p^{\text{th}}$ and above the upper $(100-p)^{\text{th}}$ percentiles with threshold values. We apply \textbf{one-sided upper clipping} at the $p^{\text{th}}$ percentile, computed only over nonzero scores, to suppress dominant layers while preserving inactive ones. This prevents deeper layers with large $\Gamma^c_i$ from disproportionately influencing the final saliency map (as shown in Figs.~\ref{fig:Overlay} and~\ref{fig:winsorcam_progession}).

Let $\Gamma^+ = \{\Gamma^c_i \in \Gamma^c \mid \Gamma^c_i > 0\}$ be the set of nonzero importance scores. The upper Winsorization threshold $T$ is defined as the $p^{\text{th}}$ percentile of $\Gamma^+$:

\begin{equation}
T = \text{Quantile}(\Gamma^+, p).
\label{eq:winsor_threshold}
\end{equation}
\noindent We then apply clipped thresholding to each $\Gamma^c_i$:
\begin{equation}
\Gamma^{c,\text{win}}_i = 
\begin{cases}
\min(\Gamma^c_i, T), & \Gamma^c_i > 0 \\
0, & \text{otherwise}
\end{cases}
\label{eq:winsor_operation}.
\end{equation}
\noindent Let $\Gamma^{c,\text{win}} = \{ \Gamma^{c,\text{win}}_1, \ldots, \Gamma^{c,\text{win}}_n \}$, the set of all Winsorized importance values for class $c$ across all $n$ layers. This step suppresses extreme values using the user-defined percentile $p$, controlling outlier influence while preserving zero values and limiting the effect of large layer-wise importance values that might otherwise produce misleading visualizations. This step operates only on scalar importance scores $\Gamma^c_i$, not spatial maps $\mathcal{W}^c_i$, preserving fine-grained visual structures while regulating layer weighting.

\step{5}{Min-Max Normalization with Zero Preservation} To ensure bounded contributions, we normalize $\Gamma^c_{\text{winsorized}}$ to $[L, H]$ (default $[0.1, 1.0]$): 
\begin{equation}
    \tilde{\Gamma}^c_i =
    \begin{cases}
        L + \frac{(\Gamma^{c,\text{win}}_i - x_{\min})}{\max(x_{\max} - x_{\min}, \epsilon)} (H - L), & \Gamma^{c,\text{win}}_i > 0 \\
        0, & \text{otherwise}
    \end{cases}
    \label{eq:normalized_gamma},
\end{equation}\noindent
where $x_{\min} = \min \Gamma^+$, $x_{\max} = \max \Gamma^+$, and $\epsilon = 10^{-6}$ prevents division by zero. Let $\tilde{\Gamma}^c = \{ \tilde{\Gamma}^c_i\}_{\{i=1, \dots, n\}}$ denote the final per-layer importance weights.

By assigning zero to non-positive importance scores and mapping positive scores to the specified range (excluding zero), this step ensures that only layers with positive contributions influence the final heatmap while preserving the semantic meaning of zero-valued weights. The parameters $L$ and $H$ can be tuned to balance visual contrast and interpretability. The default choice $L = 0.1$ and $H = 1.0$ offers clear separation between active and inactive layers while allowing true zero values to persist.

It is also worth noting that under maximum aggregation, a choice of $p=0$ will typically result in $\tilde {\Gamma }_i^c=L$ for all layers with positive importance. This occurs because all positive scores are clipped to the minimum value $x_{\min}$, and Eq.~\ref{eq:normalized_gamma} maps them to the lower bound L of the normalization range. Consequently, the resulting heatmap becomes equivalent to the naïve mean of all Grad-CAM maps, as the layer weights are uniform across active layers.

\step{6}{Final Heatmap Construction} The final Winsor-CAM heatmap is computed as a weighted linear combination of the interpolated Grad-CAM maps $\mathcal{W}^c_i$ using the normalized importance scores $\tilde{\Gamma}^c_i$:
\begin{equation}
    \mathcal{W}^c_{\text{Winsor-CAM}}(u,v) = \sum_{i=1}^{n} \tilde{\Gamma}^c_i \cdot \mathcal{W}^c_i(u,v)
    \label{eq:winsorcam_final}.
\end{equation}
The result $\mathcal{W}^c_{\text{Winsor-CAM}} \in \mathbb{R}^{H\times W}$ reflects spatially combined saliency from low- to high-level features across the network. For visualization, this map can be normalized to $[0,1]$ and overlaid on the input image.
\begin{equation}
\text{Overlay}(I, \text{Normalize}(\mathcal{W}^c_{\text{Winsor-CAM}})).
\label{eq:winsorcam_overlay}
\end{equation}

The percentile threshold $p$ (\stepref{4}) provides a continuous control knob for adjusting the balance between deep and shallow features, with higher values typically emphasizing deeper layers and lower values emphasizing earlier ones. This pipeline is illustrated in Fig.~\ref{fig:pipeline}, with progression across $p$-values shown in Fig.~\ref{fig:winsorcam_progession}.

\begin{figure}[!tbp]
    \centering
    \includegraphics[width=0.489\textwidth, trim=2.4 .75 1.8 3.3, clip]{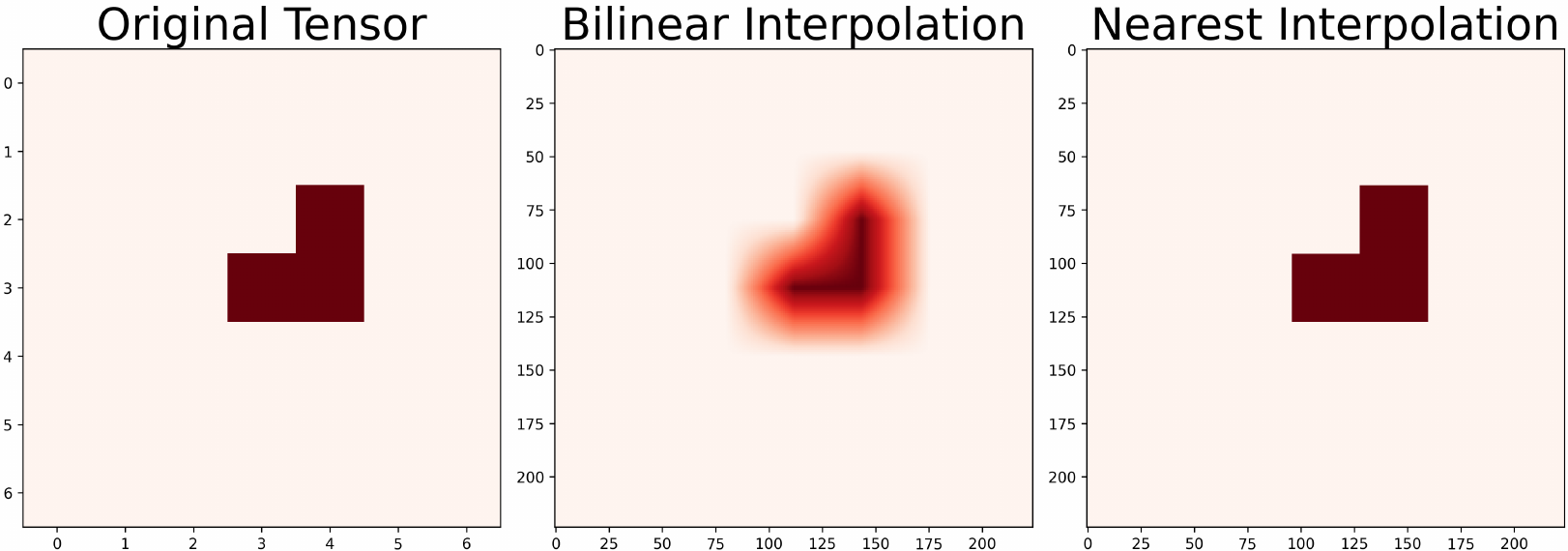}

    \caption{Interpolation's impact on feature map appearance. Starting from $7\times 7$ resolution (left), upsampled to $224\times 224$ using bilinear (center) and nearest-neighbor (right) interpolation. Nearest-neighbor preserves blocky structure; bilinear, bicubic, and other higher order polynomial methods yield visually smoother transitions.}

    \label{fig:interpolation_example}
\end{figure}

\section{Experiments}\label{sec:experiments}

Evaluating explainability methods remains challenging due to the subjective nature of visual explanation quality. Unlike supervised learning metrics (e.g., accuracy, F1), saliency-based methods may highlight features aligned with internal model reasoning but divergent from human intuition~\cite{Samek2017}. For instance, a CNN might associate wolves with forest backgrounds, emphasizing trees rather than the animal itself. Similarly, a digit classifier may fixate on specific geometric properties, such as the three points forming the digit “3,” rather than its holistic shape. Such misalignments underscore the need for both qualitative and quantitative evaluation. Our assessment incorporates visual analysis and localization metrics (IoU, center-of-mass distance, insertion/deletion AUC) to evaluate Winsor-CAM's interpretability and spatial alignment properties.

To evaluate Winsor-CAM, we conduct both qualitative (\ref{sec:qualitative_evaluation}) and quantitative (\ref{sec:quantitative_evaluation}) comparisons against standard Grad-CAM (final layer) and naïve mean aggregation of Grad-CAM maps as described in~\ref{sec:comparative_evaluation_protocol}. The backbone models used in our experiments include ResNet50~\cite{Kaiming2016}, DenseNet121~\cite{Huang2017}, VGG16~\cite{simonyan2015}, InceptionV3~\cite{Szegedy2016}, Efficientnet-B0~\cite{efficientnet2019}, and ConvNeXt-Tiny~\cite{convnext2022}, each pretrained on the ImageNet dataset~\cite{Russakovsky2015}. With experimental setups described in Sections~\ref{sec:qualitative_evaluation} and~\ref{sec:quantitative_evaluation}.

\subsection{Qualitative Evaluation}\label{sec:qualitative_evaluation}

We present qualitative comparisons of Winsor-CAM against standard Grad-CAM (final-layer) and naïve mean-layer aggregation. The Winsorization parameter $p$ (\stepref{4}) enables human-adjustable control over semantic depth: lower $p$-values emphasize shallow features (edges, textures), while higher $p$-values prioritize deeper, abstract representations.

We manually select representative $p \in \{0, 100\}$ values and examine heatmap alignment with semantically meaningful regions, assessing localization of class-discriminative features and suppression of background noise. First, we analyze interpolation's effect on saliency map appearance across different layers and resolutions. Second, we demonstrate how Winsor-CAM visualizes both high- and low-level features interpretably.

\begin{figure}[!tbp]
    \centering
    \includegraphics[width=0.489\textwidth, trim=8.75 9.3 7 7.75, clip]{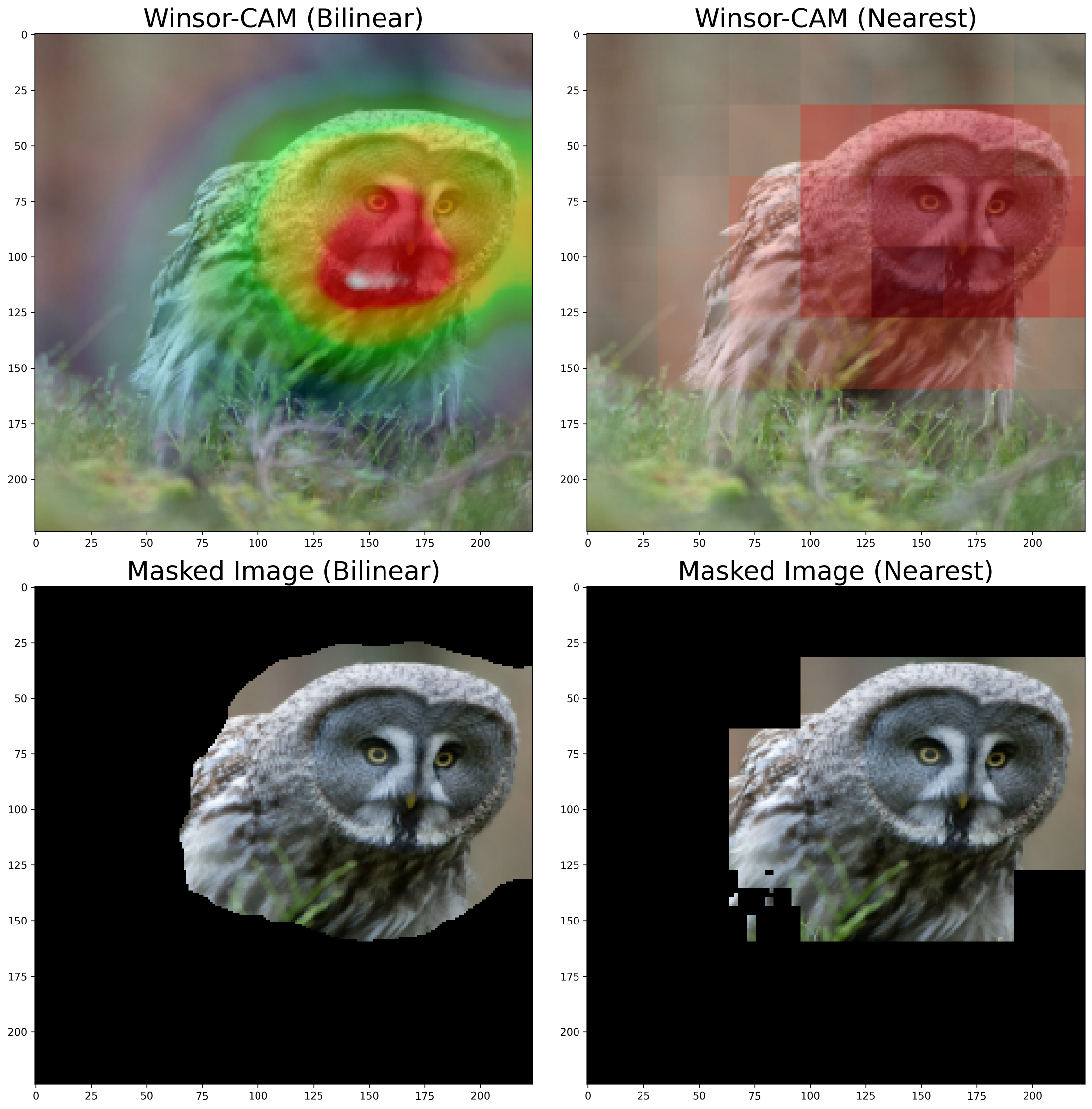}

    \caption{Effect of interpolation method on Winsor-CAM outputs using DenseNet121. Left: bilinear interpolation; right: nearest-neighbor interpolation. Top: continuous heatmaps; bottom: binarized outputs. Differences highlight how interpolation choice influences visual appearance and localization structure. ImageNet validation set, class: ``owl''.}
    \label{fig:Dense121_interpolation}
\end{figure}

\subsubsection{Interpolation's Effect on Saliency Maps}\label{sec:interpolation_effects}
A key factor that influences the visual appearance of saliency maps, regardless of the explanation method used, is the interpolation technique employed during upsampling. Since convolutional feature maps from intermediate layers are typically low-resolution (e.g., $7\times 7$), they must be upsampled to match the input resolution (e.g., $224\times 224$ for ImageNet images) before being visualized as heatmaps.

The choice of interpolation method significantly affects perceptual quality and interpretability. \textit{Nearest-neighbor interpolation} produces coarse, blocky visualizations resembling pixelation, which may obscure finer feature boundaries. In contrast, \textit{bilinear interpolation} and other polynomial-based methods produce smoother transitions and more visually coherent maps, potentially aiding human interpretation. This phenomenon is illustrated in Fig.~\ref{fig:interpolation_example}, which shows interpolation's impact on a ResNet50 heatmap.

Understanding interpolation's visual effect is particularly important when comparing explanations qualitatively, as it can bias human perception of saliency granularity or precision. Therefore, consistent interpolation settings should be maintained across methods when evaluating interpretability.

As described in Eq.~\eqref{eq:interpolation}, Winsor-CAM upscales each intermediate Grad-CAM output to match the spatial resolution of the largest feature map in the network. These upsampled maps are then combined via a weighted linear sum, as defined in Eq.~\eqref{eq:winsorcam_final}. The interpolation method used during this step can significantly affect both the visual quality of the resulting heatmaps and their behavior in subsequent post-processing steps, such as binarization. An example of the impact of interpolation on Winsor-CAM outputs is shown in Fig.~\ref{fig:Dense121_interpolation}.

Visualizations of Winsor-CAM and other methods that aggregate feature maps are typically less impacted by the choice of interpolation method than standard single-layer Grad-CAM, as they aggregate information from layers of different sizes. For instance, Grad-CAM heatmaps in ResNet-50 range from $112\times 112$ to $7\times 7$, corresponding to early and late convolutional layers, respectively. When all layers' feature maps are aggregated lower-level features from the earlier $112\times 112$ tend to smooth the later higher-level $7\times 7$ features. This is demonstrated in  Fig.~\ref{fig:Resnet50_interpolation} where the final layer Grad-CAM is compared with multi-layer Winsor-CAM outputs using both bilinear and nearest-neighbor interpolation.

\begin{figure}[tbp]
    \centering
    \includegraphics[width=0.489\textwidth, trim=.5 .5 .5 3.75, clip]{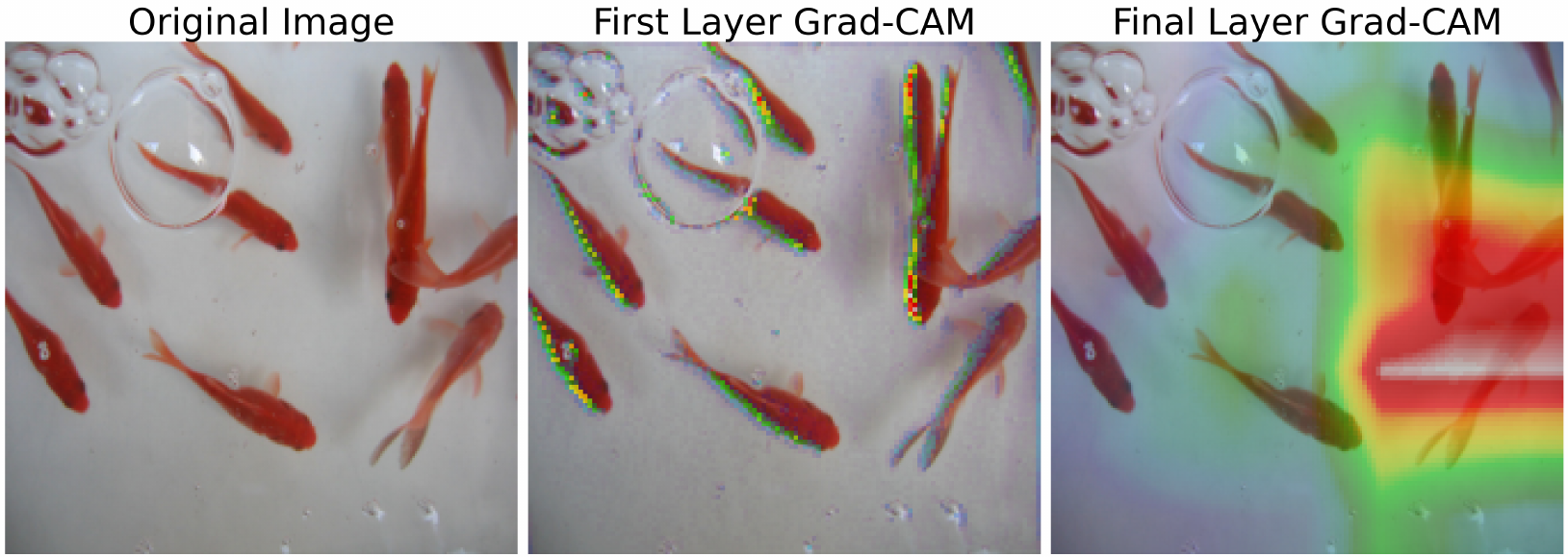}
    \caption{Comparison of Grad-CAM outputs from first and final convolutional layers of DenseNet121 trained on ImageNet explaining class ``goldfish''. First layer highlights lower-level features (edges), while final layer captures higher-level, spatially general regions. Winsor-CAM output for this image is in Fig.~\ref{fig:winsorcam_progession}.}

    \label{fig:first_vs_final}
\end{figure}

Nearest-neighbor interpolation preserves the discrete structure of feature maps and is arguably more faithful to the underlying convolutional architecture, which applies localized grid-aligned filters. However, this produces blocky, pixelated outputs that may hinder human interpretation. In contrast, bilinear, bicubic, and other higher-order interpolation methods (e.g., spline-based interpolation) produce smoother heatmaps with continuous transitions, which are generally more visually appealing and easier to interpret.

This highlights a trade-off between architectural faithfulness and human interpretability. While nearest-neighbor interpolation aligns more closely with the discrete nature of CNN operations, bilinear interpolation offers improved perceptual coherence. Most Grad-CAM implementations, including widely used libraries such as \texttt{pytorch-gradcam}~\cite{jacobgil2021pytorchgradcam}, default to bilinear interpolation for visualization. While we report quantitative results for all three interpolation methods---nearest-neighbor, bilinear, and bicubic---(see Section~\ref{sec:quantitative_evaluation}), we visualize only nearest-neighbor and bilinear in figures, as bicubic produced negligible visual and quantitative differences (see Tables~\ref{tab:model_performance_mean},~\ref{tab:model_performance_max},~\ref{tab:model_performance_mean_polyp}, and~\ref{tab:model_performance_max_polyp}) from bilinear across all evaluated heatmaps.

\subsubsection{Feature Hierarchy Visualization}\label{sec:feature_hierarchy_visualization}

To produce better visualizations, we seek to show both high-level and low-level features in a way that is interpretable to a human. As shown by~\cite{yosinski2015}, the earlier layers within a CNN capture low-level features (such as edges and colors), while later layers capture higher-level features (such as shapes and objects). This phenomenon is shown in Fig.~\ref{fig:first_vs_final}, where the first layer Grad-CAM output of a DenseNet121 trained on Imagenet is shown alongside the last layer Grad-CAM output.

Winsor-CAM aims to capture both low- and high-level features relevant to a model's prediction by aggregating importance scores across all convolutional layers, using either mean or max strategies (Eqs.~\eqref{eq:layer_importance_mean} and~\eqref{eq:layer_importance_max}). The user-controlled parameter $p$ in Eq.~\eqref{eq:winsor_threshold} modulates this aggregation by attenuating extreme layer contributions: lower $p$-values emphasize early-layer, fine-grained lower-level features, while higher values prioritize deeper, more abstract higher representations. Fig.~\ref{fig:winsorcam_progession} illustrates this progression, with Winsor-CAM outputs shown for $p$-values from $0$ to $100$ in steps of $10$. This tunable balance between shallow and deep features introduces a semantic trade-off that users can adjust based on task-specific interpretability needs.

To effectively visualize the output of Winsor-CAM across multiple $p$-values an animated GIF (Graphics Interchange Format) or video would best convey how the heatmap evolves. This dynamic representation would illustrate to a user how the model's focus shifts across the feature hierarchy as the $p$-value changes. This was not implemented in this work, as it is not a common practice in academic papers, and would be difficult to include in a static format. However, this is a possible future work that could be done to better visualize the output of Winsor-CAM. For instance, within the medical field one could create a user interface that allows a medical professional to manually adjust the $p$-value to show different features that the given model is focusing on as, perhaps, the final layer Grad-CAM output may not show all areas of medical significance.

\subsection{Quantitative Evaluation}\label{sec:quantitative_evaluation}

\subsubsection{Evaluation Metrics}\label{sec:evaluation_metrics}
For quantitative evaluation, we adopt the Intersection over Union (IoU) metric to assess the localization performance of the generated saliency maps. IoU measures the overlap between the binarized saliency mask and the ground truth segmentation mask. The score is defined as:
\begin{equation}
    \text{IoU} = \frac{\text{TP}}{\text{TP} + \text{FP} + \text{FN}}
    \label{eq:iou},
\end{equation}
where TP, FP, and FN denote the true positives, false positives, and false negatives, respectively. TP corresponds to correctly identified foreground pixels, FP to incorrectly highlighted background pixels, and FN to missed ground truth regions.

We also measure the Euclidean distance between the Center-of-Mass (CoM) of the saliency map and the ground truth mask to assess spatial alignment. The CoM coordinates $(x_c, y_c)$ are defined as:
{\small
\begin{equation}
\begin{aligned}
x_c = \frac{\sum_{i=0}^{H-1}\sum_{j=0}^{W-1} j \cdot I(i,j)}{\sum_{i=0}^{H-1}\sum_{j=0}^{W-1} I(i,j)} \mbox{ and }
y_c = \frac{\sum_{i=0}^{H-1}\sum_{j=0}^{W-1} i \cdot I(i,j)}{\sum_{i=0}^{H-1}\sum_{j=0}^{W-1} I(i,j)}
\end{aligned}
\label{eq:centroid},
\end{equation}
}
\noindent
where $I(i,j)$ is the normalized intensity at position $(i,j)$, and $H$, $W$ are image dimensions. Normalization is performed as:
\begin{equation}
I(i,j) = \frac{M(i,j) - \min(M)}{\max(M) - \min(M) + \epsilon}
\label{eq:normalization},
\end{equation}
where $M(i,j)$ represents the original intensity values from either a continuous saliency heatmap or binary segmentation mask, and $\epsilon = 10^{-6}$ prevents division by zero. CoM for ground truth masks uses binary segmentation, while saliency map CoM uses raw heatmaps before binarization. The Euclidean distance between CoMs is computed from their $(x_c, y_c)$ coordinates and normalized by the image diagonal length to ensure scale invariance. This metric complements IoU by measuring spatial alignment with the object's true location.

We evaluate explanation faithfulness using insertion and deletion metrics~\cite{petsiuk2018rise}, which quantify the change in model confidence as salient regions are progressively added to or removed from a blurred baseline image. We compute the area under the confidence curve (AUC) as a scalar measure of explanation quality. For computational efficiency and reproducibility, we implement these metrics using batched GPU processing with a fixed per-image Gaussian-blurred baseline (kernel size = 51, $\sigma$ = 10); for PolypGen, we use a black baseline to avoid blur artifacts that could resemble polyps. Insertion and deletion curves are computed over 50 uniformly spaced iterations between 0\% and 100\% of salient pixels, using the heatmap to rank pixel importance.

\subsubsection{Comparative Evaluation Protocol}\label{sec:comparative_evaluation_protocol}

For each test image, we compute the following aforementioned metrics:

\begin{itemize}
    \item \textbf{Winsor-CAM Metrics:} For each percentile value $p \in \{0, 10, 20, \ldots, 100\}$, we compute IoU, CoM distance, insertion AUC, and deletion AUC. We report the highest IoU and insertion AUC, and the lowest CoM distance and deletion AUC across all $p$-values.
    
    \item \textbf{Grad-CAM Metrics:} IoU, CoM distance, insertion AUC, and deletion AUC are computed from the final convolutional layer heatmap.
    
    \item \textbf{Naïve Aggregation Metrics:} Metrics are computed from the heatmap obtained by uniformly averaging Grad-CAM outputs across all spatial convolutional layers.
    
    \item \textbf{Other XAI Methods:} For Grad-CAM++~\cite{chattopadhay2018grad}, LayerCAM~\cite{jiang2021layercam}, ShapleyCAM~\cite{cai2025cams}, ScoreCAM~\cite{wang2020score}, and AblationCAM~\cite{desai2020ablationcam}, metrics are computed from final-layer heatmaps. For FullGrad~\cite{srinivas2019full}, which aggregates gradients across all convolutional layers, metrics are computed from its full-network attribution map.
\end{itemize}

All convolutional layers were used except in EfficientNet-B0, where only depthwise convolutional layers were retained. EfficientNet's architecture includes expansion layers, projection layers, and squeeze-and-excitation blocks. These operations manipulate channel dimensions without spatial convolution, and their inclusion would disproportionately skew Winsor-CAM's percentile-based layer weighting toward non-spatial transformations, diluting spatially informative depthwise convolution contributions.

These configurations are compared to assess the efficacy of Winsor-CAM's tunable parameter $p$ in optimizing localization performance on a per-image basis, alongside comparisons against other XAI methods. For Winsor-CAM, we report the best IoU and insertion AUC (highest) and the best CoM distance and deletion AUC (lowest) across all tested $p$-values for each image. This simulates a human-in-the-loop scenario where users select the most effective $p$-value per image, demonstrating Winsor-CAM's potential when optimally tuned. The objective is to show that Winsor-CAM not only surpasses fixed-output baselines in localization accuracy but also introduces a flexible mechanism for human-guided interpretability, unlike standard Grad-CAM, naïve mean aggregation, and all other mentioned explanation methods, which produce fixed outputs without parameter tuning.

Saliency maps are binarized using Otsu's thresholding method~\cite{Otsu1979}, which maximizes the between-class variance of pixel intensities to separate foreground from background. This binarization strategy simulates a human-tuned thresholding process without requiring manual calibration. In contrast, a fixed threshold (e.g., 0.5) would implicitly assume a sigmoidal relationship in heatmap values, which may not hold in practice and would inadequately capture variability in saliency map structures. The same binarization procedure is applied to all baseline methods.\footnote{We report the mean and standard deviation (image-to-image deviation) across the tested datasets for each method to capture both central tendency and inter-image variability.}

\subsubsection{Dataset and Training}\label{sec:dataset_and_training}
We use the PASCAL VOC 2012 dataset~\cite{pascal-voc-2012}, filtered to single-class images to ensure unambiguous correspondence between saliency maps and target classes. This yields 933 training and 928 test images stratified based on class.

The classification head of each model was modified from 1000 to 20 output classes with dropout added to prevent overfitting. Training configuration: AdamW optimizer, cross-entropy loss, learning rate 0.0001 (0.001 for EfficientNet-B0), OneCycleLR scheduler, 20 epochs, batch size 64, input size $224\times 224$ ($299\times 299$ for InceptionV3). Data augmentations included random horizontal flip (p=0.5), rotation ($\leq$7°), color jitter, random resized crop, affine translation ($\leq$10\%), and Gaussian blur (kernel=3, $\sigma \in [0.1,0.5]$). All images were normalized using standard ImageNet statistics. Training was performed on NVIDIA Tesla V100 GPU (32 GB). Model performance metrics are in Table~\ref{tab:model_metrics_voc}.

For exploratory analysis on medical imaging, we trained the same set of models on PolypGen~\cite{PolypGen2023} using Focal Loss~\cite{focalloss2020} ($\alpha=0.25$, $\gamma=2.0$), a learning rate 0.00001, and no dropout. Training, validation, and test splits (1329, 444, 452 images) were stratified by patient ID and positive/negative sequence ratio. Data augmentations mirrored the PASCAL VOC configuration. Images were not normalized to account for the substantial distributional differences between medical endoscopy images and natural images. Training was performed on the same hardware configuration. Model performance metrics are in Table~\ref{tab:model_metrics_polyp}.

\subsubsection{Evaluation Methodology}\label{sec:evaluation_methodology}

Our evaluation consists of five complementary analyses. First, we compare Winsor-CAM against two internal Grad-CAM baselines: \textit{(1)} the standard final-layer Grad-CAM output, and \textit{(2)} a naïve mean aggregation of Grad-CAM maps across all convolutional layers. The corresponding results are shown in Tables~\ref{tab:model_performance_mean} and~\ref{tab:model_performance_max}.

Second, we analyze the effect of fixing the Winsorization parameter $p$ across all images using DenseNet121 with mean aggregation and bilinear interpolation. For each $p$-value from 0 to 100 (in steps of 10), we compute the average IoU, CoM distance, insertion AUC, and deletion AUC across all test images. This analysis demonstrates how performance varies when $p$ is held constant rather than optimized per image, providing insight into the trade-offs between semantic depth and localization accuracy. Results are presented in Table~\ref{tab:winsor_iou_thresholds}.

Third, we compare Winsor-CAM against several established XAI methods---final layer Grad-CAM, Grad-CAM++, LayerCAM, ShapleyCAM, ScoreCAM, AblationCAM, and FullGrad---under a single representative configuration (DenseNet121, mean aggregation, bilinear interpolation) for reasons described in Section~\ref{sec:observations}. These results are presented in Table~\ref{tab:xai_comparison}.

\setlength{\tabcolsep}{5pt} % Reduces column separation even more
\begin{table}[tbp]
    % \scriptsize
    \tiny
    \centering
    \caption{Training and Validation Metrics for Different Models on PASCAL VOC 2012 Subset}
    \begin{tabular}{lc!{\color{lightgray}\vrule}ccc!{\color{lightgray}\vrule}ccc}
    \hline
    \textbf{Model} & \textbf{Accuracy} & \multicolumn{3}{c!{\color{lightgray}\vrule}}{\textbf{Weighted}} & \multicolumn{3}{c}{\textbf{Macro}} \\
        \cline{3-8}
        & & \textbf{F1} & \textbf{Prec.} & \textbf{Recall} & \textbf{F1} & \textbf{Prec.} & \textbf{Recall} \\
        \hline

        \multicolumn{2}{l!{\color{lightgray}\vrule}}{\textit{Training}} & \multicolumn{3}{c!{\color{lightgray}\vrule}}{} & \multicolumn{3}{c}{} \\
        ResNet50        & 0.9979 & 0.9975 & 0.9980 & 0.9979 & 0.9806 & 0.9947 & 0.9750 \\
        DenseNet121     & 0.9893 & 0.9872 & 0.9853 & 0.9893 & 0.9360 & 0.9349 & 0.9374 \\
        InceptionV3     & 0.9904 & 0.9884 & 0.9869 & 0.9904 & 0.9391 & 0.9347 & 0.9443 \\
        VGG16           & 1.0000 & 1.0000 & 1.0000 & 1.0000 & 1.0000 & 1.0000 & 1.0000 \\
        Efficientnet-B0 & 1.0000 & 1.0000 & 1.0000 & 1.0000 & 1.0000 & 1.0000 & 1.0000 \\
        ConvNeXt-tiny   & 0.9861 & 0.9850 & 0.9864 & 0.9861 & 0.9543 & 0.9853 & 0.9463 \\
        \hline
        \multicolumn{2}{l!{\color{lightgray}\vrule}}{\textit{Validating}} & \multicolumn{3}{c!{\color{lightgray}\vrule}}{} & \multicolumn{3}{c}{} \\
        ResNet50        & 0.8685 & 0.8651 & 0.8637 & 0.8685 & 0.7955 & 0.7925 & 0.8016 \\
        DenseNet121     & 0.8416 & 0.8346 & 0.8361 & 0.8416 & 0.7615 & 0.7729 & 0.7653 \\
        InceptionV3     & 0.9019 & 0.8997 & 0.9001 & 0.9019 & 0.8450 & 0.8439 & 0.8504 \\
        VGG16           & 0.8621 & 0.8582 & 0.8648 & 0.8621 & 0.8042 & 0.8407 & 0.8027 \\
        Efficientnet-B0 & 0.8405 & 0.8383 & 0.8452 & 0.8405 & 0.7882 & 0.8235 & 0.7870 \\
        ConvNeXt-tiny   & 0.8448 & 0.8381 & 0.8458 & 0.8448 & 0.7669 & 0.7892 & 0.7697 \\
    \hline
    \end{tabular}
    \label{tab:model_metrics_voc}
\end{table}

Fourth, we conduct an ablation study to assess the impact of layer selection on Winsor-CAM performance using DenseNet121 with mean aggregation and bilinear interpolation. Various layer selection strategies were evaluated, including using only the final convolutional layer, individual dense blocks (e.g., final block only, final two blocks), and all convolutional layers. This analysis isolates the contribution of different network depths to overall localization performance, with results summarized in Table~\ref{tab:densenet121_ablation_study}.

Fifth, to assess Winsor-CAM's generalizability to medical imaging contexts, we conduct an exploratory analysis on the same set of models as described in Section~\ref{sec:dataset_and_training} trained on the PolypGen polyp segmentation dataset~\cite{PolypGen2023}.

All metrics are computed following the evaluation protocol outlined in Section~\ref{sec:comparative_evaluation_protocol}, which specifies the procedures for binarization, optimal $p$-value selection, and metric calculation. We report the mean and standard deviation across the test dataset for each method to capture both central tendency and variability in performance.

\setlength{\tabcolsep}{5pt} % Reduces column separation even more
\begin{table}[tbp]
    % \scriptsize
    \tiny
    \centering
    \caption{Training and Validation Metrics for Different Models on the PolypGen Dataset}
    \begin{tabular}{lc!{\color{lightgray}\vrule}ccc!{\color{lightgray}\vrule}ccc}
    \hline
    \textbf{Model} & \textbf{Accuracy} & \multicolumn{3}{c!{\color{lightgray}\vrule}}{\textbf{Weighted}} & \multicolumn{3}{c}{\textbf{Macro}} \\
        \cline{3-8}
        & & \textbf{F1} & \textbf{Prec.} & \textbf{Recall} & \textbf{F1} & \textbf{Prec.} & \textbf{Recall} \\
        \hline

        \multicolumn{2}{l!{\color{lightgray}\vrule}}{\textit{Training}} & \multicolumn{3}{c!{\color{lightgray}\vrule}}{} & \multicolumn{3}{c}{} \\
        ResNet50        & 0.9720 & 0.9719 & 0.9720 & 0.9720 & 0.9652 & 0.9707 & 0.9602 \\
        DenseNet121     & 0.9492 & 0.9492 & 0.9491 & 0.9492 & 0.9374 & 0.9385 & 0.9364 \\
        InceptionV3     & 0.8492 & 0.8294 & 0.8742 & 0.8492 & 0.7722 & 0.9105 & 0.7352 \\
        VGG16           & 0.8970 & 0.8935 & 0.8966 & 0.8970 & 0.8653 & 0.8951 & 0.8452 \\
        Efficientnet-B0 & 0.9169 & 0.9172 & 0.9177 & 0.9169 & 0.8986 & 0.8954 & 0.9021 \\
        ConvNeXt-tiny   & 0.9735 & 0.9734 & 0.9734 & 0.9735 & 0.9672 & 0.9710 & 0.9635 \\
        \hline
        \multicolumn{2}{l!{\color{lightgray}\vrule}}{\textit{Validating}} & \multicolumn{3}{c!{\color{lightgray}\vrule}}{} & \multicolumn{3}{c}{} \\
        ResNet50        & 0.9302 & 0.9286 & 0.9274 & 0.9302 & 0.8173 & 0.8330 & 0.8035 \\
        DenseNet121     & 0.9459 & 0.9450 & 0.9443 & 0.9459 & 0.8598 & 0.8737 & 0.8473 \\
        InceptionV3     & 0.8964 & 0.8989 & 0.9020 & 0.8964 & 0.7536 & 0.7421 & 0.7670 \\
        VGG16           & 0.9257 & 0.9240 & 0.9227 & 0.9257 & 0.8055 & 0.8206 & 0.7922 \\
        Efficientnet-B0 & 0.9189 & 0.9189 & 0.9189 & 0.9189 & 0.7972 & 0.7972 & 0.7972 \\
        ConvNeXt-tiny   & 0.9122 & 0.9041 & 0.9020 & 0.9122 & 0.7408 & 0.7993 & 0.7060 \\
        \hline
        \multicolumn{2}{l!{\color{lightgray}\vrule}}{\textit{Testing}} & \multicolumn{3}{c!{\color{lightgray}\vrule}}{} & \multicolumn{3}{c}{} \\
        ResNet50        & 0.8555 & 0.8475 & 0.8447 & 0.8555 & 0.7373 & 0.7698 & 0.7162 \\
        DenseNet121     & 0.8863 & 0.8762 & 0.8800 & 0.8863 & 0.7818 & 0.8507 & 0.7449 \\
        InceptionV3     & 0.8294 & 0.8248 & 0.8215 & 0.8294 & 0.7050 & 0.7173 & 0.6953 \\
        VGG16           & 0.8744 & 0.8559 & 0.8710 & 0.8744 & 0.7371 & 0.8599 & 0.6938 \\
        Efficientnet-B0 & 0.7891 & 0.7875 & 0.7860 & 0.7891 & 0.6483 & 0.6506 & 0.6462 \\
        ConvNeXt-tiny   & 0.8768 & 0.8605 & 0.8720 & 0.8768 & 0.7474 & 0.8556 & 0.7050 \\

    \hline
    \end{tabular}
    % \vspace{-10pt}  % Reduce space AFTER caption
    \label{tab:model_metrics_polyp}
\end{table}

% pascal voc Mean
\setlength{\tabcolsep}{5pt} % Reduces column separation even more
\begin{table*}[!t]
    \centering
    
    \caption{\fontsize{7.25}{10}\selectfont Performance Metrics for Different Interpolation Methods (Mean Aggregation) on Correctly and Incorrectly Classified Images from the PASCAL VOC 2012 Subset.}
    \tiny
    \begin{tabular}{ll!{\color{lightgray}\vrule}ccc!{\color{lightgray}\vrule}ccc!{\color{lightgray}\vrule}ccc!{\color{lightgray}\vrule}ccc}
    \hline
    & & \shortstack[t]{\textbf{Winsor-}\\\textbf{CAM}} & \shortstack[t]{\textbf{Final}\\\textbf{Layer}} & \shortstack[t]{\textbf{Avg.}\\\textbf{Layer}} & 
    \shortstack[t]{\textbf{Winsor-}\\\textbf{CAM}} & \shortstack[t]{\textbf{Final}\\\textbf{Layer}} & \shortstack[t]{\textbf{Avg.}\\\textbf{Layer}} &
    \shortstack[t]{\textbf{Winsor-}\\\textbf{CAM}} & \shortstack[t]{\textbf{Final}\\\textbf{Layer}} & \shortstack[t]{\textbf{Avg.}\\\textbf{Layer}} &
    \shortstack[t]{\textbf{Winsor-}\\\textbf{CAM}} & \shortstack[t]{\textbf{Final}\\\textbf{Layer}} & \shortstack[t]{\textbf{Avg.}\\\textbf{Layer}} \\
    \hline 
    \multicolumn{14}{c}{\textbf{\textit{Correctly Classified Images}}} \\
    \hline
    \textbf{Interp.} & \textbf{Model} & 
    \multicolumn{3}{c!{\color{lightgray}\vrule}}{\textbf{IoU$\bm{\Uparrow}$}} & \multicolumn{3}{c!{\color{lightgray}\vrule}}{\textbf{Center-of-Mass Distance$\bm{\Downarrow}$}} & \multicolumn{3}{c!{\color{lightgray}\vrule}}{\textbf{Insertion$\bm{\Uparrow}$}} & \multicolumn{3}{c}{\textbf{Deletion$\bm{\Downarrow}$}} \\
    \cline{3-14}
        Bilinear & ResNet50               &\textbf{0.386±0.172} &        0.354±0.182 &0.334±0.150&\textbf{0.064±0.047}&0.073±0.044&0.075±0.050&\textbf{0.782±0.189}&        0.760±0.180 &0.752±0.197&\textbf{0.264±0.247}&0.320±0.286&0.311±0.282 \\
                 & DenseNet121            &\textbf{0.468±0.186} &        0.390±0.173 &0.428±0.170&\textbf{0.059±0.045}&0.074±0.051&0.067±0.046&\textbf{0.656±0.217}&        0.623±0.224 &0.648±0.219&\textbf{0.197±0.170}&0.242±0.181&0.207±0.177 \\ 
                 & InceptionV3            &\textbf{0.404±0.195} &        0.374±0.196 &0.370±0.180&\textbf{0.065±0.049}&0.082±0.059&0.076±0.053&\textbf{0.796±0.165}&        0.787±0.190 &0.785±0.177&\textbf{0.314±0.248}&0.355±0.253&0.315±0.254 \\
                 & VGG16                  &\textbf{0.389±0.168} &        0.367±0.155 &0.348±0.157&\textbf{0.067±0.052}&0.074±0.047&0.072±0.052&        0.873±0.176 &\textbf{0.900±0.135}&0.839±0.166&\textbf{0.196±0.199}&0.291±0.251&0.224±0.235 \\
                 & Efficientnet-B0        &        0.391±0.196  &\textbf{0.398±0.217}&0.280±0.169&\textbf{0.069±0.053}&0.074±0.060&0.086±0.059&\textbf{0.816±0.210}&        0.799±0.234 &0.788±0.206&\textbf{0.324±0.237}&0.364±0.227&0.419±0.285 \\
                 & ConvNeXt-tiny          &\textbf{0.350±0.166} &        0.314±0.159 &0.308±0.161&\textbf{0.070±0.054}&0.094±0.059&0.081±0.055&\textbf{0.678±0.236}&        0.658±0.247 &0.663±0.238&\textbf{0.303±0.256}&0.342±0.265&0.327±0.263 \\
        \hline
        Nearest  & ResNet50               &\textbf{0.372±0.164}&        0.347±0.178 &0.319±0.139&\textbf{0.064±0.047}&0.073±0.044&0.075±0.050&        0.766±0.192 &\textbf{0.789±0.169}&0.726±0.202&\textbf{0.256±0.215}&0.352±0.276&0.296±0.257 \\
                 & DenseNet121            &\textbf{0.456±0.176}&        0.379±0.167 &0.416±0.161&\textbf{0.059±0.045}&0.075±0.051&0.068±0.046&\textbf{0.661±0.212}&        0.631±0.224 &0.653±0.214&\textbf{0.190±0.148}&0.249±0.173&0.200±0.157 \\
                 & InceptionV3            &\textbf{0.399±0.191}&        0.369±0.192 &0.360±0.173&\textbf{0.081±0.060}&0.100±0.073&0.093±0.066&\textbf{0.803±0.152}&        0.798±0.181 &0.802±0.159&\textbf{0.325±0.239}&0.371±0.246&0.330±0.249 \\
                 & VGG16                  &\textbf{0.378±0.160}&        0.354±0.150 &0.338±0.151&\textbf{0.067±0.052}&0.074±0.047&0.072±0.052&        0.850±0.193 &\textbf{0.884±0.168}&0.806±0.183&\textbf{0.185±0.202}&0.245±0.219&0.211±0.237 \\
                 & Efficientnet-B0        &        0.373±0.184 &\textbf{0.387±0.210}&0.268±0.154&\textbf{0.069±0.053}&0.074±0.061&0.085±0.059&        0.806±0.221 &\textbf{0.810±0.235}&0.781±0.213&\textbf{0.324±0.232}&0.367±0.231&0.412±0.280 \\
                 & ConvNeXt-Tiny          &\textbf{0.324±0.149}&        0.299±0.155 &0.291±0.145&\textbf{0.071±0.054}&0.094±0.060&0.080±0.056&\textbf{0.673±0.234}&        0.651±0.245 &0.661±0.236&\textbf{0.322±0.254}&0.369±0.259&0.336±0.258 \\
        \hline

       Bicubic   & ResNet50               &\textbf{0.381±0.169}&        0.353±0.181 &0.328±0.144&\textbf{0.064±0.047}&0.072±0.043&0.075±0.050&\textbf{0.781±0.190}&        0.752±0.182 &0.752±0.197&\textbf{0.259±0.239}&0.318±0.287&0.310±0.280 \\
                 & DenseNet121            &\textbf{0.467±0.183}&        0.392±0.171 &0.427±0.167&\textbf{0.059±0.045}&0.070±0.045&0.068±0.046&\textbf{0.657±0.215}&        0.615±0.225 &0.649±0.218&\textbf{0.194±0.167}&0.240±0.188&0.204±0.174 \\
                 & InceptionV3            &\textbf{0.401±0.194}&        0.376±0.195 &0.365±0.177&\textbf{0.081±0.060}&0.099±0.069&0.093±0.066&\textbf{0.786±0.167}&        0.777±0.193 &0.777±0.179&\textbf{0.314±0.248}&0.352±0.253&0.315±0.254 \\
                 & VGG16                  &\textbf{0.382±0.163}&        0.360±0.154 &0.341±0.152&\textbf{0.067±0.052}&0.071±0.045&0.073±0.052&        0.868±0.183 &\textbf{0.885±0.166}&0.831±0.171&\textbf{0.193±0.201}&0.288±0.253&0.223±0.240 \\
                 & Efficientnet-B0        &        0.386±0.192 &\textbf{0.400±0.217}&0.278±0.163&\textbf{0.069±0.053}&0.070±0.054&0.085±0.059&\textbf{0.815±0.212}&        0.794±0.239 &0.793±0.207&\textbf{0.313±0.232}&0.358±0.230&0.412±0.286 \\
                 & ConvNeXt-Tiny          &\textbf{0.335±0.158}&        0.310±0.158 &0.302±0.151&\textbf{0.072±0.055}&0.081±0.050&0.080±0.055&\textbf{0.676±0.237}&        0.655±0.249 &0.664±0.237&\textbf{0.303±0.256}&0.343±0.269&0.323±0.263 \\
    \hline
    \multicolumn{14}{c}{\textbf{\textit{Incorrectly Classified Images}}} \\
    \hline
    \textbf{Interp.} & \textbf{Model} & 
    \multicolumn{3}{c!{\color{lightgray}\vrule}}{\textbf{IoU$\bm{\Downarrow}$}} & \multicolumn{3}{c!{\color{lightgray}\vrule}}{\textbf{Center-of-Mass Distance$\bm{\Uparrow}$}} & \multicolumn{3}{c!{\color{lightgray}\vrule}}{\textbf{Insertion$\bm{\Downarrow}$}} & \multicolumn{3}{c}{\textbf{Deletion$\bm{\Uparrow}$}} \\
    \cline{3-14}
        Bilinear & ResNet50               &0.198±0.164&\textbf{0.168±0.147}&        0.178±0.148 &0.111±0.075&\textbf{0.136±0.086}&0.117±0.073&        0.537±0.260 &        0.516±0.242 &\textbf{0.497±0.251}&0.126±0.175&\textbf{0.160±0.231}&0.156±0.212 \\
                 & DenseNet121            &0.316±0.206&\textbf{0.232±0.196}&        0.277±0.185 &0.091±0.065&\textbf{0.128±0.093}&0.101±0.068&        0.335±0.188 &\textbf{0.289±0.187}&        0.326±0.185 &0.089±0.101&\textbf{0.121±0.110}&0.094±0.095 \\
                 & InceptionV3            &0.209±0.163&\textbf{0.157±0.162}&        0.176±0.150 &0.112±0.077&\textbf{0.171±0.114}&0.127±0.082&        0.526±0.283 &\textbf{0.426±0.271}&        0.496±0.286 &0.123±0.115&\textbf{0.204±0.167}&0.143±0.137 \\
                 & VGG16                  &0.181±0.133&        0.194±0.155 &\textbf{0.173±0.124}&0.110±0.072&\textbf{0.121±0.082}&0.109±0.070&\textbf{0.599±0.280}&        0.672±0.263 &        0.602±0.248 &0.091±0.129&\textbf{0.139±0.189}&0.096±0.144 \\
                 & Efficientnet-B0        &0.250±0.179&\textbf{0.101±0.161}&        0.176±0.145 &0.097±0.067&\textbf{0.152±0.105}&0.114±0.076&        0.592±0.261 &\textbf{0.405±0.311}&        0.554±0.277 &0.152±0.171&\textbf{0.352±0.216}&0.208±0.216 \\
                 & ConvNeXt-Tiny          &0.255±0.165&         0.225±0.162&\textbf{0.206±0.146}&0.100±0.068&\textbf{0.117±0.064}&0.111±0.070&        0.383±0.203 &\textbf{0.343±0.198}&        0.352±0.192 &0.128±0.145&\textbf{0.150±0.164}&0.145±0.154 \\
        \hline
        Nearest  & ResNet50               &0.190±0.152&\textbf{0.167±0.148}&        0.172±0.137 &0.111±0.075&\textbf{0.137±0.087}&0.116±0.073&0.509±0.259&        0.557±0.228 &\textbf{0.473±0.248}&0.128±0.174&\textbf{0.164±0.200}&0.153±0.214 \\
                 & DenseNet121            &0.306±0.196&\textbf{0.224±0.193}&        0.269±0.177 &0.092±0.065&\textbf{0.128±0.094}&0.102±0.068&0.355±0.189&\textbf{0.293±0.188}&        0.343±0.187 &0.090±0.074&\textbf{0.126±0.088}&0.096±0.071 \\
                 & InceptionV3            &0.201±0.154&\textbf{0.155±0.158}&        0.172±0.143 &0.137±0.094&\textbf{0.210±0.140}&0.154±0.100&0.538±0.271&\textbf{0.448±0.278}&        0.511±0.278& 0.130±0.112&\textbf{0.211±0.162}&0.148±0.128 \\
                 & VGG16                  &0.175±0.126&        0.190±0.152 &\textbf{0.168±0.118}&0.110±0.072&\textbf{0.121±0.082}&0.109±0.070&0.554±0.294&        0.628±0.313 &\textbf{0.548±0.256}&0.095±0.139&\textbf{0.153±0.219}&0.093±0.144 \\
                 & Efficientnet-B0        &0.237±0.163&\textbf{0.100±0.152}&        0.170±0.134 &0.097±0.068&\textbf{0.152±0.105}&0.113±0.075&0.577±0.266&\textbf{0.426±0.324}&        0.547±0.281 &0.163±0.179&\textbf{0.351±0.222}&0.211±0.211 \\
                 & ConvNeXt-Tiny          &0.236±0.148&        0.216±0.155 &\textbf{0.198±0.135}&0.101±0.068&\textbf{0.117±0.064}&0.110±0.070&0.377±0.200&\textbf{0.333±0.194}&        0.357±0.195 &0.135±0.137&\textbf{0.166±0.158}&0.149±0.146 \\
        \hline

       Bicubic   & ResNet50               &0.196±0.159&\textbf{0.167±0.146}&        0.176±0.143 &        0.111±0.075 &\textbf{0.133±0.084}&0.116±0.073&        0.537±0.261 &        0.509±0.244 &\textbf{0.496±0.250}&0.123±0.170&\textbf{0.160±0.233}&0.156±0.214 \\
                 & DenseNet121            &0.314±0.201&\textbf{0.233±0.197}&        0.278±0.184 &        0.092±0.065 &\textbf{0.110±0.077}&0.101±0.068&        0.341±0.190 &\textbf{0.280±0.190}&        0.332±0.186 &0.087±0.096&\textbf{0.120±0.121}&0.093±0.087 \\
                 & InceptionV3            &0.205±0.160&\textbf{0.157±0.161}&        0.175±0.146 &        0.137±0.094 &\textbf{0.199±0.128}&0.155±0.101&        0.518±0.285 &\textbf{0.425±0.270}&        0.491±0.286 &0.123±0.117&\textbf{0.198±0.154}&0.143±0.141 \\
                 & VGG16                  &0.179±0.129&        0.190±0.153 &\textbf{0.170±0.120}&\textbf{0.109±0.072}&\textbf{0.109±0.074}&0.108±0.070&\textbf{0.587±0.288}&        0.611±0.311 &        0.594±0.245 &0.092±0.138&\textbf{0.135±0.186}&0.097±0.159 \\
                 & Efficientnet-B0        &0.243±0.171&\textbf{0.099±0.159}&        0.174±0.139 &        0.098±0.068 &\textbf{0.136±0.097}&0.112±0.075&        0.592±0.264 &\textbf{0.394±0.307}&        0.558±0.278 &0.148±0.173&\textbf{0.342±0.209}&0.199±0.212 \\
                 & ConvNeXt-Tiny          &0.243±0.152&        0.223±0.160 &\textbf{0.203±0.139}&        0.101±0.068 &\textbf{0.108±0.060}&0.110±0.070&        0.388±0.202 &\textbf{0.337±0.201}&        0.354±0.193 &0.119±0.133&\textbf{0.151±0.167}&0.145±0.154 \\
    \hline
    \end{tabular}
    % \vspace{-10pt}  % Reduce space AFTER caption
    \label{tab:model_performance_mean}
\end{table*}

% pascal voc Max
\setlength{\tabcolsep}{5pt} % Reduces column separation even more
\begin{table*}[!t]
    \centering
    \caption{\fontsize{7.25}{10}\selectfont Performance Metrics for Different Interpolation Methods (Max Aggregation) on Correctly and Incorrectly Classified Images from PASCAL VOC 2012 Subset}
    \tiny
    \begin{tabular}{ll!{\color{lightgray}\vrule}ccc!{\color{lightgray}\vrule}ccc!{\color{lightgray}\vrule}ccc!{\color{lightgray}\vrule}ccc}
    \hline
    & & \shortstack[t]{\textbf{Winsor-}\\\textbf{CAM}} & \shortstack[t]{\textbf{Final}\\\textbf{Layer}} & \shortstack[t]{\textbf{Avg.}\\\textbf{Layer}} & 
    \shortstack[t]{\textbf{Winsor-}\\\textbf{CAM}} & \shortstack[t]{\textbf{Final}\\\textbf{Layer}} & \shortstack[t]{\textbf{Avg.}\\\textbf{Layer}} &
    \shortstack[t]{\textbf{Winsor-}\\\textbf{CAM}} & \shortstack[t]{\textbf{Final}\\\textbf{Layer}} & \shortstack[t]{\textbf{Avg.}\\\textbf{Layer}} &
    \shortstack[t]{\textbf{Winsor-}\\\textbf{CAM}} & \shortstack[t]{\textbf{Final}\\\textbf{Layer}} & \shortstack[t]{\textbf{Avg.}\\\textbf{Layer}} \\
    \hline 
    \multicolumn{14}{c}{\textbf{\textit{Correctly Classified Images}}} \\
    \hline
    \textbf{Interp.} & \textbf{Model} & 
    \multicolumn{3}{c!{\color{lightgray}\vrule}}{\textbf{IoU$\bm{\Uparrow}$}} & \multicolumn{3}{c!{\color{lightgray}\vrule}}{\textbf{Center-of-Mass Distance$\bm{\Downarrow}$}} & \multicolumn{3}{c!{\color{lightgray}\vrule}}{\textbf{Insertion$\bm{\Uparrow}$}} & \multicolumn{3}{c}{\textbf{Deletion$\bm{\Downarrow}$}} \\
    \cline{3-14}
        Bilinear & ResNet50              &\textbf{0.377±0.158}&        0.354±0.182 &0.334±0.150&\textbf{0.067±0.047}&        0.073±0.044 &0.075±0.050&\textbf{0.791±0.185}&0.760±0.180&0.752±0.197&\textbf{0.268±0.269}&0.320±0.286&0.311±0.282 \\
                 & DenseNet121           &\textbf{0.453±0.175}&        0.390±0.173 &0.428±0.170&\textbf{0.067±0.050}&        0.082±0.059 &0.076±0.053&\textbf{0.664±0.217}&0.623±0.224&0.648±0.219&\textbf{0.196±0.170}&0.242±0.181&0.207±0.177 \\ 
                 & InceptionV3           &\textbf{0.396±0.188}&        0.374±0.196 &0.370±0.180&\textbf{0.067±0.050}&        0.082±0.059 &0.076±0.053&\textbf{0.813±0.166}&0.787±0.190&0.785±0.177&\textbf{0.295±0.245}&0.355±0.253&0.315±0.254 \\
                 & VGG16                 &\textbf{0.404±0.163}&        0.367±0.155 &0.348±0.157&\textbf{0.063±0.048}&        0.074±0.047 &0.072±0.052&\textbf{0.905±0.130}&0.900±0.135&0.839±0.166&\textbf{0.185±0.190}&0.291±0.251&0.224±0.235 \\
                 & Efficientnet-B0       &        0.360±0.175 &\textbf{0.398±0.217}&0.280±0.169&        0.077±0.056 &\textbf{0.074±0.060}&0.086±0.059&\textbf{0.850±0.162}&0.799±0.234&0.788±0.206&\textbf{0.330±0.257}&0.364±0.227&0.419±0.285 \\
                 & ConvNeXt-Tiny         &\textbf{0.354±0.163}&        0.314±0.159 &0.308±0.161&\textbf{0.074±0.054}&        0.094±0.059 &0.081±0.055&\textbf{0.691±0.231}&0.658±0.247&0.663±0.238&\textbf{0.292±0.255}&0.342±0.265&0.327±0.263 \\
        \hline
        Nearest  & ResNet50              &\textbf{0.363±0.145}&        0.347±0.178 &0.319±0.139&\textbf{0.067±0.047}&        0.073±0.044 & 0.075±0.050&\textbf{0.793±0.183}&        0.789±0.169 &0.726±0.202&\textbf{0.262±0.238}&0.352±0.276&0.296±0.257 \\
                 & DenseNet121           &\textbf{0.447±0.167}&        0.379±0.167 &0.416±0.161&\textbf{0.060±0.044}&        0.075±0.051 & 0.068±0.046&\textbf{0.671±0.212}&        0.631±0.224 &0.653±0.214&\textbf{0.188±0.148}&0.249±0.173&0.200±0.157 \\
                 & InceptionV3           &\textbf{0.389±0.183}&        0.369±0.192 &0.360±0.173&\textbf{0.067±0.050}&        0.082±0.059 & 0.076±0.054&\textbf{0.823±0.152}&        0.798±0.181 &0.802±0.159&\textbf{0.306±0.238}&0.371±0.246&0.330±0.249 \\
                 & VGG16                 &\textbf{0.392±0.155}&        0.354±0.150 &0.338±0.151&\textbf{0.063±0.048}&        0.074±0.047 & 0.072±0.052&        0.877±0.157 &\textbf{0.884±0.168}&0.806±0.183&\textbf{0.172±0.194}&0.245±0.219&0.211±0.237 \\
                 & Efficientnet-B0       &        0.344±0.162 &\textbf{0.387±0.210}&0.268±0.154&        0.077±0.056 &\textbf{0.074±0.061}& 0.085±0.059&\textbf{0.844±0.175}&        0.810±0.235 &0.781±0.213&\textbf{0.332±0.258}&0.367±0.231&0.412±0.280 \\
                 & ConvNeXt-Tiny         &\textbf{0.336±0.147}&        0.299±0.155 &0.291±0.145&\textbf{0.074±0.054}&        0.094±0.060 & 0.080±0.056&\textbf{0.691±0.229}&        0.651±0.245 &0.661±0.236&\textbf{0.306±0.248}&0.369±0.259&0.336±0.258 \\
        \hline

       Bicubic   & ResNet50               &\textbf{0.372±0.153}&        0.353±0.181 &0.328±0.144&\textbf{0.067±0.047}&        0.072±0.043 &0.075±0.050&\textbf{0.792±0.184}&0.752±0.182&0.752±0.197&\textbf{0.263±0.265}&0.318±0.287&0.310±0.280 \\
                 & DenseNet121            &\textbf{0.453±0.173}&        0.392±0.171 &0.427±0.167&\textbf{0.060±0.044}&        0.070±0.045 &0.068±0.046&\textbf{0.665±0.216}&0.615±0.225&0.649±0.218&\textbf{0.191±0.166}&0.240±0.188&0.204±0.174 \\
                 & InceptionV3            &\textbf{0.392±0.186}&        0.376±0.195 &0.365±0.177&\textbf{0.067±0.050}&        0.081±0.057 &0.076±0.054&\textbf{0.804±0.169}&0.777±0.193&0.777±0.179&\textbf{0.294±0.245}&0.352±0.253&0.315±0.254 \\
                 & VGG16                  &\textbf{0.396±0.158}&        0.360±0.154 &0.341±0.152&\textbf{0.064±0.048}&        0.071±0.045 &0.073±0.052&\textbf{0.899±0.138}&0.885±0.166&0.831±0.171&\textbf{0.181±0.192}&0.288±0.253&0.223±0.240 \\
                 & Efficientnet-B0        &        0.359±0.171 &\textbf{0.400±0.217}&0.278±0.163&        0.076±0.056 &\textbf{0.070±0.054}&0.085±0.059&\textbf{0.852±0.162}&0.794±0.239&0.793±0.207&\textbf{0.320±0.255}&0.358±0.230&0.412±0.286 \\
                 & ConvNeXt-Tiny          &\textbf{0.347±0.154}&        0.310±0.158 &0.302±0.151&\textbf{0.074±0.054}&        0.081±0.050 &0.080±0.055&\textbf{0.691±0.231}&0.655±0.249&0.664±0.237&\textbf{0.288±0.253}&0.343±0.269&0.323±0.263 \\
    \hline
    \multicolumn{14}{c}{\textbf{\textit{Incorrectly Classified Images}}} \\
    \hline
    \textbf{Interp.} & \textbf{Model} & 
    \multicolumn{3}{c!{\color{lightgray}\vrule}}{\textbf{IoU$\bm{\Downarrow}$}} & \multicolumn{3}{c!{\color{lightgray}\vrule}}{\textbf{Center-of-Mass Distance$\bm{\Uparrow}$}} & \multicolumn{3}{c!{\color{lightgray}\vrule}}{\textbf{Insertion$\bm{\Downarrow}$}} & \multicolumn{3}{c}{\textbf{Deletion$\bm{\Uparrow}$}} \\
    \cline{3-14}
        Bilinear & ResNet50              &0.200±0.161&\textbf{0.168±0.147}&        0.178±0.148 &0.112±0.073&\textbf{0.136±0.086}&0.117±0.073&0.548±0.254&        0.516±0.242 &\textbf{0.497±0.251}&0.130±0.198&\textbf{0.160±0.231}&0.156±0.212 \\
                 & DenseNet121           &0.300±0.197&\textbf{0.232±0.196}&        0.277±0.185 &0.094±0.067&\textbf{0.128±0.093}&0.101±0.068&0.341±0.192&\textbf{0.289±0.187}&        0.326±0.185 &0.088±0.093&\textbf{0.121±0.110}&0.094±0.095 \\
                 & InceptionV3           &0.192±0.157&\textbf{0.157±0.162}&        0.176±0.150 &0.121±0.081&\textbf{0.171±0.114}&0.127±0.082&0.532±0.280&\textbf{0.426±0.271}&        0.496±0.286 &0.125±0.124&\textbf{0.204±0.167}&0.143±0.137 \\
                 & VGG16                 &0.216±0.147&        0.194±0.155 &\textbf{0.173±0.124}&0.100±0.071&\textbf{0.121±0.082}&0.109±0.070&0.701±0.242&        0.672±0.263 &\textbf{0.602±0.248}&0.074±0.117&\textbf{0.139±0.189}&0.096±0.144 \\
                 & Efficientnet-B0       &0.237±0.168&\textbf{0.101±0.161}&        0.176±0.145 &0.100±0.071&\textbf{0.152±0.105}&0.114±0.076&0.656±0.241&\textbf{0.405±0.311}&        0.554±0.277 &0.139±0.167&\textbf{0.352±0.216}&0.208±0.216 \\
                 & ConvNeXt-Tiny         &0.240±0.159&        0.225±0.162 &\textbf{0.206±0.146}&0.103±0.068&\textbf{0.117±0.064}&0.111±0.070&0.377±0.194&\textbf{0.343±0.198}&        0.352±0.192 &0.123±0.148&\textbf{0.150±0.164}&0.145±0.154 \\
        \hline
        Nearest  & ResNet50              &0.194±0.150&\textbf{0.167±0.148}&        0.172±0.137 &0.111±0.073&\textbf{0.137±0.087}&0.116±0.073&0.552±0.242&        0.557±0.228 &\textbf{0.473±0.248}&0.129±0.188&\textbf{0.164±0.200}&0.153±0.214 \\
                 & DenseNet121           &0.295±0.188&\textbf{0.224±0.193}&        0.269±0.177 &0.095±0.067&\textbf{0.128±0.094}&0.102±0.068&0.359±0.190&\textbf{0.293±0.188}&        0.343±0.187 &0.088±0.064&\textbf{0.126±0.088}&0.096±0.071 \\
                 & InceptionV3           &0.188±0.150&\textbf{0.156±0.159}&        0.172±0.143 &0.120±0.081&\textbf{0.172±0.114}&0.126±0.082&0.540±0.269&\textbf{0.447±0.279}&        0.512±0.278 &0.125±0.118&\textbf{0.211±0.163}&0.148±0.128 \\
                 & VGG16                 &0.210±0.140&        0.190±0.152 &\textbf{0.168±0.118}&0.100±0.071&\textbf{0.121±0.082}&0.109±0.070&0.659±0.263&        0.628±0.313 &\textbf{0.548±0.256}&0.072±0.126&\textbf{0.153±0.219}&0.093±0.144 \\
                 & Efficientnet-B0       &0.227±0.154&\textbf{0.100±0.152}&        0.170±0.134 &0.101±0.071&\textbf{0.152±0.105}&0.113±0.075&0.645±0.250&\textbf{0.426±0.324}&        0.547±0.281 &0.145±0.180&\textbf{0.351±0.222}&0.211±0.211 \\
                 & ConvNeXt-Tiny         &0.230±0.148&        0.216±0.155 &\textbf{0.198±0.135}&0.103±0.068&\textbf{0.117±0.064}&0.110±0.070&0.384±0.197&\textbf{0.333±0.194}&        0.357±0.195 &0.130±0.138&\textbf{0.166±0.158}&0.149±0.146 \\
        \hline

       Bicubic   & ResNet50               &0.197±0.156&\textbf{0.167±0.146}&        0.176±0.143 &0.111±0.073&\textbf{0.133±0.084}&0.116±0.073&0.553±0.254&        0.509±0.244 &\textbf{0.496±0.250}&0.126±0.192&\textbf{0.160±0.233}&0.156±0.214 \\
                 & DenseNet121            &0.300±0.195&\textbf{0.233±0.197}&        0.278±0.184 &0.095±0.067&\textbf{0.110±0.077}&0.101±0.068&0.347±0.193&\textbf{0.280±0.190}&        0.332±0.186 &0.085±0.086&\textbf{0.120±0.121}&0.093±0.087 \\
                 & InceptionV3            &0.190±0.154&\textbf{0.157±0.161}&        0.175±0.146 &0.120±0.081&\textbf{0.163±0.104}&0.126±0.082&0.526±0.277&\textbf{0.425±0.270}&        0.491±0.286 &0.123±0.124&\textbf{0.198±0.154}&0.143±0.141 \\
                 & VGG16                  &0.211±0.143&        0.190±0.153 &\textbf{0.170±0.120}&0.101±0.070&\textbf{0.109±0.074}&0.108±0.070&0.694±0.241&        0.611±0.311 &\textbf{0.594±0.245}&0.075±0.131&\textbf{0.135±0.186}&0.097±0.159 \\
                 & Efficientnet-B0       &0.232±0.160&\textbf{0.099±0.159}&        0.174±0.139 &0.101±0.071&\textbf{0.136±0.097}&0.112±0.075&0.658±0.243&\textbf{0.394±0.307}&        0.558±0.278 &0.133±0.164&\textbf{0.342±0.209}&0.199±0.212 \\
                 & ConvNeXt-Tiny         &0.235±0.151&        0.223±0.160 &\textbf{0.203±0.139}&0.103±0.068&\textbf{0.108±0.060}&0.110±0.070&0.381±0.196&\textbf{0.337±0.201}&        0.354±0.193 &0.124±0.149&\textbf{0.151±0.167}&0.145±0.154 \\
    \hline
    \end{tabular}
    \label{tab:model_performance_max}
\end{table*}

\subsubsection{Observations}\label{sec:observations}

\textbf{Comparison with Internal Grad-CAM Baselines.}
Tables~\ref{tab:model_performance_mean} and~\ref{tab:model_performance_max} show that Winsor-CAM consistently produces higher IoU values and lower CoM distances compared to both final-layer Grad-CAM and naïve mean aggregation across nearly all models and configurations. DenseNet121 was chosen based on superior performance. An exception is EfficientNet-B0, where final-layer Grad-CAM slightly outperforms Winsor-CAM in IoU and CoM distance, likely due to its architecture and the layer filtering described in Section~\ref{sec:comparative_evaluation_protocol}.

Winsor-CAM also achieves superior deletion AUC across all models and higher insertion AUC in most cases. Bilinear and bicubic interpolation methods outperform nearest-neighbor interpolation, and mean aggregation surpasses max aggregation for IoU and CoM distance, though max aggregation shows advantages for insertion/deletion AUC.

\begin{figure*}[tbp]
    \centering

    \includegraphics[width=.95\textwidth,keepaspectratio, trim=13 7.1 12.5 24.25, clip]{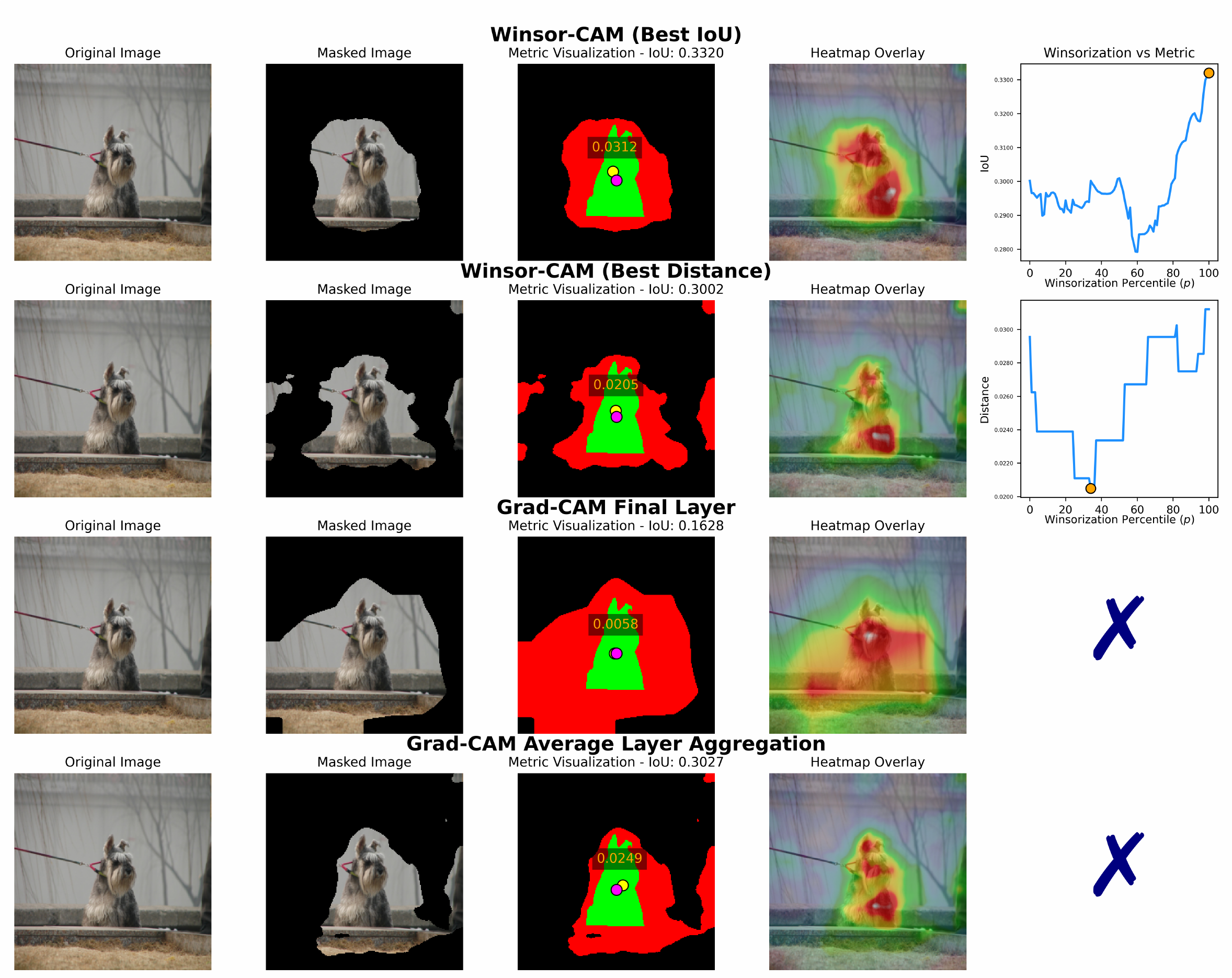}
    \caption{Comparison of Winsor-CAM and Grad-CAM on PASCAL VOC 2012 image (DenseNet121, mean aggregation). Rows: (1) Winsor-CAM best IoU, (2) Winsor-CAM best CoM distance, (3) final-layer Grad-CAM, (4) naïve mean aggregation. IoU: green = TP, red = FP, blue = FN, black = background. CoM: magenta = ground truth center, yellow = heatmap center, orange line = distance. ``Winsorization vs. Metric'' plots show Winsor-CAM performance across $p$-values (omitted for fixed Grad-CAM output). Target class: ``dog''. See Section~\ref{sec:dataset_and_training}.}
    \label{fig:winsorcam_ious}
\end{figure*}

\begin{figure*}[tbp]
    \centering
    \includegraphics[width=.95\textwidth,keepaspectratio, trim=13 7.1 12.5 24.25, clip]{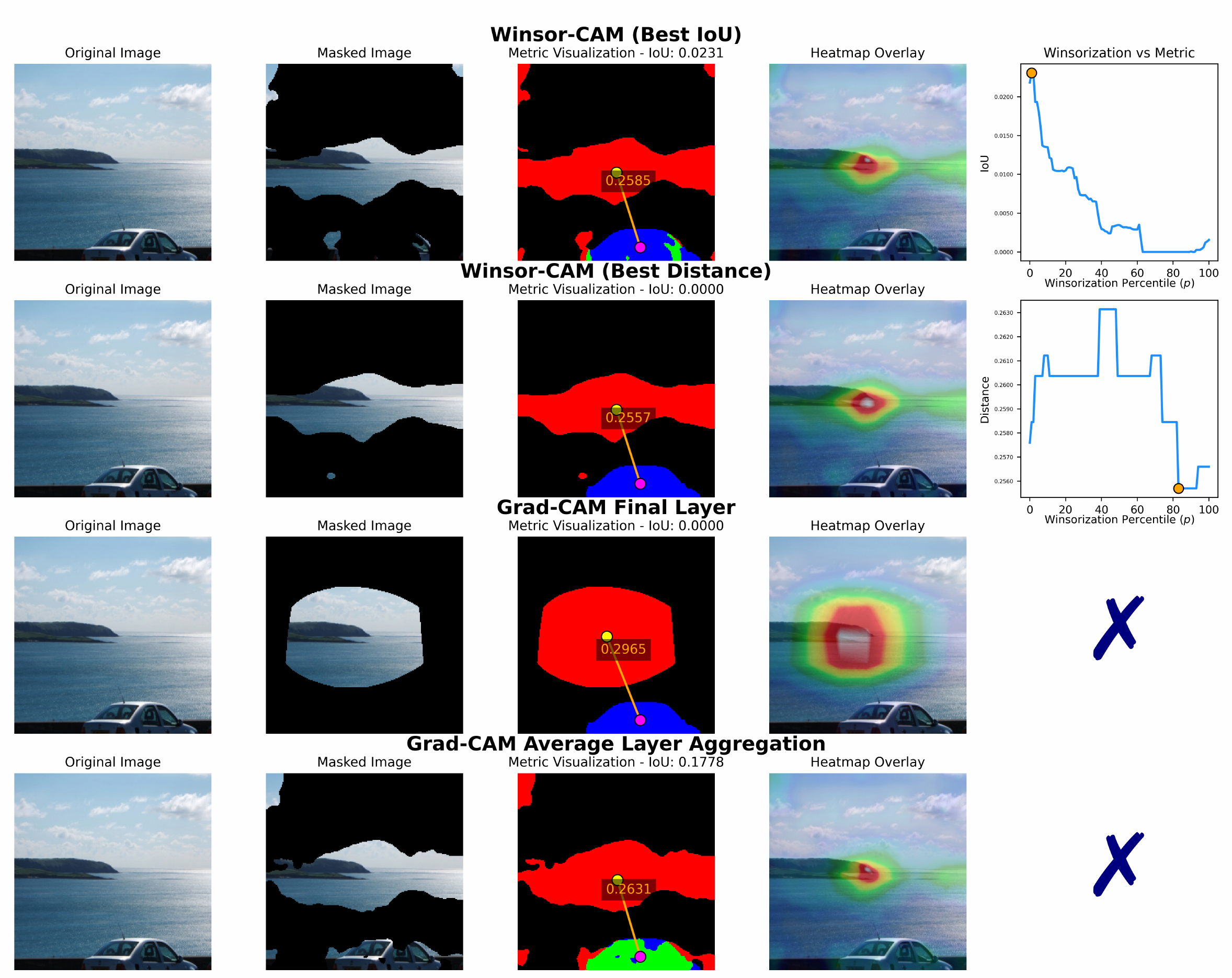}
    \caption{Comparative analysis using the format of Fig.~\ref{fig:winsorcam_ious} for a misclassified image, where the model predicted ``boat'' instead of the ground truth ``car.'' This example illustrates how the localization metrics and heatmaps differ in failure cases.}
    \label{fig:winsorcam_ious_incor}
\end{figure*}

% pascal_voc winsor p thresholds
\setlength{\tabcolsep}{7pt} % Reduces column separation even more
\begin{table}[!t]
    \tiny
    \centering
    \caption{\fontsize{7.25}{5}\selectfont Mean IoU, CoM Distance, Insertion, and Deletion at Different Winsor-CAM Percentile Thresholds ($p$) for DenseNet121 (Correct Predictions) on the PASCAL VOC 2012 Subset.}
    \begin{tabular}{c!{\color{lightgray}\vrule}c!{\color{lightgray}\vrule}c!{\color{lightgray}\vrule}c!{\color{lightgray}\vrule}c}
    \hline
    \textbf{$p$} & \textbf{IoU$\bm{\Uparrow}$} & \textbf{CoM Distance$\bm{\Downarrow}$} & \textbf{Insertion$\bm{\Uparrow}$} & \textbf{Deletion$\bm{\Downarrow}$} \\
    \hline
    0   &         0.442±0.182 &         0.066±0.047 & \textbf{0.643±0.220}& \textbf{0.208±0.177} \\
    10  &         0.441±0.183 &         0.066±0.047 &         0.642±0.220 &         0.209±0.178 \\
    20  &         0.441±0.183 &         0.066±0.047 &         0.641±0.220 &         0.209±0.179 \\
    30  &         0.442±0.183 &         0.066±0.047 &         0.641±0.220 &         0.209±0.179 \\
    40  &         0.443±0.183 &         0.065±0.047 &         0.640±0.219 &         0.210±0.180 \\
    50  & \textbf{0.445±0.184}&         0.065±0.046 &         0.640±0.219 &         0.211±0.179 \\
    60  & \textbf{0.445±0.184}& \textbf{0.065±0.046}&         0.639±0.218 &         0.211±0.180 \\
    70  & \textbf{0.445±0.184}& \textbf{0.064±0.046}&         0.638±0.218 &         0.213±0.180 \\
    80  & \textbf{0.445±0.185}& \textbf{0.064±0.046}&         0.637±0.217 &         0.214±0.180 \\
    90  & \textbf{0.445±0.185}& \textbf{0.064±0.045}&         0.636±0.217 &         0.215±0.180 \\
    100 & \textbf{0.443±0.185}& \textbf{0.064±0.045}&         0.637±0.217 &         0.215±0.180 \\
    \hline
    \end{tabular}
    \label{tab:winsor_iou_thresholds}
\end{table}

\begin{figure}[tbp]
    \centering
    \includegraphics[width=0.48\textwidth, trim=12 13.1 10.5 11.5, clip]{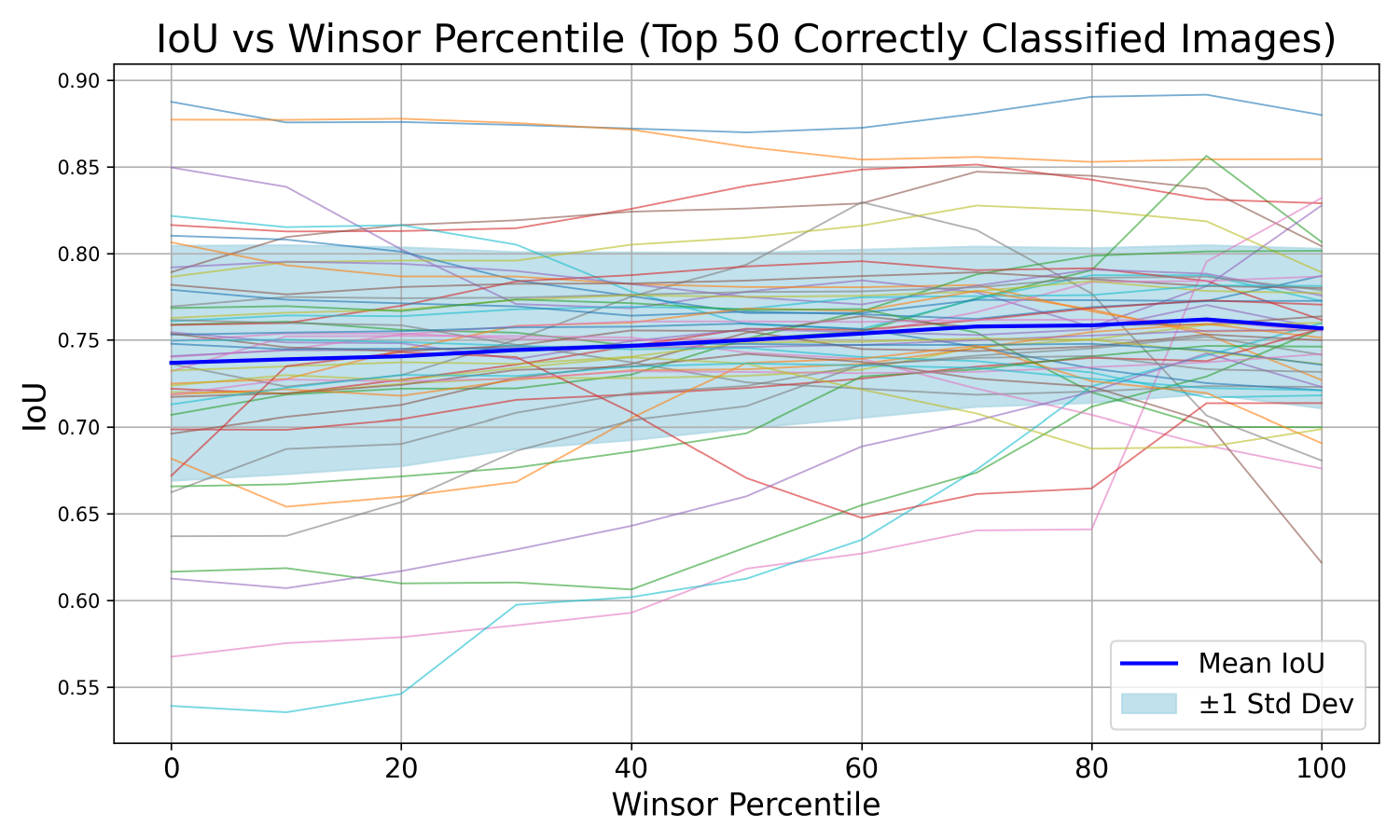}
    \caption{Mean and standard deviation of IoU across the top 50 correctly classified images (ranked by IoU) for DenseNet121 trained on ImageNet, shown for Winsor-CAM with $p$-values from 0 to 100 in increments of 10 using mean layer aggregation. The bold blue line shows mean IoU with light blue shading indicating ±1 standard deviation. Individual IoU values for each image and $p$-value are plotted in multiple colors to show large individual image deviations across the feature hierarchy.}
    \label{fig:mean_and_indiv}
\end{figure}

A key finding is that final-layer Grad-CAM yields more accurate CoMs than naïve mean aggregation, but underperforms in IoU. Winsor-CAM achieves both higher IoU and lower CoM distances than both baselines, while also improving deletion AUC and generally achieving higher insertion AUC. This suggests that final-layer Grad-CAM captures coarse localization but misses finer details from earlier layers, while naïve mean aggregation dilutes meaningful patterns with noise. Winsor-CAM addresses this trade-off through weighted contributions and outlier suppression, yielding improved spatial alignment.

Based on these findings, DenseNet121 with mean aggregation and bilinear interpolation was selected as the representative configuration for all subsequent analyses (fixed $p$-value analysis, and ablation study), achieving the best localization performance among all tested architectures: 0.468±0.186 IoU, 0.059±0.045 CoM distance, with competitive insertion AUC (0.656±0.217) and deletion AUC (0.197±0.170).

\smallskip
\noindent \textbf{Per-Image Variability and Semantic Tunability.}
We split results by prediction correctness because correctly classified cases are our primary interest. For incorrect predictions, IoU scores are generally lower as the model attends to wrong-class features or fails to localize the target object. Despite this, Winsor-CAM yields higher IoU and lower CoM distances than both baselines, even on misclassified examples, likely due to the evaluation protocol (Section~\ref{sec:comparative_evaluation_protocol}) selecting optimal $p$-values per image.

Figs.~\ref{fig:winsorcam_ious} and~\ref{fig:winsorcam_ious_incor} illustrate these effects for correctly and incorrectly classified images, showing IoU and CoM distance overlays with plots of metric variation across $p$-values. In Fig.~\ref{fig:winsorcam_ious}, Winsor-CAM with best IoU for class ``dog'' binarizes cleanly using Otsu's method. While naïve mean aggregation appears aligned, noise exists at image corners (noise) and edges that Winsor-CAM suppresses. The final layer aligns with the dog's center but highlights much of the background. Winsor-CAM (Best Distance) shows that minimizing CoM distance with low $p$ emphasizes earlier layers, introducing some noise.
The variability in optimal $p$-values across images (Fig.~\ref{fig:mean_and_indiv}) highlights Winsor-CAM's utility in human-in-the-loop settings, where experts adjust semantic depth interactively. Selecting the best $p$-value per image simulates an oracle, demonstrating Winsor-CAM's upper-bound performance when optimally tuned. Unlike standard Grad-CAM or fixed aggregation methods offering single static outputs, Winsor-CAM reveals a spectrum of explanations adaptable to user needs. This flexibility is valuable where stakeholders have different interpretive priorities, a radiologist might emphasize high-level semantic features (high $p$) to understand anatomical focus, while a model developer might prefer low-level features (low $p$) to understand edge and texture sensitivities. These findings underscore the need for explanation methods that are quantitatively robust and contextually adaptable.

\smallskip
\noindent \textbf{Fixed $\mathbf{p}$-Value Analysis: Trade-offs in Parameter Selection.}
Table~\ref{tab:winsor_iou_thresholds} demonstrates how Winsor-CAM performance varies when percentile parameter $p$ is fixed across all images rather than optimized per image, revealing important trade-offs. IoU benefits from moderate Winsorization ($p=50$ to $70$), peaking at 0.445±0.184, while CoM distance improves with less aggressive Winsorization ($p=90$ to $100$), reaching 0.064±0.045. Very low $p$-values ($0$ to $20$) produce better insertion AUC (0.643±0.220) and deletion AUC (0.208±0.177), suggesting that emphasizing earlier layers helps capture fine-grained features relevant to the predicted class.

This reveals a fundamental characteristic of Winsor-CAM: different $p$-values emphasize different model behaviors. Lower $p$ suppresses extreme layer contributions and emphasizes early-layer features (textures, edges), while higher $p$ retains broader contributions from deeper layers (coarser, high-level saliency). Fig.~\ref{fig:winsorcam_progession} illustrates how heatmaps and layer-wise importance evolve as $p$ varies from 0 to 100. When $p$ is fixed, the choice represents a trade-off between spatial precision (lower $p$) versus semantic coherence (higher $p$). However, Fig.~\ref{fig:mean_and_indiv} shows that individual images often have substantially different optimal $p$-values, motivating the human-in-the-loop approach where users adjust $p$ based on interpretive needs.

% pascal_voc xai compare
\setlength{\tabcolsep}{4pt} % Reduces column separation even more
\begin{table}[!t]
    \tiny
    \centering
    \caption{\fontsize{7.25}{5}\selectfont Comparison of Winsor-CAM and other XAI methods on correctly classified images (DenseNet121, mean aggregation, bilinear interpolation) on the PASCAL VOC 2012 Subset}
    \begin{tabular}{l!{\color{lightgray}\vrule}cccc}
        \hline
        \textbf{Method} & \textbf{IoU$\bm{\Uparrow}$} & \textbf{CoM Distance$\bm{\Downarrow}$} & \textbf{Insertion$\bm{\Uparrow}$} & \textbf{Deletion$\bm{\Downarrow}$} \\
        \hline
                Winsor-CAM    & \textbf{0.468±0.186} & \textbf{0.059±0.045} & \textbf{0.656±0.217} & \textbf{0.197±0.170} \\
                Grad-CAM      &         0.390±0.173 &          0.074±0.051 &          0.623±0.224 &          0.242±0.181 \\
                Grad-CAM++    &         0.382±0.180 &          0.071±0.045 &          0.612±0.223 &          0.255±0.198 \\
                LayerCAM      &         0.392±0.175 &          0.069±0.044 &          0.615±0.224 &          0.247±0.189 \\
                ShapleyCAM    &         0.390±0.173 &          0.074±0.051 &          0.623±0.224 &          0.242±0.181 \\
                ScoreCAM      &         0.305±0.193 &          0.092±0.068 &          0.575±0.226 &          0.305±0.245 \\
                AblationCAM   &         0.334±0.211 &          0.101±0.091 &          0.594±0.259 &          0.275±0.177 \\
                FullGrad      &         0.433±0.187 &          0.072±0.050 &          0.588±0.221 &          0.251±0.188 \\
        \hline
    \end{tabular}
    % \vspace{-15pt}  % Reduce space AFTER caption
    \label{tab:xai_comparison}
\end{table}

% densenet121 ablation study Pascal_voc
\setlength{\tabcolsep}{4pt} % Reduces column separation even more
\begin{table*}[!t]
    \centering
    \caption{\fontsize{7.5}{10}\selectfont Ablation Study: Layer Selection Impact on DenseNet121 Performance (Mean Aggregation, Bilinear Interpolation) from the PASCAL VOC 2012 Subset}
    \tiny
    \begin{tabular}{ll!{\color{lightgray}\vrule}ccc!{\color{lightgray}\vrule}ccc!{\color{lightgray}\vrule}ccc!{\color{lightgray}\vrule}ccc}
        \hline
        & & \shortstack[t]{\textbf{Winsor-}\\\textbf{CAM}} & \shortstack[t]{\textbf{Final}\\\textbf{Layer}} & \shortstack[t]{\textbf{Avg.}\\\textbf{Layer}} & 
        \shortstack[t]{\textbf{Winsor-}\\\textbf{CAM}} & \shortstack[t]{\textbf{Final}\\\textbf{Layer}} & \shortstack[t]{\textbf{Avg.}\\\textbf{Layer}} &
        \shortstack[t]{\textbf{Winsor-}\\\textbf{CAM}} & \shortstack[t]{\textbf{Final}\\\textbf{Layer}} & \shortstack[t]{\textbf{Avg.}\\\textbf{Layer}} &
        \shortstack[t]{\textbf{Winsor-}\\\textbf{CAM}} & \shortstack[t]{\textbf{Final}\\\textbf{Layer}} & \shortstack[t]{\textbf{Avg.}\\\textbf{Layer}} \\
        \hline
        \textbf{Classification} & \textbf{Configuration} & 
    \multicolumn{3}{c!{\color{lightgray}\vrule}}{\textbf{IoU$\bm{\Uparrow}$}} & \multicolumn{3}{c!{\color{lightgray}\vrule}}{\textbf{Center-of-Mass Distance$\bm{\Downarrow}$}} & \multicolumn{3}{c!{\color{lightgray}\vrule}}{\textbf{Insertion$\bm{\Uparrow}$}} & \multicolumn{3}{c}{\textbf{Deletion$\bm{\Downarrow}$}} \\
        \cline{3-14} 
        \multirow{5}{*}{\textbf{Correct}} 
    & Final Conv                & --- &       \textbf{0.390±0.173}       & ---    & --- &        \textbf{0.074±0.051} & --- & --- &                    \textbf{0.623±0.224} & --- & --- &                   \textbf{0.242±0.181} & --- \\
    & Final dense block         &         0.450±0.190  & --- &         0.425±0.178 &         0.060±0.043  & --- &         0.066±0.043  &         0.641±0.215  & --- &         0.640±0.216  &         0.215±0.175  & --- &         0.222±0.182 \\
    & Final two dense blocks    &         0.465±0.185  & --- &         0.419±0.171 & \textbf{0.059±0.045} & --- &         0.068±0.045  &         0.645±0.218  & --- &         0.642±0.221  &         0.206±0.177  & --- &         0.216±0.188 \\
    & Final three dense blocks  &         0.467±0.186  & --- &         0.424±0.171 & \textbf{0.059±0.045} & --- &         0.068±0.046  &         0.650±0.217  & --- &         0.646±0.219  &         0.200±0.174  & --- &         0.210±0.184 \\
    & All Layers                & \textbf{0.468±0.186} & --- & \textbf{0.428±0.170}& \textbf{0.059±0.045} & --- & \textbf{0.067±0.046} & \textbf{0.656±0.217} & --- & \textbf{0.648±0.219} & \textbf{0.197±0.170} & --- & \textbf{0.207±0.177} \\
    \hline
    \textbf{Classification} & \textbf{Configuration} & 
    \multicolumn{3}{c!{\color{lightgray}\vrule}}{\textbf{IoU$\bm{\Downarrow}$}} & \multicolumn{3}{c!{\color{lightgray}\vrule}}{\textbf{Center-of-Mass Distance$\bm{\Uparrow}$}} & \multicolumn{3}{c!{\color{lightgray}\vrule}}{\textbf{Insertion$\bm{\Downarrow}$}} & \multicolumn{3}{c}{\textbf{Deletion$\bm{\Uparrow}$}} \\
    \cline{3-14}
    \multirow{5}{*}{\textbf{Incorrect}} 
    & Final Conv              &---&      \textbf{0.232±0.196} & --- &  ---&                  \textbf{0.128±0.093} &--- &    --- &                \textbf{0.289±0.187} & --- & --- &                 \textbf{0.121±0.110} & - \\
    & Final dense block       & \textbf{0.310±0.210}& --- &         0.278±0.194 & \textbf{0.091±0.066} & --- & \textbf{0.102±0.070}& \textbf{0.320±0.192}& --- &         0.315±0.192 & \textbf{0.099±0.112}& --- & \textbf{0.102±0.111} \\
    & Final two dense blocks  &         0.317±0.209 & --- & \textbf{0.271±0.186}& \textbf{0.091±0.065} & --- &         0.101±0.068 &         0.319±0.189 & --- & \textbf{0.311±0.191}&         0.093±0.109 & --- &         0.098±0.109 \\
    & Final three dense blocks&         0.316±0.206 & --- &         0.276±0.185 & \textbf{0.091±0.065} & --- &         0.101±0.068 &         0.328±0.189 & --- &         0.320±0.187 &         0.090±0.105 & --- &         0.095±0.103 \\
    & All Layers              &         0.316±0.206 & --- &         0.277±0.185 & \textbf{0.091±0.065} & --- &         0.101±0.068 &         0.335±0.188 & --- &         0.326±0.185 &         0.089±0.101 & --- &         0.094±0.095 \\
    \hline
\end{tabular}
\label{tab:densenet121_ablation_study}
\end{table*}

\smallskip
\noindent \textbf{Comparison with Established XAI Methods.} 
Table~\ref{tab:xai_comparison} shows that Winsor-CAM outperforms Grad-CAM, Grad-CAM++, LayerCAM, ScoreCAM, AblationCAM, ShapleyCAM, and FullGrad under the representative configuration. Notably, Winsor-CAM achieves substantial improvements over FullGrad, which also aggregates information across all layers. With optimal per-image $p$-value selection, Winsor-CAM achieves 0.468±0.186 IoU versus FullGrad's 0.433±0.187, 0.059±0.045 CoM distance versus 0.072±0.050, 0.656±0.217 insertion AUC versus 0.588±0.221, and 0.197±0.170 deletion AUC versus 0.251±0.188. Furthermore, as shown in Table~\ref{tab:winsor_iou_thresholds}, even the worst-performing fixed $p$-value configuration outperforms FullGrad in every metric, demonstrating that Winsorization-based outlier suppression provides consistent and measurable benefits over uniform aggregation approaches, regardless of the specific $p$-value chosen.

\smallskip
\noindent \textbf{Effect of Layer Selection: Ablation Study.}
Table~\ref{tab:densenet121_ablation_study} demonstrates that including more layers generally improves Winsor-CAM's localization performance. Using only the final dense block yields 0.450±0.190 IoU, while using all layers achieves 0.468±0.186. Similarly, CoM distance improves from 0.060±0.043 to 0.059±0.045, insertion AUC from 0.641±0.215 to 0.656±0.217, and deletion AUC from 0.215±0.175 to 0.197±0.170. These results confirm that incorporating earlier layers enhances spatial alignment when outlier contributions are properly suppressed through Winsorization.

The marginal differences between using the final three dense blocks (0.467±0.186 IoU, 0.059±0.045 CoM) versus all layers (0.468±0.186 IoU, 0.059±0.045 CoM) suggest that most discriminative information is concentrated in deeper stages, though early-layer features still contribute to localization and fidelity. The progression from final block only to all layers shows consistent improvement, with substantial gains when moving from single-block to multi-block configurations. This supports the hypothesis that multi-layer aggregation is beneficial, but also indicates diminishing returns as more shallow layers are included, consistent with the understanding that deeper layers capture more semantically relevant features for classification tasks.

% polypgen Mean
\setlength{\tabcolsep}{5pt} % Reduces column separation even more
\begin{table*}[!t]
    \centering
    \caption{\fontsize{7.75}{10}\selectfont Performance Metrics for Different Interpolation Methods (Mean Aggregation) on Correctly and Incorrectly Classified Images from PolypGen}
    % \fontsize{5}{6}\selectfont  % Even smaller 
    \tiny
    \begin{tabular}{ll!{\color{lightgray}\vrule}ccc!{\color{lightgray}\vrule}ccc!{\color{lightgray}\vrule}ccc!{\color{lightgray}\vrule}ccc}
    \hline
    & & \shortstack[t]{\textbf{Winsor-}\\\textbf{CAM}} & \shortstack[t]{\textbf{Final}\\\textbf{Layer}} & \shortstack[t]{\textbf{Avg.}\\\textbf{Layer}} & 
    \shortstack[t]{\textbf{Winsor-}\\\textbf{CAM}} & \shortstack[t]{\textbf{Final}\\\textbf{Layer}} & \shortstack[t]{\textbf{Avg.}\\\textbf{Layer}} &
    \shortstack[t]{\textbf{Winsor-}\\\textbf{CAM}} & \shortstack[t]{\textbf{Final}\\\textbf{Layer}} & \shortstack[t]{\textbf{Avg.}\\\textbf{Layer}} &
    \shortstack[t]{\textbf{Winsor-}\\\textbf{CAM}} & \shortstack[t]{\textbf{Final}\\\textbf{Layer}} & \shortstack[t]{\textbf{Avg.}\\\textbf{Layer}} \\
    \hline 
    \multicolumn{14}{c}{\textbf{\textit{Correctly Classified Images}}} \\
    \hline
    \textbf{Interp.} & \textbf{Model} & 
    \multicolumn{3}{c!{\color{lightgray}\vrule}}{\textbf{IoU$\bm{\Uparrow}$}} & \multicolumn{3}{c!{\color{lightgray}\vrule}}{\textbf{Center-of-Mass Distance$\bm{\Downarrow}$}} & \multicolumn{3}{c!{\color{lightgray}\vrule}}{\textbf{Insertion$\bm{\Uparrow}$}} & \multicolumn{3}{c}{\textbf{Deletion$\bm{\Downarrow}$}} \\
    \cline{3-14}
        Bilinear & ResNet50               &\textbf{0.293±0.154}&        0.081±0.136 &0.266±0.138&\textbf{0.131±0.088}&        0.157±0.090 &0.135±0.088&\textbf{0.928±0.041}&        0.734±0.124 &0.918±0.048&        0.664±0.143 &\textbf{0.360±0.168}&0.786±0.097 \\
                 & DenseNet121            &\textbf{0.367±0.187}&        0.313±0.170 &0.265±0.139&\textbf{0.114±0.089}&        0.126±0.091 &0.128±0.090&\textbf{0.815±0.085}&        0.768±0.106 &0.793±0.081&\textbf{0.433±0.089}&        0.461±0.120 &0.530±0.085 \\
                 & InceptionV3            &\textbf{0.339±0.183}&        0.318±0.194 &0.221±0.130&\textbf{0.126±0.094}&        0.129±0.088 &0.139±0.095&\textbf{0.626±0.108}&        0.567±0.138 &0.592±0.094&        0.374±0.070 &\textbf{0.270±0.075}&0.442±0.079 \\
                 & VGG16                  &        0.264±0.131 &\textbf{0.265±0.141}&0.216±0.105&\textbf{0.119±0.076}&        0.122±0.072 &0.129±0.080&\textbf{0.776±0.077}&        0.610±0.105 &0.771±0.078&        0.432±0.132 &\textbf{0.349±0.112}&0.536±0.113 \\
                 & Efficientnet-B0        &        0.241±0.146 &\textbf{0.342±0.168}&0.179±0.128&        0.135±0.093 &\textbf{0.119±0.079}&0.151±0.093&        0.756±0.113 &\textbf{0.808±0.108}&0.708±0.124&        0.579±0.102 &\textbf{0.530±0.105}&0.627±0.098 \\
                 & ConvNeXt-Tiny          &\textbf{0.239±0.149}&        0.110±0.112 &0.171±0.122&\textbf{0.123±0.092}&        0.167±0.093 &0.143±0.087&\textbf{0.877±0.083}&        0.778±0.116 &0.839±0.099&\textbf{0.604±0.136}&        0.612±0.175 &0.706±0.127 \\
        \hline
        Nearest  & ResNet50               &\textbf{0.278±0.143}&        0.067±0.115 &0.253±0.130&\textbf{0.131±0.088}&        0.158±0.090 &0.136±0.088&\textbf{0.934±0.028}&        0.703±0.117 &0.930±0.031&        0.750±0.096 &\textbf{0.381±0.160}&0.843±0.063 \\
                 & DenseNet121            &\textbf{0.362±0.182}&        0.302±0.166 &0.257±0.133&\textbf{0.115±0.089}&        0.126±0.091 &0.129±0.090&\textbf{0.809±0.075}&        0.790±0.088 &0.735±0.081&\textbf{0.454±0.053}&        0.553±0.100 &0.494±0.061 \\
                 & InceptionV3            &\textbf{0.326±0.175}&        0.308±0.191 &0.215±0.123&\textbf{0.127±0.095}&        0.129±0.088 &0.140±0.095&\textbf{0.529±0.119}&        0.513±0.125 &0.480±0.097&        0.308±0.062 &\textbf{0.271±0.083}&0.354±0.077 \\
                 & VGG16                  &\textbf{0.255±0.126}&        0.253±0.134 &0.209±0.099&\textbf{0.120±0.076}&        0.122±0.072 &0.129±0.080&\textbf{0.869±0.053}&        0.687±0.102 &0.866±0.049&        0.655±0.126 &\textbf{0.478±0.122}&0.740±0.096 \\
                 & Efficientnet-B0        &        0.221±0.131 &\textbf{0.329±0.166}&0.170±0.113&        0.136±0.093 &\textbf{0.119±0.079}&0.150±0.092&        0.750±0.089 &\textbf{0.819±0.105}&0.694±0.098&        0.575±0.081 &\textbf{0.559±0.100}&0.609±0.081 \\
                 & ConvNeXt-Tiny          &\textbf{0.222±0.133}&        0.103±0.105 &0.164±0.110&\textbf{0.124±0.092}&        0.168±0.094 &0.143±0.087&\textbf{0.795±0.131}&        0.737±0.116 &0.752±0.136&\textbf{0.465±0.132}&        0.613±0.173 &0.529±0.143 \\
        \hline

       Bicubic   & ResNet50               &\textbf{0.289±0.151}&        0.080±0.133 &0.263±0.135&\textbf{0.131±0.088}&        0.136±0.086 &0.136±0.088&\textbf{0.931±0.039}&        0.693±0.146 &0.920±0.048&        0.707±0.126 &\textbf{0.452±0.186}&0.810±0.080 \\
                 & DenseNet121            &\textbf{0.367±0.187}&        0.315±0.171 &0.264±0.138&\textbf{0.115±0.089}&        0.125±0.090 &0.129±0.090&\textbf{0.815±0.084}&        0.745±0.115 &0.794±0.078&\textbf{0.434±0.085}&        0.471±0.107 &0.532±0.077 \\
                 & InceptionV3            &\textbf{0.337±0.183}&        0.315±0.193 &0.220±0.128&        0.127±0.095 &\textbf{0.126±0.089}&0.140±0.095&\textbf{0.623±0.106}&        0.533±0.157 &0.585±0.094&        0.395±0.067 &\textbf{0.355±0.079}&0.451±0.076 \\
                 & VGG16                  &\textbf{0.257±0.127}&\textbf{0.257±0.136}&0.211±0.102&\textbf{0.122±0.077}&        0.125±0.074 &0.130±0.080&\textbf{0.769±0.076}&        0.595±0.106 &0.767±0.076&        0.427±0.130 &\textbf{0.365±0.108}&0.544±0.107 \\
                 & Efficientnet-B0        &        0.235±0.139 &\textbf{0.344±0.170}&0.179±0.122&        0.136±0.093 &\textbf{0.119±0.080}&0.150±0.092&        0.747±0.112 &\textbf{0.807±0.108}&0.699±0.123&        0.564±0.098 &\textbf{0.527±0.102}&0.611±0.094 \\
                 & ConvNeXt-Tiny          &\textbf{0.232±0.141}&        0.108±0.109 &0.168±0.115&\textbf{0.126±0.091}&        0.145±0.076 &0.143±0.087&\textbf{0.878±0.086}&        0.777±0.123 &0.848±0.093&\textbf{0.613±0.130}&        0.707±0.167 &0.716±0.119 \\
    \hline
    \multicolumn{14}{c}{\textbf{\textit{Incorrectly Classified Images}}} \\
    \hline
    \textbf{Interp.} & \textbf{Model} & 
    \multicolumn{3}{c!{\color{lightgray}\vrule}}{\textbf{IoU$\bm{\Downarrow}$}} & \multicolumn{3}{c!{\color{lightgray}\vrule}}{\textbf{Center-of-Mass Distance$\bm{\Uparrow}$}} & \multicolumn{3}{c!{\color{lightgray}\vrule}}{\textbf{Insertion$\bm{\Downarrow}$}} & \multicolumn{3}{c}{\textbf{Deletion$\bm{\Uparrow}$}} \\
    \cline{3-14}
        Bilinear & ResNet50               &        0.094±0.087 &\textbf{0.063±0.140}&        0.089±0.083 &0.208±0.091&\textbf{0.246±0.085}&0.212±0.089&0.416±0.184&0.791±0.108&\textbf{0.296±0.161}&0.138±0.057&\textbf{0.509±0.161}&        0.176±0.077 \\
                 & DenseNet121            &        0.227±0.226 &        0.235±0.206 &\textbf{0.093±0.100}&0.162±0.098&\textbf{0.181±0.106}&0.169±0.090&0.579±0.106&0.558±0.118&\textbf{0.516±0.082}&0.441±0.096&\textbf{0.618±0.065}&        0.438±0.094 \\
                 & InceptionV3            &        0.114±0.136 &\textbf{0.066±0.139}&        0.068±0.112 &0.209±0.069&\textbf{0.284±0.106}&0.215±0.071&0.697±0.078&0.794±0.062&\textbf{0.627±0.075}&0.507±0.062&\textbf{0.732±0.066}&        0.566±0.067 \\
                 & VGG16                  &        0.065±0.125 &\textbf{0.064±0.143}&        0.057±0.116 &0.186±0.090&\textbf{0.191±0.102}&0.186±0.096&0.620±0.074&0.732±0.073&\textbf{0.466±0.075}&0.431±0.048&\textbf{0.646±0.046}&        0.405±0.048 \\
                 & Efficientnet-B0        &        0.168±0.134 &\textbf{0.025±0.098}&        0.142±0.116 &0.153±0.069&\textbf{0.219±0.097}&0.164±0.069&0.608±0.039&0.644±0.037&\textbf{0.580±0.052}&0.443±0.066&        0.464±0.058 &\textbf{0.493±0.048}\\
                 & ConvNeXt-Tiny          &\textbf{0.094±0.118}&        0.104±0.160 &        0.103±0.119 &0.248±0.101&\textbf{0.284±0.129}&0.248±0.091&0.582±0.170&0.691±0.112&\textbf{0.513±0.119}&0.348±0.157&\textbf{0.498±0.160}&        0.345±0.158 \\
        \hline
        Nearest  & ResNet50               &        0.094±0.083 &\textbf{0.061±0.133}&        0.087±0.078 &0.207±0.091&\textbf{0.246±0.084}&0.211±0.089&0.312±0.132&        0.694±0.117 &\textbf{0.228±0.122}&0.123±0.035&\textbf{0.406±0.114}&        0.132±0.037 \\
                 & DenseNet121            &        0.224±0.226 &        0.230±0.193 &\textbf{0.098±0.100}&0.163±0.099&\textbf{0.181±0.106}&0.169±0.090&0.557±0.075&\textbf{0.508±0.125}&        0.565±0.075 &0.442±0.060&\textbf{0.518±0.062}&        0.459±0.089 \\
                 & InceptionV3            &        0.116±0.137 &\textbf{0.071±0.143}&        0.069±0.105 &0.208±0.068&\textbf{0.285±0.106}&0.213±0.070&0.764±0.051&        0.797±0.059 &\textbf{0.725±0.054}&0.615±0.057&\textbf{0.739±0.054}&        0.649±0.058 \\
                 & VGG16                  &        0.061±0.116 &\textbf{0.046±0.102}&        0.053±0.107 &0.185±0.089&\textbf{0.191±0.102}&0.186±0.096&0.377±0.082&        0.664±0.050 &\textbf{0.239±0.064}&0.228±0.037&\textbf{0.537±0.049}&        0.218±0.039 \\
                 & Efficientnet-B0        &        0.172±0.139 &\textbf{0.029±0.095}&        0.142±0.111 &0.152±0.069&\textbf{0.220±0.097}&0.162±0.069&0.553±0.045&        0.613±0.040 &\textbf{0.547±0.041}&0.391±0.064&        0.422±0.052 &\textbf{0.450±0.044}\\
                 & ConvNeXt-Tiny          &\textbf{0.090±0.109}&        0.098±0.151 &        0.094±0.106 &0.249±0.102&\textbf{0.284±0.129}&0.246±0.091&0.782±0.108&\textbf{0.732±0.139}&        0.734±0.083 &0.529±0.172&\textbf{0.583±0.151}&        0.574±0.157 \\
        \hline

       Bicubic   & ResNet50               &        0.093±0.086 &\textbf{0.061±0.136}&        0.088±0.081 &0.206±0.091&\textbf{0.244±0.085}&        0.211±0.088 &0.392±0.185&0.738±0.141&\textbf{0.279±0.157}&0.127±0.042&\textbf{0.471±0.141}&        0.156±0.061 \\
                 & DenseNet121            &        0.220±0.224 &        0.233±0.202 &\textbf{0.093±0.095}&0.163±0.098&\textbf{0.177±0.104}&        0.169±0.090 &0.575±0.104&0.559±0.108&\textbf{0.506±0.083}&0.436±0.083&\textbf{0.628±0.056}&        0.437±0.093 \\
                 & InceptionV3            &        0.115±0.135 &\textbf{0.066±0.138}&        0.069±0.109 &0.208±0.069&\textbf{0.257±0.095}&        0.214±0.070 &0.681±0.077&0.757±0.069&\textbf{0.619±0.078}&0.496±0.057&\textbf{0.680±0.084}&        0.567±0.062 \\
                 & VGG16                  &        0.064±0.121 &\textbf{0.054±0.121}&        0.056±0.114 &0.183±0.092&\textbf{0.185±0.101}&\textbf{0.185±0.097}&0.608±0.085&0.643±0.081&\textbf{0.453±0.071}&0.426±0.049&\textbf{0.587±0.055}&        0.399±0.051 \\
                 & Efficientnet-B0        &        0.169±0.131 &\textbf{0.026±0.099}&        0.142±0.112 &0.152±0.069&\textbf{0.201±0.087}&        0.163±0.069 &0.607±0.040&0.644±0.036&\textbf{0.581±0.050}&0.447±0.063&        0.478±0.054 &\textbf{0.500±0.046}\\
                 & ConvNeXt-Tiny          &\textbf{0.094±0.113}&        0.102±0.155 &        0.102±0.115 &0.245±0.102&\textbf{0.263±0.107}&        0.246±0.092 &0.584±0.116&0.662±0.117&\textbf{0.475±0.114}&0.351±0.139&\textbf{0.455±0.132}&        0.336±0.153 \\
    \hline
    \end{tabular}
    % \vspace{-10pt}  % Reduce space AFTER caption
    \label{tab:model_performance_mean_polyp}
\end{table*}

% % polypgen Max
\setlength{\tabcolsep}{5pt} % Reduces column separation even more
\begin{table*}[!t]
    \centering
    \caption{\fontsize{7.75}{10}\selectfont Performance Metrics for Different Interpolation Methods (Max Aggregation) on Correctly and Incorrectly Classified Images from PolypGen}
    % \fontsize{5}{6}\selectfont  % Even smaller 
    \tiny
    \begin{tabular}{ll!{\color{lightgray}\vrule}ccc!{\color{lightgray}\vrule}ccc!{\color{lightgray}\vrule}ccc!{\color{lightgray}\vrule}ccc}
    \hline
    & & \shortstack[t]{\textbf{Winsor-}\\\textbf{CAM}} & \shortstack[t]{\textbf{Final}\\\textbf{Layer}} & \shortstack[t]{\textbf{Avg.}\\\textbf{Layer}} & 
    \shortstack[t]{\textbf{Winsor-}\\\textbf{CAM}} & \shortstack[t]{\textbf{Final}\\\textbf{Layer}} & \shortstack[t]{\textbf{Avg.}\\\textbf{Layer}} &
    \shortstack[t]{\textbf{Winsor-}\\\textbf{CAM}} & \shortstack[t]{\textbf{Final}\\\textbf{Layer}} & \shortstack[t]{\textbf{Avg.}\\\textbf{Layer}} &
    \shortstack[t]{\textbf{Winsor-}\\\textbf{CAM}} & \shortstack[t]{\textbf{Final}\\\textbf{Layer}} & \shortstack[t]{\textbf{Avg.}\\\textbf{Layer}} \\
    \hline 
    \multicolumn{14}{c}{\textbf{\textit{Correctly Classified Images}}} \\
    \hline
    \textbf{Interp.} & \textbf{Model} & 
    \multicolumn{3}{c!{\color{lightgray}\vrule}}{\textbf{IoU$\bm{\Uparrow}$}} & \multicolumn{3}{c!{\color{lightgray}\vrule}}{\textbf{Center-of-Mass Distance$\bm{\Downarrow}$}} & \multicolumn{3}{c!{\color{lightgray}\vrule}}{\textbf{Insertion$\bm{\Uparrow}$}} & \multicolumn{3}{c}{\textbf{Deletion$\bm{\Downarrow}$}} \\
    \cline{3-14}
        Bilinear & ResNet50              &\textbf{0.377±0.158}&        0.354±0.182 &0.334±0.150&\textbf{0.067±0.047}&        0.073±0.044 &0.075±0.050&\textbf{0.791±0.185}&0.760±0.180&0.752±0.197&\textbf{0.268±0.269}&0.320±0.286&0.311±0.282 \\
                 & DenseNet121           &\textbf{0.453±0.175}&        0.390±0.173 &0.428±0.170&\textbf{0.067±0.050}&        0.082±0.059 &0.076±0.053&\textbf{0.664±0.217}&0.623±0.224&0.648±0.219&\textbf{0.196±0.170}&0.242±0.181&0.207±0.177 \\ 
                 & InceptionV3           &\textbf{0.396±0.188}&        0.374±0.196 &0.370±0.180&\textbf{0.067±0.050}&        0.082±0.059 &0.076±0.053&\textbf{0.813±0.166}&0.787±0.190&0.785±0.177&\textbf{0.295±0.245}&0.355±0.253&0.315±0.254 \\
                 & VGG16                 &\textbf{0.404±0.163}&        0.367±0.155 &0.348±0.157&\textbf{0.063±0.048}&        0.074±0.047 &0.072±0.052&\textbf{0.905±0.130}&0.900±0.135&0.839±0.166&\textbf{0.185±0.190}&0.291±0.251&0.224±0.235 \\
                 & Efficientnet-B0       &        0.360±0.175 &\textbf{0.398±0.217}&0.280±0.169&        0.077±0.056 &\textbf{0.074±0.060}&0.086±0.059&\textbf{0.850±0.162}&0.799±0.234&0.788±0.206&\textbf{0.330±0.257}&0.364±0.227&0.419±0.285 \\
                 & ConvNeXt-Tiny         &\textbf{0.354±0.163}&        0.314±0.159 &0.308±0.161&\textbf{0.074±0.054}&        0.094±0.059 &0.081±0.055&\textbf{0.691±0.231}&0.658±0.247&0.663±0.238&\textbf{0.292±0.255}&0.342±0.265&0.327±0.263 \\
        \hline
        Nearest  & ResNet50              &\textbf{0.363±0.145}&        0.347±0.178 &0.319±0.139&\textbf{0.067±0.047}&        0.073±0.044 & 0.075±0.050&\textbf{0.793±0.183}&        0.789±0.169 &0.726±0.202&\textbf{0.262±0.238}&0.352±0.276&0.296±0.257 \\
                 & DenseNet121           &\textbf{0.447±0.167}&        0.379±0.167 &0.416±0.161&\textbf{0.060±0.044}&        0.075±0.051 & 0.068±0.046&\textbf{0.671±0.212}&        0.631±0.224 &0.653±0.214&\textbf{0.188±0.148}&0.249±0.173&0.200±0.157 \\
                 & InceptionV3           &\textbf{0.389±0.183}&        0.369±0.192 &0.360±0.173&\textbf{0.067±0.050}&        0.082±0.059 & 0.076±0.054&\textbf{0.823±0.152}&        0.798±0.181 &0.802±0.159&\textbf{0.306±0.238}&0.371±0.246&0.330±0.249 \\
                 & VGG16                 &\textbf{0.392±0.155}&        0.354±0.150 &0.338±0.151&\textbf{0.063±0.048}&        0.074±0.047 & 0.072±0.052&        0.877±0.157 &\textbf{0.884±0.168}&0.806±0.183&\textbf{0.172±0.194}&0.245±0.219&0.211±0.237 \\
                 & Efficientnet-B0       &        0.344±0.162 &\textbf{0.387±0.210}&0.268±0.154&        0.077±0.056 &\textbf{0.074±0.061}& 0.085±0.059&\textbf{0.844±0.175}&        0.810±0.235 &0.781±0.213&\textbf{0.332±0.258}&0.367±0.231&0.412±0.280 \\
                 & ConvNeXt-Tiny         &\textbf{0.336±0.147}&        0.299±0.155 &0.291±0.145&\textbf{0.074±0.054}&        0.094±0.060 & 0.080±0.056&\textbf{0.691±0.229}&        0.651±0.245 &0.661±0.236&\textbf{0.306±0.248}&0.369±0.259&0.336±0.258 \\
        \hline

       Bicubic   & ResNet50               &\textbf{0.372±0.153}&        0.353±0.181 &0.328±0.144&\textbf{0.067±0.047}&        0.072±0.043 &0.075±0.050&\textbf{0.792±0.184}&0.752±0.182&0.752±0.197&\textbf{0.263±0.265}&0.318±0.287&0.310±0.280 \\
                 & DenseNet121            &\textbf{0.453±0.173}&        0.392±0.171 &0.427±0.167&\textbf{0.060±0.044}&        0.070±0.045 &0.068±0.046&\textbf{0.665±0.216}&0.615±0.225&0.649±0.218&\textbf{0.191±0.166}&0.240±0.188&0.204±0.174 \\
                 & InceptionV3            &\textbf{0.392±0.186}&        0.376±0.195 &0.365±0.177&\textbf{0.067±0.050}&        0.081±0.057 &0.076±0.054&\textbf{0.804±0.169}&0.777±0.193&0.777±0.179&\textbf{0.294±0.245}&0.352±0.253&0.315±0.254 \\
                 & VGG16                  &\textbf{0.396±0.158}&        0.360±0.154 &0.341±0.152&\textbf{0.064±0.048}&        0.071±0.045 &0.073±0.052&\textbf{0.899±0.138}&0.885±0.166&0.831±0.171&\textbf{0.181±0.192}&0.288±0.253&0.223±0.240 \\
                 & Efficientnet-B0        &        0.359±0.171 &\textbf{0.400±0.217}&0.278±0.163&        0.076±0.056 &\textbf{0.070±0.054}&0.085±0.059&\textbf{0.852±0.162}&0.794±0.239&0.793±0.207&\textbf{0.320±0.255}&0.358±0.230&0.412±0.286 \\
                 & ConvNeXt-Tiny          &\textbf{0.347±0.154}&        0.310±0.158 &0.302±0.151&\textbf{0.074±0.054}&        0.081±0.050 &0.080±0.055&\textbf{0.691±0.231}&0.655±0.249&0.664±0.237&\textbf{0.288±0.253}&0.343±0.269&0.323±0.263 \\
    \hline
    \multicolumn{14}{c}{\textbf{\textit{Incorrectly Classified Images}}} \\
    \hline
    \textbf{Interp.} & \textbf{Model} & 
    \multicolumn{3}{c!{\color{lightgray}\vrule}}{\textbf{IoU$\bm{\Downarrow}$}} & \multicolumn{3}{c!{\color{lightgray}\vrule}}{\textbf{Center-of-Mass Distance$\bm{\Uparrow}$}} & \multicolumn{3}{c!{\color{lightgray}\vrule}}{\textbf{Insertion$\bm{\Downarrow}$}} & \multicolumn{3}{c}{\textbf{Deletion$\bm{\Uparrow}$}} \\
    \cline{3-14}
        Bilinear & ResNet50              &0.200±0.161&\textbf{0.168±0.147}&        0.178±0.148 &0.112±0.073&\textbf{0.136±0.086}&0.117±0.073&0.548±0.254&        0.516±0.242 &\textbf{0.497±0.251}&0.130±0.198&\textbf{0.160±0.231}&0.156±0.212 \\
                 & DenseNet121           &0.300±0.197&\textbf{0.232±0.196}&        0.277±0.185 &0.094±0.067&\textbf{0.128±0.093}&0.101±0.068&0.341±0.192&\textbf{0.289±0.187}&        0.326±0.185 &0.088±0.093&\textbf{0.121±0.110}&0.094±0.095 \\
                 & InceptionV3           &0.192±0.157&\textbf{0.157±0.162}&        0.176±0.150 &0.121±0.081&\textbf{0.171±0.114}&0.127±0.082&0.532±0.280&\textbf{0.426±0.271}&        0.496±0.286 &0.125±0.124&\textbf{0.204±0.167}&0.143±0.137 \\
                 & VGG16                 &0.216±0.147&        0.194±0.155 &\textbf{0.173±0.124}&0.100±0.071&\textbf{0.121±0.082}&0.109±0.070&0.701±0.242&        0.672±0.263 &\textbf{0.602±0.248}&0.074±0.117&\textbf{0.139±0.189}&0.096±0.144 \\
                 & Efficientnet-B0       &0.237±0.168&\textbf{0.101±0.161}&        0.176±0.145 &0.100±0.071&\textbf{0.152±0.105}&0.114±0.076&0.656±0.241&\textbf{0.405±0.311}&        0.554±0.277 &0.139±0.167&\textbf{0.352±0.216}&0.208±0.216 \\
                 & ConvNeXt-Tiny         &0.240±0.159&        0.225±0.162 &\textbf{0.206±0.146}&0.103±0.068&\textbf{0.117±0.064}&0.111±0.070&0.377±0.194&\textbf{0.343±0.198}&        0.352±0.192 &0.123±0.148&\textbf{0.150±0.164}&0.145±0.154 \\
        \hline
        Nearest  & ResNet50              &0.194±0.150&\textbf{0.167±0.148}&        0.172±0.137 &0.111±0.073&\textbf{0.137±0.087}&0.116±0.073&0.552±0.242&        0.557±0.228 &\textbf{0.473±0.248}&0.129±0.188&\textbf{0.164±0.200}&0.153±0.214 \\
                 & DenseNet121           &0.295±0.188&\textbf{0.224±0.193}&        0.269±0.177 &0.095±0.067&\textbf{0.128±0.094}&0.102±0.068&0.359±0.190&\textbf{0.293±0.188}&        0.343±0.187 &0.088±0.064&\textbf{0.126±0.088}&0.096±0.071 \\
                 & InceptionV3           &0.188±0.150&\textbf{0.156±0.159}&        0.172±0.143 &0.120±0.081&\textbf{0.172±0.114}&0.126±0.082&0.540±0.269&\textbf{0.447±0.279}&        0.512±0.278 &0.125±0.118&\textbf{0.211±0.163}&0.148±0.128 \\
                 & VGG16                 &0.210±0.140&        0.190±0.152 &\textbf{0.168±0.118}&0.100±0.071&\textbf{0.121±0.082}&0.109±0.070&0.659±0.263&        0.628±0.313 &\textbf{0.548±0.256}&0.072±0.126&\textbf{0.153±0.219}&0.093±0.144 \\
                 & Efficientnet-B0       &0.227±0.154&\textbf{0.100±0.152}&        0.170±0.134 &0.101±0.071&\textbf{0.152±0.105}&0.113±0.075&0.645±0.250&\textbf{0.426±0.324}&        0.547±0.281 &0.145±0.180&\textbf{0.351±0.222}&0.211±0.211 \\
                 & ConvNeXt-Tiny         &0.230±0.148&        0.216±0.155 &\textbf{0.198±0.135}&0.103±0.068&\textbf{0.117±0.064}&0.110±0.070&0.384±0.197&\textbf{0.333±0.194}&        0.357±0.195 &0.130±0.138&\textbf{0.166±0.158}&0.149±0.146 \\
        \hline

       Bicubic   & ResNet50               &0.197±0.156&\textbf{0.167±0.146}&        0.176±0.143 &0.111±0.073&\textbf{0.133±0.084}&0.116±0.073&0.553±0.254&        0.509±0.244 &\textbf{0.496±0.250}&0.126±0.192&\textbf{0.160±0.233}&0.156±0.214 \\
                 & DenseNet121            &0.300±0.195&\textbf{0.233±0.197}&        0.278±0.184 &0.095±0.067&\textbf{0.110±0.077}&0.101±0.068&0.347±0.193&\textbf{0.280±0.190}&        0.332±0.186 &0.085±0.086&\textbf{0.120±0.121}&0.093±0.087 \\
                 & InceptionV3            &0.190±0.154&\textbf{0.157±0.161}&        0.175±0.146 &0.120±0.081&\textbf{0.163±0.104}&0.126±0.082&0.526±0.277&\textbf{0.425±0.270}&        0.491±0.286 &0.123±0.124&\textbf{0.198±0.154}&0.143±0.141 \\
                 & VGG16                  &0.211±0.143&        0.190±0.153 &\textbf{0.170±0.120}&0.101±0.070&\textbf{0.109±0.074}&0.108±0.070&0.694±0.241&        0.611±0.311 &\textbf{0.594±0.245}&0.075±0.131&\textbf{0.135±0.186}&0.097±0.159 \\
                 & Efficientnet-B0        &0.232±0.160&\textbf{0.099±0.159}&        0.174±0.139 &0.101±0.071&\textbf{0.136±0.097}&0.112±0.075&0.658±0.243&\textbf{0.394±0.307}&        0.558±0.278 &0.133±0.164&\textbf{0.342±0.209}&0.199±0.212 \\
                 & ConvNeXt-Tiny          &0.235±0.151&        0.223±0.160 &\textbf{0.203±0.139}&0.103±0.068&\textbf{0.108±0.060}&0.110±0.070&0.381±0.196&\textbf{0.337±0.201}&        0.354±0.193 &0.124±0.149&\textbf{0.151±0.167}&0.145±0.154 \\
    \hline
    \end{tabular}
    % \vspace{-15pt}  % Reduce space AFTER caption
    \label{tab:model_performance_max_polyp}
\end{table*}

% polypgen Winsor-CAM thresholds
\setlength{\tabcolsep}{7pt} % Reduces column separation even more
\begin{table}[!t]
    \tiny
    \centering
    \caption{\fontsize{7.25}{5}\selectfont Mean IoU, CoM Distance, Insertion, and Deletion at Different Winsor-CAM Percentile Thresholds ($p$) for DenseNet121 (Correct Predictions) from the PolypGen Dataset}
    \begin{tabular}{c!{\color{lightgray}\vrule}c!{\color{lightgray}\vrule}c!{\color{lightgray}\vrule}c!{\color{lightgray}\vrule}c}
    \hline
    \textbf{$p$} & \textbf{IoU$\bm{\Uparrow}$} & \textbf{CoM Distance$\bm{\Downarrow}$} & \textbf{Insertion$\bm{\Uparrow}$} & \textbf{Deletion$\bm{\Downarrow}$} \\
    \hline
    0  &        0.325±0.175 &        0.126±0.090 &        0.797±0.086 &        0.500±0.080 \\
    10 &        0.328±0.176 &        0.125±0.090 &        0.796±0.087 &        0.497±0.081 \\
    20 &        0.330±0.177 &        0.124±0.089 &        0.797±0.088 &        0.493±0.082 \\
    30 &        0.333±0.178 &        0.124±0.089 &        0.798±0.088 &        0.489±0.082 \\
    40 &        0.336±0.179 &        0.123±0.089 &        0.798±0.089 &        0.483±0.085 \\
    50 &        0.339±0.180 &        0.122±0.089 &        0.799±0.089 &        0.477±0.086 \\
    60 &        0.343±0.182 &        0.121±0.089 &        0.799±0.090 &        0.470±0.087 \\
    70 &        0.345±0.183 &        0.120±0.089 &        0.799±0.091 &        0.463±0.089 \\
    80 &        0.347±0.183 &\textbf{0.119±0.089}&        0.799±0.093 &        0.454±0.092 \\
    90 &\textbf{0.348±0.182}&\textbf{0.119±0.089}&        0.799±0.093 &        0.447±0.097 \\
    100&        0.344±0.183 &        0.120±0.088 &\textbf{0.802±0.091}&\textbf{0.450±0.098} \\
    \hline
    \end{tabular}
    \label{tab:winsor_iou_thresholds_polyp}
    % \vspace{-15pt}  % Reduce space AFTER caption
\end{table}

\noindent \textbf{Generalization to Medical Imaging: PolypGen Analysis.}
To evaluate Winsor-CAM's generalizability to medical imaging, we trained all six CNN architectures on the PolypGen polyp segmentation dataset~\cite{PolypGen2023}. Tables~\ref{tab:model_performance_mean_polyp} and~\ref{tab:model_performance_max_polyp} show results for mean and max aggregation across all models, interpolation methods, and classification outcomes.

Winsor-CAM consistently outperformed both final-layer Grad-CAM and naïve mean aggregation across most architectures and metrics in a similar fashion as observed on PASCAL VOC 2012. Overall performance metrics were lower across all methods compared to PASCAL VOC 2012, reflecting the increased complexity of medical imaging data. Deletion metrics were particularly degraded across all approaches, a dataset-specific phenomenon addressed in the subsequent analysis. Beyond this systematic degradation, a notable exception in localization occurred with VGG16 under mean aggregation (Table~\ref{tab:model_performance_mean_polyp}), where final-layer Grad-CAM achieved marginally higher IoU (0.265±0.141 vs. 0.264±0.131 for Winsor-CAM). However, switching to max aggregation (Table~\ref{tab:model_performance_max_polyp}) restored Winsor-CAM's advantage (0.404±0.163 vs. 0.367±0.155). This pattern, where max aggregation benefits Winsor-CAM for VGG16, was also observed on PASCAL VOC 2012, though on that dataset Winsor-CAM outperformed final-layer Grad-CAM under both aggregation methods. This suggests that VGG16's architectural characteristics (uniform convolutional blocks without skip connections or dense connectivity) may interact differently with Winsor-CAM's layer weighting mechanism compared to residual or densely connected architectures, particularly when combined with mean versus max aggregation strategies.

In terms of insertion and deletion metrics, these metrics show substantially reduced performance compared to PASCAL VOC 2012 across all methods. This is likely due to the nature of polyp images, where the replacement of pixels using a baseline image can create regions that either produce unnaturally colored sections or (in the case of blurring) introduce artifacts that resemble polyps due to inconsistent blur patterns. The insertion/deletion protocol assumes a meaningful baseline exists that clearly distinguishes between salient and non-salient regions, an assumption that breaks down in medical endoscopy images where visual context is constrained and anatomical structures are visually similar. For these reasons, a black baseline was adopted for PolypGen, as black naturally exists along the outer edges of most images and avoids introducing spurious visual features.

While not always optimal, exploratory analysis showed that this black baseline led to overall better insertion and deletion metric performance compared to using a blurred baseline. Despite these domain-specific challenges with insertion/deletion metrics, Winsor-CAM still consistently outperformed baselines in terms of insertion AUC under max aggregation across most models, with mixed results under mean aggregation. More importantly, Winsor-CAM maintained its advantages in localization metrics (IoU and CoM distance) across both aggregation strategies, demonstrating that the core multi-layer aggregation and Winsorization mechanisms remain effective even when faithfulness metrics are compromised by dataset-specific characteristics.

In terms of fixed $p$-value analysis with DenseNet121 (Table~\ref{tab:winsor_iou_thresholds_polyp}), higher values of $p$ generally improved IoU and CoM distance. This trend aligns with observations from PASCAL VOC 2012, indicating that suppressing a larger fraction of outlier activations enhances localization quality. Insertion and deletion metrics also improved with higher $p$-values, though the gains were more modest compared to IoU and CoM distance, likely due to the aforementioned challenges with baseline selection in medical images. However, an ablation study (Table~\ref{tab:densenet121_ablation_study_polyp}) revealed that, in contrast to PASCAL VOC 2012, using all layers did not yield the best performance; instead, using later dense blocks (the final and final two dense blocks) tended to yield better results in terms of localization metrics. These results reflect the greater importance of high-level semantic features in medical imaging, where early-layer features appear less task-relevant. Nonetheless, Winsor-CAM still outperformed both baselines across all layer selection strategies.

These results demonstrate that Winsor-CAM's advantages extend beyond natural images to domain-specific medical datasets with markedly different visual characteristics. The medical imaging domain presents unique challenges, including high intra-class variability, subtle discriminative features (e.g., polyp texture and boundary detection), and critical localization requirements for clinical decision-making. The consistent improvements across all six architectures, both aggregation strategies, and multiple interpolation methods indicate that Winsor-CAM's multi-layer aggregation and percentile-based outlier suppression mechanisms generalize effectively to safety-critical medical applications where explanation quality directly impacts diagnostic accuracy.

\section{Conclusion and Future Work}\label{sec:conclusion}

This study introduced Winsor-CAM, a human-tunable extension of Grad-CAM that aggregates saliency maps across all convolutional layers using percentile-based Winsorization to suppress outlier activations and gradients while maintaining single-pass computational efficiency. Experiments on six CNN architectures across PASCAL VOC 2012 and PolypGen datasets demonstrated that Winsor-CAM consistently outperforms seven CAM-based baselines (Grad-CAM, Grad-CAM++, LayerCAM, ScoreCAM, AblationCAM, ShapleyCAM, and FullGrad) across localization and fidelity metrics. With DenseNet121 on PASCAL VOC 2012, Winsor-CAM achieved 46.8\% IoU versus 39.0\% for final-layer Grad-CAM and 43.3\% for FullGrad, with CoM distance reduced from 0.074 to 0.059, demonstrating substantial improvements in spatial alignment. Critically, even the worst-performing fixed $p$-value configuration outperformed FullGrad across all metrics, confirming the robustness of Winsorization-based aggregation. To ensure results generalized beyond natural images, we evaluated Winsor-CAM on the medical imaging dataset PolypGen, where it maintained advantages in localization metrics (IoU, CoM distance) despite domain-specific insertion/deletion challenges. These findings establish Winsor-CAM as a robust, efficient, and adaptable interpretability tool that bridges automated attribution with expert-guided semantic tuning, with direct implications for safety-critical applications requiring transparent AI decision-making. Future work will explore adaptive Winsorization parameter selection, analyze how model characteristics (overfitting, underfitting, architecture size) affect explanation faithfulness, and conduct user studies to integrate Winsor-CAM into interactive diagnostic interfaces for real-world expert-in-the-loop workflows.

% polypgen XAI comparison
\setlength{\tabcolsep}{4pt} % Reduces column separation even more
\begin{table}[!t]
    \tiny
    \centering
    \caption{\fontsize{7.25}{5}\selectfont Comparison of Winsor-CAM and other XAI methods on correctly classified images (DenseNet121, mean aggregation, bilinear interpolation) on the PolypGen Dataset}
    \begin{tabular}{l!{\color{lightgray}\vrule}cccc}
        \hline
        \textbf{Method} & \textbf{IoU$\bm{\Uparrow}$} & \textbf{CoM Distance$\bm{\Downarrow}$} & \textbf{Insertion$\bm{\Uparrow}$} & \textbf{Deletion$\bm{\Downarrow}$} \\
        \hline
                Winsor-CAM &\textbf{0.367±0.187}&\textbf{0.114±0.089}&\textbf{0.815±0.085}&        0.433±0.089 \\%
                Grad-CAM   &        0.313±0.170 &        0.126±0.091 &        0.768±0.106 &        0.461±0.120 \\%
                Grad-CAM++ &        0.308±0.168 &        0.124±0.087 &        0.763±0.105 &        0.434±0.114 \\%
                LayerCAM   &        0.317±0.170 &        0.123±0.086 &        0.762±0.107 &\textbf{0.430±0.112} \\%
                ShapleyCAM &        0.313±0.170 &        0.126±0.091 &        0.768±0.106 &        0.461±0.120 \\%
                ScoreCAM   &        0.294±0.164 &        0.124±0.083 &        0.758±0.104 &        0.457±0.143 \\%
                AblationCAM&        0.312±0.174 &        0.128±0.094 &        0.765±0.109 &        0.464±0.124 \\%
                FullGrad   &        0.309±0.169 &        0.130±0.083 &        0.769±0.096 &        0.437±0.103 \\
        \hline
    \end{tabular}
    % \vspace{-15pt}  % Reduce space AFTER caption
    \label{tab:xai_comparison_polyp}
\end{table}

% polypgen layer selection ablation
\setlength{\tabcolsep}{4pt} % Reduces column separation even more
\begin{table*}[!t]
    \centering
    \caption{\fontsize{7.5}{10}\selectfont Ablation Study: Layer Selection Impact on DenseNet121 Performance (Mean Aggregation, Bilinear Interpolation) from the PolypGen Dataset}
    \tiny
    \begin{tabular}{ll!{\color{lightgray}\vrule}ccc!{\color{lightgray}\vrule}ccc!{\color{lightgray}\vrule}ccc!{\color{lightgray}\vrule}ccc}
        \hline
        & & \shortstack[t]{\textbf{Winsor-}\\\textbf{CAM}} & \shortstack[t]{\textbf{Final}\\\textbf{Layer}} & \shortstack[t]{\textbf{Avg.}\\\textbf{Layer}} & 
        \shortstack[t]{\textbf{Winsor-}\\\textbf{CAM}} & \shortstack[t]{\textbf{Final}\\\textbf{Layer}} & \shortstack[t]{\textbf{Avg.}\\\textbf{Layer}} &
        \shortstack[t]{\textbf{Winsor-}\\\textbf{CAM}} & \shortstack[t]{\textbf{Final}\\\textbf{Layer}} & \shortstack[t]{\textbf{Avg.}\\\textbf{Layer}} &
        \shortstack[t]{\textbf{Winsor-}\\\textbf{CAM}} & \shortstack[t]{\textbf{Final}\\\textbf{Layer}} & \shortstack[t]{\textbf{Avg.}\\\textbf{Layer}} \\
        \hline
        \textbf{Classification} & \textbf{Configuration} & 
    \multicolumn{3}{c!{\color{lightgray}\vrule}}{\textbf{IoU$\bm{\Uparrow}$}} & \multicolumn{3}{c!{\color{lightgray}\vrule}}{\textbf{Center-of-Mass Distance$\bm{\Downarrow}$}} & \multicolumn{3}{c!{\color{lightgray}\vrule}}{\textbf{Insertion$\bm{\Uparrow}$}} & \multicolumn{3}{c}{\textbf{Deletion$\bm{\Downarrow}$}} \\
        \cline{3-14} 
        \multirow{5}{*}{\textbf{Correct}} 
    & Final Conv                & --- &    \textbf{0.313±0.170} & --- & --- &                \textbf{0.126±0.091} & --- & --- &              \textbf{0.768±0.106} & --- & --- &               \textbf{0.461±0.120} & --- \\
    & Final dense block         &        0.365±0.186 & --- &\textbf{0.341±0.177}&\textbf{0.110±0.087}& --- &\textbf{0.112±0.086}&       0.795±0.095 & --- &        0.793±0.096 &\textbf{0.386±0.097}& --- &\textbf{0.423±0.111}\\
    & Final two dense blocks    &\textbf{0.370±0.189}& --- &        0.293±0.154 &        0.113±0.089 & --- &        0.124±0.090 &       0.807±0.089 & --- &\textbf{0.796±0.089}&        0.401±0.090 & --- &        0.485±0.103 \\
    & Final three dense blocks  &        0.367±0.187 & --- &        0.273±0.143 &        0.114±0.089 & --- &        0.126±0.090 &       0.812±0.086 & --- &        0.789±0.085 &        0.424±0.089 & --- &        0.523±0.091 \\
    & All Layers                &        0.367±0.187 & --- &        0.265±0.139 &        0.114±0.089 & --- &        0.128±0.090&\textbf{0.815±0.085}& --- &        0.793±0.081 &        0.433±0.089 & --- &        0.530±0.085 \\
    \hline
    \textbf{Classification} & \textbf{Configuration} & 
    \multicolumn{3}{c!{\color{lightgray}\vrule}}{\textbf{IoU$\bm{\Downarrow}$}} & \multicolumn{3}{c!{\color{lightgray}\vrule}}{\textbf{Center-of-Mass Distance$\bm{\Uparrow}$}} & \multicolumn{3}{c!{\color{lightgray}\vrule}}{\textbf{Insertion$\bm{\Downarrow}$}} & \multicolumn{3}{c}{\textbf{Deletion$\bm{\Uparrow}$}} \\
    \cline{3-14}
    \multirow{5}{*}{\textbf{Incorrect}} 
    & Final Conv               & --- &    \textbf{0.235±0.206} & --- & ---                &\textbf{0.181±0.106} & --- & --- &               \textbf{0.558±0.118}& --- & --- &                 \textbf{0.618±0.065}& --- \\
    & Final dense block       &        0.251±0.234 & --- &        0.132±0.214 &\textbf{0.168±0.106}& --- &\textbf{0.187±0.103}&        0.621±0.105 & --- &         0.617±0.077 &\textbf{0.569±0.087}& --- &\textbf{0.537±0.106}\\
    & Final two dense blocks  &        0.227±0.229 & --- &        0.101±0.129 &        0.162±0.097 & --- &        0.170±0.090 &        0.607±0.100 & --- &         0.550±0.060 &        0.515±0.099 & --- &        0.496±0.089 \\
    & Final three dense blocks&        0.226±0.228 & --- &        0.097±0.115 &        0.162±0.097 & --- &        0.169±0.091 &        0.586±0.105 & --- &         0.527±0.072 &        0.467±0.098 & --- &        0.455±0.087 \\
    & All Layers              &\textbf{0.227±0.226}& --- &\textbf{0.093±0.100}&        0.162±0.098 & --- &        0.169±0.090 &\textbf{0.579±0.106}& --- & \textbf{0.516±0.082}&        0.441±0.096 & --- &        0.438±0.094 \\
    \hline

\end{tabular}

\label{tab:densenet121_ablation_study_polyp}
\end{table*}

\bibliographystyle{IEEEtran}
\bibliography{bibliography}
\balance
% \section*{Biography Section}

\begin{IEEEbiography}[{\includegraphics[width=1in,height=1.25in,clip,keepaspectratio]{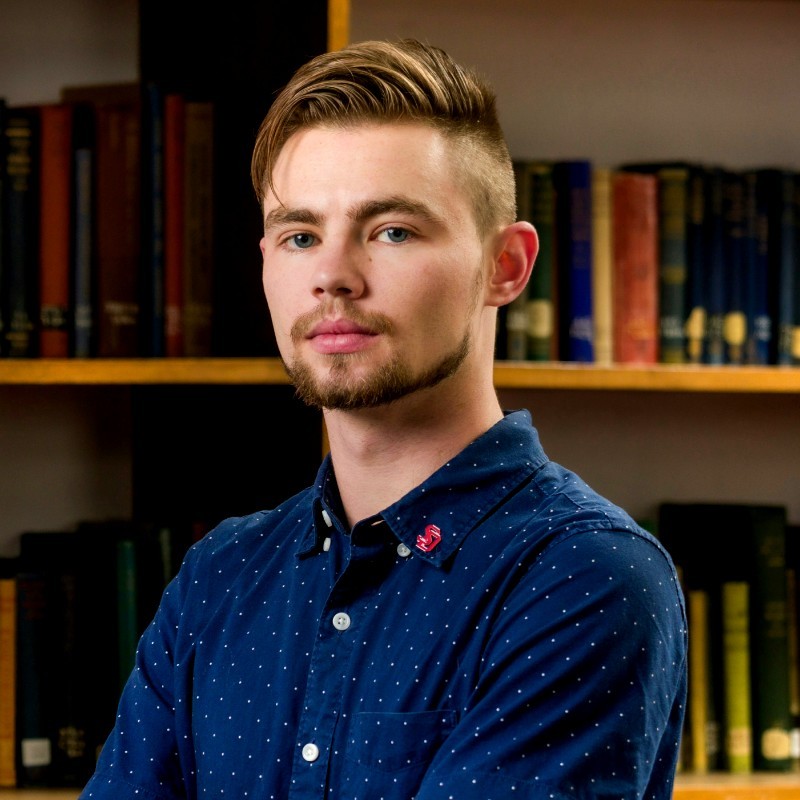}}]{Casey Wall} is a Ph.D. student in Data Science and Engineering at the University of South Dakota and a researcher with USD Artificial Intelligence Research. He received the B.S. degree from Morningside University and the M.S. degree from the University of South Dakota, where he was supported by the NSF Research Traineeship Program. His master's thesis focused on explainable artificial intelligence and biometrics, and he is the author of a Springer-published book on the subject. He has conducted research at the L3i Laboratory, La Rochelle Université, France, contributing to projects in historical document image analysis, with publications in IJDAR and ICADL.
\end{IEEEbiography}

\begin{IEEEbiography}[{\includegraphics[width=1in,height=1.25in,clip,keepaspectratio]{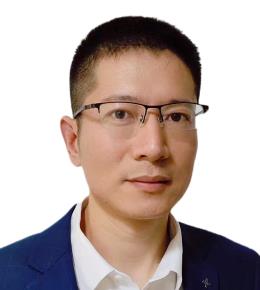}}]{Longwei Wang}
\textit{(Member, IEEE)} is an Assistant Professor of Computer Science at the University of South Dakota. He received his Ph.D. from Auburn University. His research focuses on robust and interpretable AI, integrating equivariant neural architectures with explainability-guided refinement. He also works on multimodal learning, signal processing, and wireless networks. Dr. Wang serves on technical program committees for major AI conferences and reviews for leading journals in AI, machine learning, and wireless networks.
\end{IEEEbiography}

\begin{IEEEbiography}[{\includegraphics[width=1in,height=1.25in,clip,keepaspectratio]{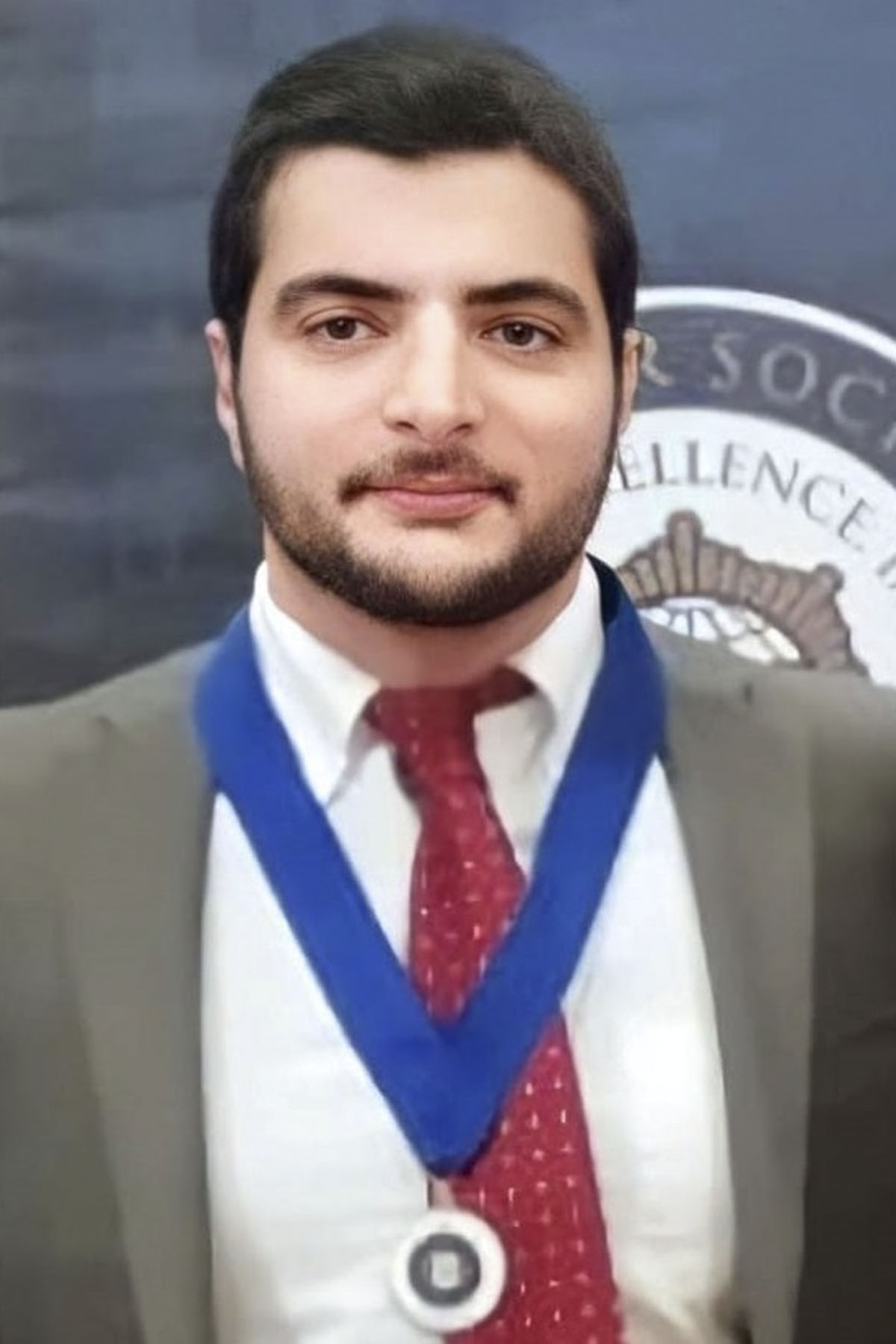}}]{Rodrigue Rizk} \textit{(Member, IEEE)} received the B.E. degree in Computer Engineering (Summa Cum Laude) from Notre Dame University, where he was valedictorian. He received the M.S. and Ph.D. degrees in Computer Engineering from the University of Louisiana at Lafayette, maintaining a perfect GPA. He is currently a professor and Graduate Program Director with the University of South Dakota, Vice Director of Engineering with USD Artificial Intelligence Research, and Chair of IEEE Siouxland. His research interests include artificial intelligence, deep learning, reinforcement learning, mechanistic interpretability, quantum computing, neuromorphic and high-performance computing, AI hardware acceleration, agentic AI, and healthcare technologies. Dr. Rizk is the author of the book \textit{Cracking the Machine Learning Code: Technicality or Innovation?} (Springer, 2024). He has authored numerous peer-reviewed publications in top-tier venues, serves on the Technical Program Committees of several leading international conferences, and acts as a reviewer for major journals and NSF and NIH review panels. He is a Lifetime Member of Phi Kappa Phi and ACM, and an active member of IEEE, including the IEEE Standards Association, the IEEE Computer Society, and the IEEE Quantum Technical Community. He is the recipient of the prestigious Richard G. and Mary B. Neiheisel Endowed Fellowship and has received multiple honors, including the President’s Award for Educational Excellence and Outstanding Academic Achievement and the Ragin’ Leadership Academy Award.
\end{IEEEbiography}

\begin{IEEEbiography}[{\includegraphics[width=1in,height=1.25in,clip,keepaspectratio]{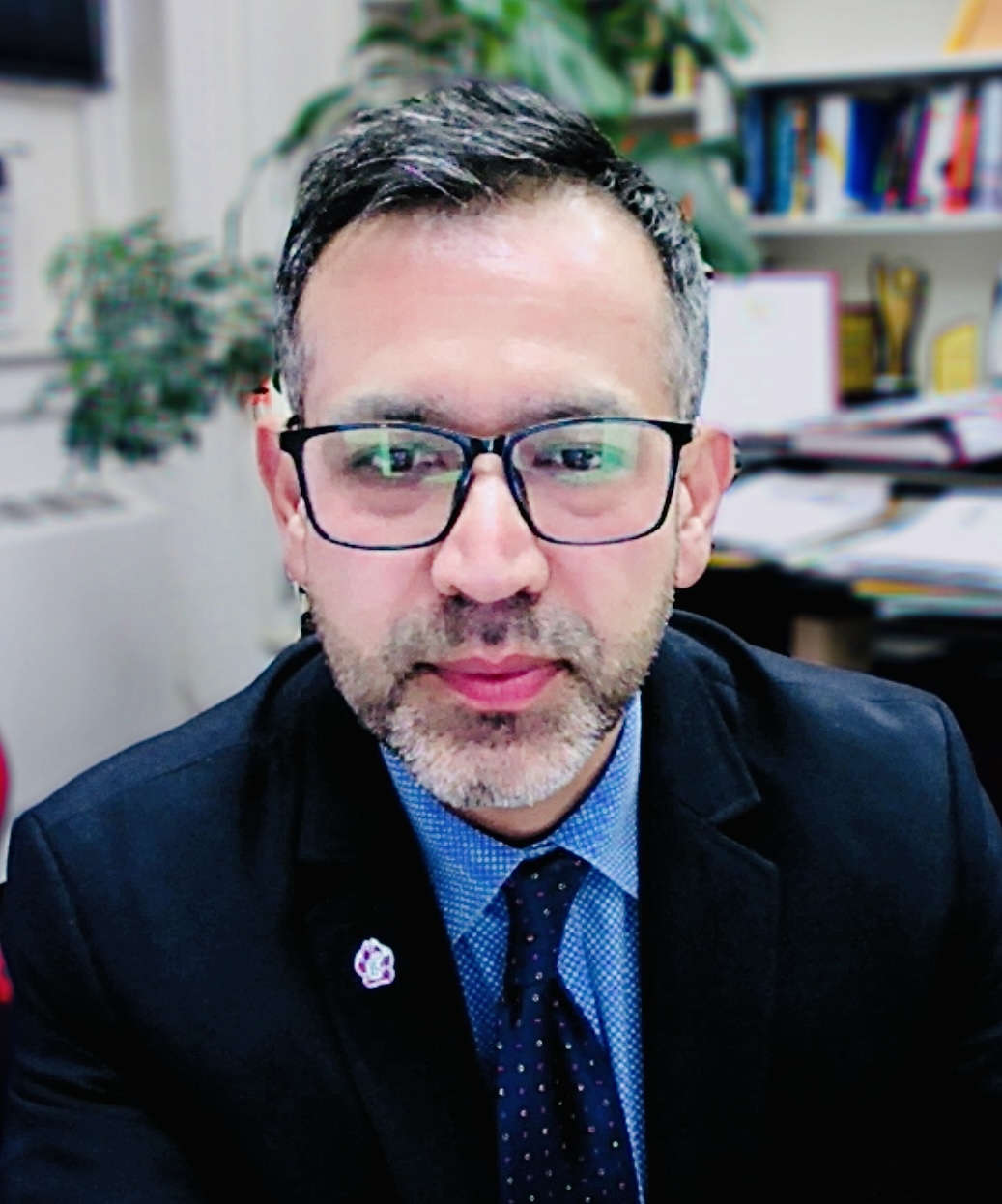}}]{KC Santosh} \textit{(Senior Member, IEEE)} – highly Accomplished AI expert – is Chair of the Department of Computer Science at the University of South Dakota (USD), where he also serves as Founding Director of the USD Artificial Intelligence Research. His extensive background includes serving as Research Fellow at NIH and Postdoctoral Research Scientist (ITESOFT co.) at LORIA Research Center, France. With over \$9 million in funding from sources such as the DOD, NSF, ED, and SDBOR, 12 published books, and more than 300 peer-reviewed research articles (including in IEEE TPAMI, IEEE TAI, and IEEE TMI), he contributes as Associate Editor for prestigious journals such as IEEE Transactions on AI, IEEE Transactions on Medical Imaging, International Journal of Machine Learning and Cybernetics, and International Journal of Pattern Recognition \& AI, and actively serves as Program Chair for reputed conferences such as the IEEE Conference on AI, CogMI, CBMS, ICDAR and GREC, and serves as a U.S. Speaker for AI Education as well as a member of the NIST Center for AI Standards and Innovation.
\end{IEEEbiography}

\end{document}